\documentclass{article} %
\def\wtcaccent{DB7043} %
\PassOptionsToPackage{table}{xcolor}
\usepackage{amsmath,amssymb,amsthm}
\usepackage{style}
\usepackage{natbib}
\usepackage{amsmath,amssymb}
\usetikzlibrary{arrows.meta,positioning}

\usepackage{amsmath,amsfonts,bm}

\def\eqref#1{equation~\ref{#1}}

\def\1{\bm{1}}

\DeclareMathAlphabet{\mathsfit}{\encodingdefault}{\sfdefault}{m}{sl}
\SetMathAlphabet{\mathsfit}{bold}{\encodingdefault}{\sfdefault}{bx}{n}

\usepackage{url}
\usepackage[most]{tcolorbox}
\usepackage{booktabs, threeparttable}
\usepackage{makecell}
\usepackage{amsmath}
\usepackage{subcaption}
\usepackage[table]{xcolor}
\usepackage{pifont}
\usepackage{enumitem}
\usepackage{simpleicons}
\usepackage{fontawesome5}
\newcommand{\websiteurl}{https://secretly-elastic.ashita.nl}
\newcommand{\githuburl}{https://github.com/WalterSimoncini/secretly-elastic}

\usepackage[colorlinks=true,linkcolor=accent,citecolor=accent,filecolor=accent,urlcolor=accent]{hyperref}

\newcommand{\deltapos}[1]{\textcolor{green!55!black}{#1}}
\newcommand{\deltaneg}[1]{\textcolor{red!70!black}{#1}}

\definecolor{plotorange}{HTML}{EB6834}
\definecolor{plotteal}{HTML}{009E73}
\definecolor{plotblue}{HTML}{0072B2}

\definecolor{recipehighlight}{HTML}{fdede7}
\newcommand{\recipetag}[1]{{\setlength{\fboxsep}{1pt}\colorbox{recipehighlight}{\strut #1}}}

\title{Your Transformer Is Secretly Elastic}

\author{Lorem Dolor}

\newtcolorbox{walter}{colback=pink!20, colframe=pink!90!black, title=Walter}

\begin{document}
\begin{wtctitleblock}
\wtctitle{Your Transformer Is Secretly Elastic}{}
\begin{center}
{\sffamily\bfseries Walter Simoncini\textsuperscript{1}, Dheeraj Varghese\textsuperscript{1}, Gertjan J.~Burghouts\textsuperscript{2}, Cees G.M.~Snoek\textsuperscript{1}\par}
\smallskip
{\sffamily\small
  \textsuperscript{1}University of Amsterdam\quad
  \textsuperscript{2}TNO, Intelligent Imaging\par}
\end{center}
\begin{abstract}
Scaling up transformers has produced models with broad capabilities, deployed across heterogeneous hardware and fluctuating workloads. Yet, model families offer only a few fixed sizes (e.g., S, B, L), forcing users to commit to a model based on hardware and latency constraints. This rigidity is often suboptimal under varying query loads. We introduce a retraining-free method that elastifies pre-trained transformers into a single model that adapts on-the-fly across a continuum of computational budgets. Our key contribution is a differentiable relaxation of the functional unit-ranking problem: unlike prior work, which optimizes a single importance weight per MLP block or attention head, we rank every individual MLP neuron and attention head jointly, on a common scale, by minimizing the model's loss under soft pruning across many sparsity targets, all at once. Across 27 vision and text transformer encoders ranging from 22M to 8B parameters, our method yields smooth degradation up to 60\% sparsity and outperforms one-shot baselines under deep pruning. For example, pruning DINOv3 ViT-H+/16 to 50\% sparsity removes 427M parameters while reducing ImageNet-1k linear-probing accuracy by only 0.8\%, without the need for weight correction or retraining. At 60\% sparsity, a pruned AugReg ViT-B/16 model achieves up to $1.80\times$ higher throughput and $1.26\times$ lower single-query latency on an A100 GPU. Elastic models are produced in minutes, with construction time scaling sub-linearly with parameter count.
\end{abstract}
\wtctitlefooter{%
  \textbf{Correspondence:} \href{mailto:w.simoncini@uva.nl}{w.simoncini@uva.nl}\par\smallskip
  {\fontsize{9}{11}\selectfont
  \href{\websiteurl}{{\color{accent}\faGlobe}~\textbf{Website}}\qquad
  \href{\githuburl}{{\color{accent}\simpleicon{github}}~\textbf{GitHub}}}
}
\end{wtctitleblock}
\begin{figure}[!h]
  \centering
  \includegraphics[width=\linewidth]{media/main-figure.pdf}
  \caption{\textbf{Differentiable soft pruning.} Initial importance scores for MLP neurons and attention heads are transformed into learnable log-scores and median-centered by type, aligning their reference points without changing within-type rankings. At each optimization step, the joint ranking determines a threshold $\kappa_b$ and soft mask $M_b$ for each sparsity target $b \in \mathcal{B}$. We run the same calibration batch through the frozen model under each mask, average losses across targets, and update only the log-scores. Mask gradients concentrate near each threshold, so multiple targets expose different regions of the ranking to useful gradients. The optimized ranking can then be hard-pruned to any inference budget without further optimization.}
  \label{fig:main-figure}
\end{figure}

\clearpage
\section{Introduction}
Advances in pre-training recipes have enabled scaling frontier AI models to trillions of parameters, achieving impressive results in a variety of tasks. Their deployment spans applications with very different computational requirements, from offline processing to individual requests with strict latency limits. However, these varied requirements are poorly served by the few fixed model sizes typically released. Under fluctuating workloads, a model chosen for typical demand may be too expensive during a traffic peak, while a smaller model may sacrifice performance when more compute is available. Structured pruning~\citep{li2017pruning, molchanov2019importance, fptp, optin, berisha2025variance} can produce intermediate model sizes, but each pruned model is fixed to its target budget. Serving $k$ budgets this way requires producing, storing, and switching between $k$ pruned models as demand changes. For retrieval, separate indices may also be needed if embeddings from different pruned models are incompatible. Elastic inference instead lets a single model operate at different computational budgets chosen at inference time. Many existing methods require a specific nested architecture and training from scratch~\citep{devvrit2024matformer} or substantial fine-tuning~\citep{valipour2023sortednet, flextron, zhu2025ea, taghibakhshi2025nemotron}, which is costly for billion-parameter models. Choosing suitable data for fine-tuning can also be difficult when the model's pre-training distribution is unknown.

\citet{simoncini2026elastic} make pre-trained vision transformers elastic without retraining by optimizing importance scores across multiple computational budgets. Their evolutionary search adjusts one scaling factor per MLP block and attention head, allowing sparsity to be redistributed across the network while keeping the search tractable. However, this parameterization preserves the initial neuron ordering within each MLP block. Retaining an important neuron with a low initial score therefore requires retaining every higher-ranked neuron in that block. Building on their elastic formulation, we introduce \textit{differentiable soft pruning} to recover more of this \emph{hidden elasticity} by optimizing an individual importance score for \textit{every} MLP neuron and attention head.

The key step is to replace hard truncation of the ranking with continuous masks, making the model's loss differentiable with respect to individual importance scores. Continuous masks are well established in differentiable pruning~\citep{louizos2018learning, xia2022structured, xia2024sheared}, but each of these methods learns a single mask for a fixed sparsity target, typically also updating the model weights. We instead aim to improve the ranking using information from multiple sparsity targets while keeping the pre-trained weights frozen. Our hypothesis is that this joint optimization will yield better masks for any sparsity target. To this end, at each optimization step we determine a pruning cutoff in the current ranking for each selected sparsity target and soften the hard selection with a sigmoid mask. We average the resulting losses to jointly update an individual importance score for every MLP neuron and attention head. These updates reflect how units function together under pruning and allow neuron orderings to change within each block. Gradient descent makes this search tractable with any differentiable objective, including the model's original training loss. Because mask gradients are strongest near each cutoff, optimizing multiple budgets exposes different regions of the ranking to useful gradients. In our experiments, the jointly optimized ranking outperforms separate rankings optimized for individual targets. The resulting ranking also supports hard pruning at intermediate budgets without further optimization.

We evaluate 27 vision and text transformer encoders on image classification, semantic segmentation, text embedding, and retrieval. While the evaluated pruning baselines remain competitive under shallow pruning, our method improves the performance-compute trade-off across a range of sparsity levels, with the largest gains under deep pruning. For example, pruning DINOv3 ViT-H+/16 to 50\% sparsity removes 427M parameters with only a 0.8\% drop in ImageNet-1k linear probing accuracy, without weight correction or retraining. Furthermore, savings translate to speedups: AugReg ViT-B/16 reaches up to $1.80\times$ dense throughput on an A100 at 60\% sparsity.

\section{Related Works}

\noindent\textbf{Network pruning.} Network pruning~\citep{lecun1989optimal, hassibi1992second} compresses models by removing parameters while limiting performance degradation. Unstructured pruning~\citep{han2015learning, frankle2018the, lee2021layeradaptive} removes individual weights, semi-structured pruning~\citep{frantar2023sparsegpt} enforces regular sparsity patterns, and structured pruning~\citep{li2017pruning, molchanov2019importance, fptp, optin, berisha2025variance} removes entire functional units, such as MLP neurons or attention heads. We adopt structured pruning, which enables inference acceleration without specialized hardware. To decide which weights or units to remove, several criteria have been developed, including weight magnitude~\citep{han2015learning}, second-order saliency~\citep{lecun1989optimal, hassibi1992second}, Wanda~\citep{sun2024simple}, which scales weight magnitudes by input activation norms, and magnitude-augmented SNIP~\citep{kohama2023single}, which combines gradient-based loss sensitivity with a magnitude term. We initialize unit scores using a Hessian-diagonal approximation of second-order saliency, then optimize the induced unit ranking.

Beyond parameter pruning, token pruning~\citep{rao2021dynamicvit, Tang_2022_CVPR, kong2022spvit, bergner2025token} and merging~\citep{bolya2023token, kim2024token, tran2024accelerating} speed up inference by reducing the sequence length. These methods act on a different axis than ours, since they shrink the number of tokens while we shrink the network, and they are therefore complementary. Ours keeps all tokens, which preserves dense patch-level features.

Differentiable pruning methods make pruning trainable by replacing the hard keep/remove decision with a continuous mask, so the loss can be minimized by gradient descent~\citep{louizos2018learning, savarese2020winning}. 
\citet{louizos2018learning} parameterize masks as hard-concrete gates, optimized jointly with network weights under an expected $L_0$ penalty. This formulation extends to attention heads~\citep{voita-etal-2019-analyzing} and multiple structural granularities~\citep{xia2022structured}. Sheared LLaMA~\citep{xia2024sheared} further constrains these masks to produce a prescribed target architecture. Alternatives include differentiable attention-head selection~\citep{li-etal-2021-differentiable}, learning weight importance scores through a straight-through estimator~\citep{sanh2020movement}, and learning layer-wise pruning thresholds~\citep{kusupati2020soft, azarian2020learned}. 
We use the same idea: a soft mask over every neuron and head lets us optimize importance scores directly against the loss. Our approach differs in two ways. First, these methods produce one sparse network per sparsity target, whereas we learn a single ranking of all units from which any sparsity can be obtained by removing units from the bottom. Second, they typically train the model weights together with the masks, whereas we keep the pre-trained weights frozen for efficiency.

\noindent\textbf{Elastic inference.} Elastic models adapt their computational cost at inference time to meet different compute budgets depending on request load or resource availability. Early work explored elastic CNNs~\citep{yu2018slimmable, yu2019universally, vu2020any, grimaldi2022dynamic}. For transformers, approaches include early exit~\citep{xin2020deebert, zhou2020bert}, variable depth~\citep{Fan2020Reducing, zhu2025ea, taghibakhshi2025nemotron}, and nested MLP widths~\citep{devvrit2024matformer}, attention heads~\citep{flextron, haberer2024hydravit}, or embedding dimensions~\citep{valipour2023sortednet, zhang2024slicing}. At inference, subnetworks can be selected manually, chosen automatically given a compute budget~\citep{zhu2025ea, taghibakhshi2025nemotron, flextron}, or selected adaptively based on the input~\citep{flextron, hojjat2026thinkingvit}. These approaches require training from scratch~\citep[e.g.,][]{devvrit2024matformer} or substantial fine-tuning~\citep[e.g.,][]{flextron, valipour2023sortednet, zhu2025ea, taghibakhshi2025nemotron}.
\cite{simoncini2026elastic} showed that pre-trained ViTs can be made elastic without retraining, by searching for a global importance ranking with evolutionary search. To keep the black-box search tractable, they optimize one scale per MLP block and attention head, leaving the within-block neuron order fixed to the initial scores. We keep their retraining-free, ranking-based formulation, but replace the search with gradient-based optimization, which makes per-neuron scores tractable to optimize and lets the ranking be repaired within a block.

\section{Method}

We aim to make a pre-trained transformer elastic: a single model that can be pruned to any computational budget at test time, without retraining. To this end, we rank a transformer's functional units, i.e., its MLP neurons and attention heads. Whereas standard pruning commits to a single sparsity, i.e., a fraction of FLOPs removed, a ranking realizes any sparsity by removing units from its bottom until the target is met. We first estimate an initial importance score for every unit (Section~\ref{sec:method-importance-scores}). These scores induce an initial ranking, but one that may not reflect downstream performance or be directly comparable across blocks and unit types. We therefore propose an algorithm that optimizes this ranking via differentiable soft pruning, placing all units on a common scale and preserving performance across sparsity targets (Section~\ref{sec:method-score-refinement}). An overview of our method is shown in~\autoref{fig:main-figure}.

\subsection{Preliminary: Importance Score Estimation}
\label{sec:method-importance-scores}

Ranking units requires per-unit importance scores, which we initialize following prior work. Given a transformer with $L$ blocks, MLP hidden dimension $H$ and $A$ attention heads per block, we seek MLP scores $s^m \in \mathbb{R}_{\geq0}^{L \times H}$ and head scores $s^h \in \mathbb{R}_{\geq0}^{L \times A}$. The importance of a parameter $\theta_i$ is often defined as the increase in a loss $\mathcal{L}$ caused by its removal, which a second-order Taylor expansion around the weights $\theta$ of a converged model reduces to the Hessian~\citep{lecun1989optimal, hassibi1992second}. We approximate the Hessian by its diagonal and estimate it as the empirical Fisher over $N$ calibration samples~\citep{fptp, simoncini2026elastic}:
\begin{equation}
    \mathcal{H}_i \approx \frac{1}{N} \sum^{N}_{k=1} \left( \frac{\partial \mathcal{L}_k}{\partial \theta_i} \right)^2.
\end{equation}
The per-sample loss $\mathcal{L}_k$ is modality specific: for vision encoders we use the self-supervised DINO objective~\citep{caron2021emerging}, which yields gradients for supervised and self-supervised models alike, and for text encoders the symmetrized in-batch InfoNCE objective~\citep{oord2018representation, gunther2023jina}. Both are detailed in Appendix~\ref{sec:appendix-importance-scores}.

In either case, scores $\mathcal{H}_i$ are per-weight, whereas we prune whole units, so we average scores over the input-projection column and output-projection row for an MLP neuron, and, following~\citet{simoncini2026elastic}, over the value-projection slice for an attention head, yielding initial scores $s^m$ and $s^h$. Yet, these scores are a weak basis for pruning: ($i$) the diagonal approximation ignores unit interactions, ($ii$) the proxy loss may not align with downstream performance, and ($iii$) scores are not comparable across blocks~\citep{molchanov2017pruning} and unit types. 

\cite{simoncini2026elastic} optimize initial importance scores through black-box search over one score scaling factor per MLP block and attention head. This enables reallocating sparsity across blocks and ranking individual heads while keeping the search space dimensionality manageable. However, neuron rankings within each MLP block remain fixed: keeping an important neuron with a low initial score also requires keeping all higher-ranked neurons in that block. We mitigate these limitations by optimizing all unit scores jointly, on a common scale, across sparsity targets via differentiable soft-pruning, which we introduce next.

\subsection{Differentiable soft pruning}
\label{sec:method-score-refinement}
Starting from the initial importance scores, we reorder functional units to preserve performance at every sparsity target, by minimizing the loss under pruning. Pruning, however, is a hard truncation of the ranking, and is hence non-differentiable with respect to the importance scores. We therefore relax pruning into a soft mask, so that the loss becomes differentiable with respect to scores, and optimize them directly via gradient descent while keeping the model weights frozen. We next describe the score parametrization, the soft masking, and the objective.

\noindent\textbf{Score parametrization.} We use learnable log-scores $z^m$ for MLP neurons and $z^h$ for attention heads, which can vary freely while their exponentials remain positive. To establish a common reference point for the joint ranking, we center each score type once at initialization: $z^m {=} \log(s^m) {-} \text{median}(\log(s^m))$ and $z^h {=} \log(s^h) {-} \text{median}(\log(s^h))$. This aligns their medians at zero without changing either distribution's spread or the ranking within each type. We use the median as it is less sensitive to outliers than the mean. We then jointly optimize the combined log-scores $z$, which induce a single ranking over all units.

\noindent\textbf{Soft masking.} For a sparsity target $b$, hard pruning walks the ranking from the top and keeps units until their cumulative FLOP cost reaches $(1 {-} b)$ of the dense model. Let $\kappa_b$ be the score of the last unit kept, held constant via stop-gradient. We relax the hard selection $\mathbf{1}[z \geq \kappa_b]$ to soft masks $M^m_b$ for MLP neurons and $M^h_b$ for attention heads, given by a sigmoid of the signed distance from the threshold at temperature $\tau$:
\begin{equation}
    M^m_b = \sigma \left( \frac{z^m - \kappa_b}{\tau} \right) \in (0, 1)^{L \times H},
    \quad
    M^h_b = \sigma \left( \frac{z^h - \kappa_b}{\tau} \right) \in (0, 1)^{L \times A}.
\end{equation}

Scores above the threshold receive mask values closer to one, while scores below it receive values closer to zero. The temperature $\tau$ controls the sharpness of this transition. We recompute $\kappa_b$ at each optimization step so that the cutoff continues to reflect its associated sparsity target as scores and the unit ranking change. During optimization, the soft masks scale activations in the forward pass as:
\begin{equation}
    \text{MLP}_l = (\phi(xW_1) \odot M_{b,l}^m)W_2
    \quad
    \text{Attn}_l = \left( \text{Cat}\left( \text{head}_1(x), ..., \text{head}_A(x) \right) \odot M_{b,l}^h \right)W_o,
\end{equation}
where $l$ indexes the transformer block, and each head's mask scales all its output channels. 

\noindent\textbf{Objective.} As our goal is an elastic model covering a wide sparsity range, we optimize scores by minimizing the loss of masked models over a set $\mathcal{B}$ of sparsity targets:
\begin{equation}
    z^* = \arg\min_z \frac{1}{|\mathcal{B}|} \sum_{b \in \mathcal{B}} \mathbb{E}_{(x,y) \sim D} \left[ \ell \left( f\left(x, M_b(z)\right), y \right) \right],
\end{equation}
where $f(x, M_b(z))$ is the network with the mask for sparsity target $b$ applied, $D$ the calibration set, and $\ell$ the optimization loss. We use the model's native loss where applicable, e.g., cross-entropy for a classifier, and otherwise minimize the cosine distance between soft-pruned and dense embeddings, with $y$ denoting the dense embedding. The masked model's loss reflects how units function together under pruning, capturing interactions that the Hessian-diagonal initialization ignores, and optimizing a separate score for every MLP neuron and attention head gives us full freedom to reorder units in response to these interactions.

\noindent\textbf{Optimization details.} At each optimization step, we average the loss over 14 sparsity targets, from 5\% to 70\% in 5\% increments, using the same calibration batch. For a single sparsity target, units far above or below the pruning threshold have mask values near one or zero, where the sigmoid is nearly saturated. Their scores therefore receive little gradient through the masks, while units near the threshold are more sensitive to score changes. Across multiple targets, a unit far from one threshold may lie close to another and receive a useful gradient from that target. We hypothesize that this broader gradient signal, together with the freedom to reorder individual units, helps preserve performance across sparsity targets. The final scores $s^* = \exp(z^*)$ define the ranking used for pruning at inference time. Unlike methods that sparsify uniformly across blocks~\citep{frantar2023sparsegpt}, search locally~\citep{fptp}, or keep separate rankings per unit type~\citep{simoncini2026elastic}, we let the optimization decide where sparsity goes, across blocks and functional units.

\section{Experiments}
\label{sec:experiments}

We compare against one-shot pruning baselines on vision and text tasks in Section~\ref{sec:preserve-representations}, study the effect of model size in Section~\ref{sec:scaling}, evaluate elastic deployment against a fixed dense index and measure speedups in Section~\ref{sec:elastic-inference}, and ablate our design choices in Section~\ref{sec:ablations}. Unless otherwise specified, we evaluate six sparsity targets from 10\% to 60\% in 10\% increments, report the computational cost in theoretical GFLOPs, and evaluate all pruning methods without weight correction to isolate the effect of unit selection, deferring weight correction to Appendix~\ref{sec:appendix-weight-correction}. %

\subsection{Comparison with Pruning Methods}
\label{sec:preserve-representations}

\noindent\textbf{Setup.} We evaluate pruned vision encoders through k-NN classification on ImageNet-1k~\citep{deng2009imagenet}, in-context semantic segmentation on Pascal VOC 2012~\citep{everingham2010pascal} with the Hummingbird protocol~\citep{balazevic2023towards}, and pruned text encoders on the 41-task MTEB(eng, v2) suite~\citep{muennighoff2023mteb, enevoldsen2025mmteb}. For classification, we consider five ViT-B/16 models from the AugReg~\citep{steiner2022how}, DINOv3~\citep{simeoni2026dinov3}, DeiT~\citep{touvron2021training}, DeiT-III~\citep{touvron2022deit}, and SigLIPv2~\citep{tschannen2025siglip} families. For segmentation, we consider AugReg, which supports all baselines, and DINOv3, for its strong patch-level representations. For both vision tasks, we compare our pruning method against five one-shot correction- and retraining-free baselines: FPTP~\citep{fptp}, SparseGPT~\citep{frantar2023sparsegpt}, OPTIN~\citep{optin}, VBP~\citep{berisha2025variance}, and SnapViT~\citep{simoncini2026elastic}. FPTP and OPTIN require a classification head, so we evaluate them only on the supervised backbones. For text, we consider the bidirectional BGE-L~\citep{xiao2024c} and E5-L~\citep{wang2022text} encoders and the decoder-based Qwen3 0.6B~\citep{zhang2025qwen3} and E5-Mistral 7B~\citep{wang2024improving} embedders, and compare against SnapViT and SlimGPT~\citep{ling2024slimgpt}.

\begin{figure}[t]
  \centering
  \includegraphics[width=\linewidth]{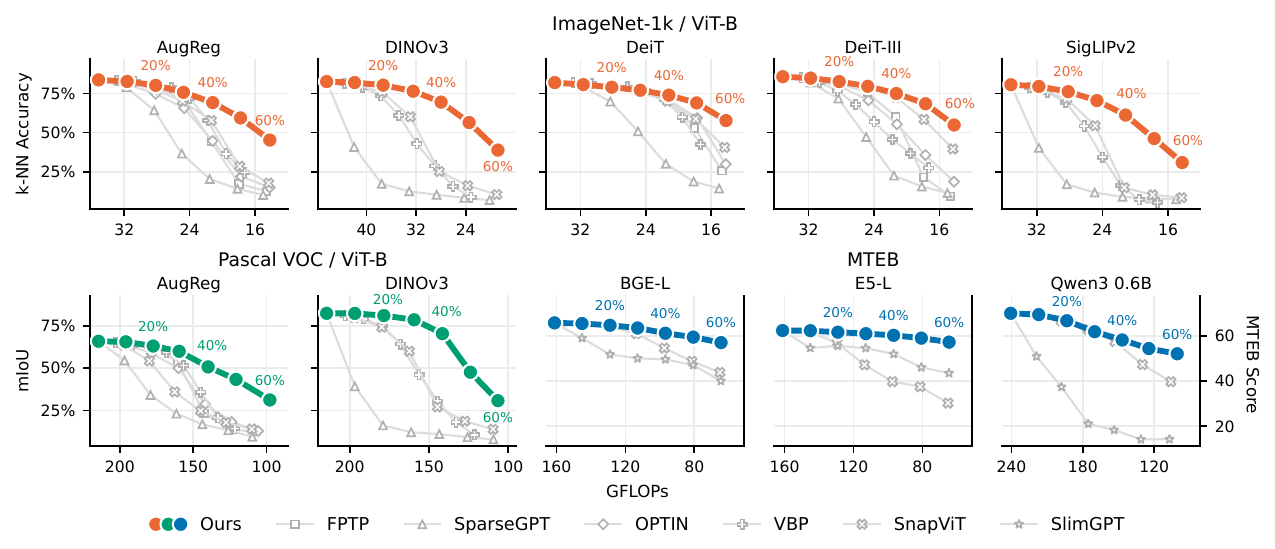}
  \caption{\textbf{Our method yields smooth degradation curves across vision and text tasks.} Performance versus compute for our method and six baselines from dense to 60\% sparsity. Curve labels indicate FLOP sparsity. Top: ImageNet-1k top-1 k-NN accuracy. Bottom left: Pascal VOC 2012 Hummingbird mIoU. Bottom right: mean MTEB score. Baselines remain competitive under shallow pruning, but our method retains higher performance as sparsity increases, with the largest gains under moderate and deep pruning.}
  \label{fig:results-overview}
\end{figure}

\noindent\textbf{Image classification.} The top row of \autoref{fig:results-overview} shows ImageNet-1k k-NN accuracy for the five ViT-B/16 backbones. Baselines are competitive under shallow pruning, with SnapViT within 0.9\% of our method on DeiT and DeiT-III at up to 30\% sparsity. Past 30\%, every baseline degrades sharply while our method degrades smoothly, leading the runner-up by 9.6\% on DeiT to 40.3\% on DINOv3 at 50\% sparsity, and by 15.3\% on DeiT-III to 28.4\% on DINOv3 at 60\%. The gap is largest on the self-supervised and image-text backbones: at 40\% sparsity, our method retains 69.4\% on DINOv3 and 61.1\% on SigLIPv2, whereas every baseline falls below 30\%. Our method thus widens the usable sparsity band of every backbone, matching the baselines under shallow pruning and retaining substantially higher accuracy past 30\% sparsity. Appendix~\ref{sec:appendix-additional-backbones} extends the comparison to nine transfer datasets and linear probing, where the same trends hold, except on EuroSAT~\citep{helber2019eurosat}, where SnapViT and OPTIN retain higher accuracy.

\noindent\textbf{In-context semantic segmentation.} The bottom-left panels of \autoref{fig:results-overview} report mIoU on Pascal VOC 2012 for AugReg and DINOv3 ViT-B/16 under the retrieval-based Hummingbird protocol. On AugReg, FPTP and VBP match our method up to 20\% sparsity, but our method leads from 30\% onward, by up to 25.1\% over OPTIN, the runner-up, at 50\%. On DINOv3, our method leads at every sparsity: at 30\%, it loses only 3.8\% mIoU against 22.4\% for SnapViT, the runner-up. Past 40\%, however, DINOv3 degrades sharply with our method as well, losing 34.8\% at 50\% sparsity. Our method thus preserves patch-level representations better than the baselines without any task-specific calibration, with the largest gains under moderate pruning.

\noindent\textbf{Text embedding.} The bottom-right panels of \autoref{fig:results-overview} report the MTEB score for pruned BGE-L, E5-L, and Qwen3 0.6B models. Our method outperforms SlimGPT at every sparsity on all three encoders, by up to 38.0 points on Qwen3 0.6B at 60\%. SnapViT is on par with our method under shallow pruning, exceeding it by at most 1.0 point at up to 30\% sparsity, but falls behind under deep pruning: at 60\%, E5-L retains 57.3 with our method against 30.2 for SnapViT. Bidirectional encoders are more robust. For example, our method loses only 5.1 points on E5-L at 60\% sparsity, against 18.0 on Qwen3 0.6B. Our method thus preserves text embeddings more consistently than the baselines under deep pruning, on bidirectional and decoder-based encoders alike. Appendix~\ref{sec:appendix-text-embeddings} extends the comparison to all nine text encoders, where the same trends hold: SnapViT is competitive under shallow pruning, but our method outperforms both baselines on every model under deep pruning. It further isolates retrieval tasks, which are more sensitive to pruning.

\subsection{Effect of Model Size}
\label{sec:scaling}

\begin{figure}[t]
  \centering
  \includegraphics[width=\linewidth]{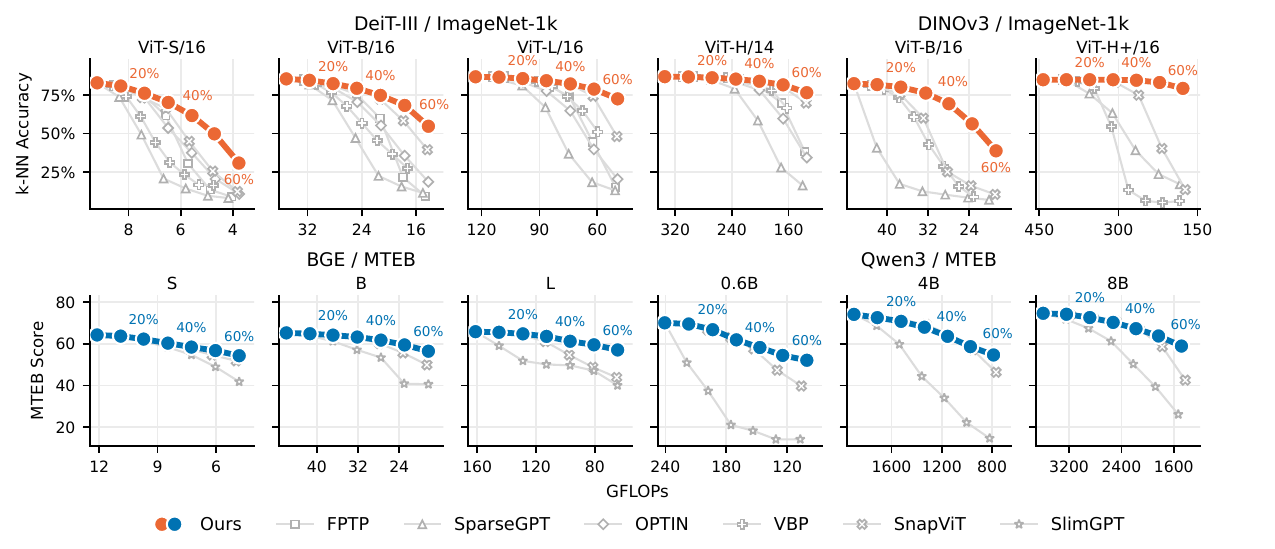}
  \caption{\textbf{Larger vision models tolerate more pruning.} Performance versus compute for our method and baselines from dense to 60\% sparsity. Curve labels indicate FLOP sparsity. Top: ImageNet-1k top-1 k-NN accuracy for DeiT-III and DINOv3. Bottom: mean MTEB score for BGE and Qwen3. Size helps BGE modestly and Qwen3 inconsistently. Our advantage grows with size except on DeiT-III, where SnapViT remains competitive.}
  \label{fig:results-scaling}
\end{figure}

\noindent\textbf{Setup.} We use the same protocols and baselines as in Section~\ref{sec:preserve-representations}, but vary model size within a family. We evaluate DeiT-III from ViT-S/16 to ViT-H/14 and DINOv3 ViT-B/16 and ViT-H+/16 via k-NN on ImageNet-1k, and for text BGE-S, B, and L and Qwen3 0.6B, 4B, and 8B on MTEB. %

\noindent\textbf{Image classification.} The top row of \autoref{fig:results-scaling} shows that degradation flattens with scale. At 50\% sparsity, our method loses 33.2\% on DeiT-III ViT-S/16 but only 5.3\% on ViT-H/14, and 26.1\% on DINOv3 ViT-B/16 against 1.8\% on ViT-H+/16, which stays within 0.4\% of its dense accuracy up to 40\%. Baselines benefit from scale on DeiT-III, where SnapViT matches our method up to 40\% sparsity on ViT-L/16 and ViT-H/14 and trails it by only 1.1\% at 50\% on ViT-H/14. On DINOv3, in contrast, the advantage grows with scale: at 60\% sparsity, our method retains 79.4\% on ViT-H+/16 while every baseline falls below 20\%. Larger models thus tolerate more pruning with our method, while its advantage depends on the backbone, shrinking on supervised DeiT-III and widening on self-supervised DINOv3. Appendix~\ref{sec:appendix-additional-backbones} reports the same trend for the AugReg and SigLIPv2 families and under linear probing, where DINOv3 ViT-H+/16 loses only 0.8\% at 50\% sparsity. Within the DINO WebSSL family~\citep{fan2025scaling}, we further find that pruning larger models can outperform deploying smaller dense counterparts: pruned 3B and 7B models pareto-dominate the dense 2B and 5B models, respectively, as shown in Appendix \autoref{fig:dino-webssl-pareto}.

\noindent\textbf{Text embedding.} The bottom row of \autoref{fig:results-scaling} shows that, unlike in vision, size matters less than family. At 60\% sparsity, BGE losses shrink only from 10.0 MTEB points for BGE-S to 8.7 for BGE-L, while Qwen3 losses go from 18.0 points for 0.6B to 19.5 for 4B and 15.7 for 8B, without a consistent trend. The advantage over SnapViT nonetheless grows with size within each family, from 2.5 to 13.3 points on BGE and from 8.3 to 16.4 points on Qwen3 4B and 8B at 60\% sparsity. Under deep pruning, bidirectional encoders better retain performance than decoder-based embedders regardless of size, while our advantage over the baselines grows with size in both families. Appendix~\ref{sec:appendix-text-embeddings} reports results for all nine encoders, and compares our pruned E5-Mistral 7B~\citep{wang2024improving} with EffiR~\citep{lei2026making} in retrieval.

\subsection{Elastic Deployment}
\label{sec:elastic-inference}

\begin{figure}[t]
  \centering
  \includegraphics[width=\linewidth]{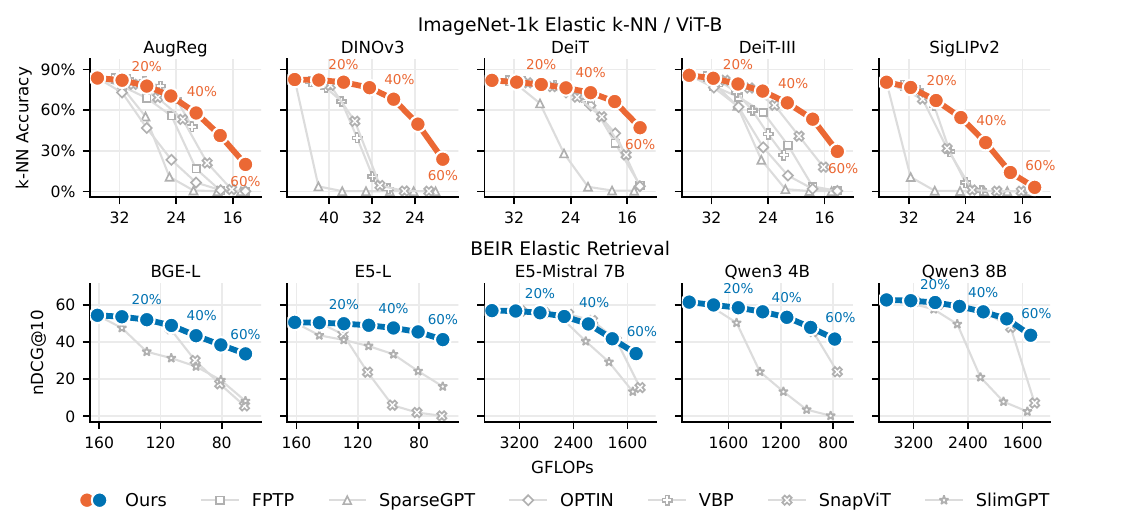}
  \caption{\textbf{Our method better preserves compatibility with a fixed dense index.} Performance versus compute for our method and baselines from dense to 60\% sparsity, with the index encoded once by the dense model and queries encoded by its pruned counterparts. Curve labels indicate FLOP sparsity. Top: ImageNet-1k top-1 k-NN accuracy. Bottom: mean BEIR nDCG@10. Baselines remain competitive under shallow pruning, but our method supports deeper query encoder pruning without rebuilding the index.}
  \label{fig:results-elastic}
\end{figure}

\noindent\textbf{Setup.} The preceding evaluations encode index samples and queries with the same pruned model. In deployment, however, the index can be built once offline with the dense encoder, while incoming queries are encoded elastically depending on the current workload~\citep{devvrit2024matformer}. We evaluate this asymmetric setting with the k-NN protocol of Section~\ref{sec:preserve-representations} on ImageNet-1k for the same five ViT-B/16 backbones, and through text retrieval on BEIR~\citep{thakur2021beir}, using the MTEB implementation, for BGE-L, E5-L, E5-Mistral 7B, and Qwen3 4B and 8B.

\noindent\textbf{Image classification.} In the top row of \autoref{fig:results-elastic}, baselines remain competitive under shallow pruning but collapse under deep pruning: at 50\% sparsity, the best baseline falls below 6\% accuracy on AugReg, DINOv3, and SigLIPv2, while VBP cannot reach high sparsity as it prunes only MLP neurons. Our method instead degrades gradually, leading the runner-up by 13.8\% on SigLIPv2 to 49.4\% on DINOv3 at 50\% sparsity, with its largest lead of 66.2\% on DINOv3 at 40\%. Absolute accuracy is backbone-dependent: at 50\% sparsity, DeiT retains 66.3\% versus 13.9\% for SigLIPv2. Our method thus better preserves compatibility with a fixed dense index than the baselines across pretraining regimes, although deep pruning remains costly on some backbones.

\noindent\textbf{Text retrieval.} The bottom row of \autoref{fig:results-elastic} shows that our method degrades smoothly, losing at most 3.0 nDCG@10 points at 20\% sparsity and 10.9 at 40\%. SnapViT remains competitive under shallow pruning, leading by at most 1.8 points at up to 40\% sparsity, but collapses under deep pruning, where our method outperforms both baselines on every backbone at 50\% and 60\%, by up to 36.5 points on Qwen3 8B at 60\%. The bidirectional E5-L is again the most robust, losing only 1.5 points at 30\% sparsity and 9.3 at 60\%. Elastic queries thus remain compatible with a fixed dense index, allowing query budgets to change without rebuilding the index, and our method extends this compatibility to deep pruning. Appendix~\ref{sec:appendix-elastic-text-retrieval} confirms these trends across all nine encoders.

\noindent\textbf{Measured speedups.} While the preceding experiments report theoretical GFLOPs, practical inference gains also depend on hardware and execution settings. We therefore measure throughput and latency for dense and pruned models. For AugReg ViT-B/16 at 60\% sparsity, we achieve up to $1.80\times$ the dense model's throughput at batch size 128 and a $1.26\times$ speedup in single-query latency on an A100 in fp16, using compilation and GPU-friendly pruning. Switching between budgets has a mean cost of 0.46 ms, excluding the initial compilation. Appendix~\ref{sec:appendix-speedups} details the measurement protocol, execution settings, and the effect of pruning granularity on speed and accuracy. These results show that elastic models support faster inference with a negligible switching overhead.

\subsection{Ablations}
\label{sec:ablations}

In this section, we investigate how score granularity helps address neglected unit interactions, whether ranking optimization can recover from poor initial rankings, and whether optimizing jointly across sparsity targets improves the resulting ranking. Appendix~\ref{sec:appendix-ablations} provides additional ablations on the optimization loss, calibration data, the unified ranking, and optimization settings.

\noindent\textbf{Per-unit optimization mitigates the diagonal approximation.} Our method learns an individual score for every neuron and head, allowing the ranking to adapt to cross-unit interactions ignored by the diagonal initialization. To test how much this freedom matters, we compare progressively finer score parametrizations at equal compute, using the same 300-step optimization protocol and fixed MLP-to-head sparsity allocations. We consider one scaling factor per MLP or MHA sublayer, one per MLP block and attention head as in SnapViT, one per neuron group and attention head with progressively finer groups formed according to the initial importance ranking, and finally an independent score per neuron and head as in our standard parametrization. We implement each scaling factor as an additive offset to the corresponding log-scores. As shown in~\autoref{tab:free-scores}, accuracy increases monotonically with optimization granularity, from 29.4\% with sublayer-level scaling factors to 59.7\% with per-unit scores. Applying our protocol to the SnapViT parametrization improves accuracy from 28.4\% to 34.6\%, but its search space still trails per-unit optimization by 25.1\%. Thus, under the same optimization protocol, allowing finer groups of units to change their relative order improves accuracy, with independent per-unit scores performing best.

\begin{table}[!htbp]
  \begin{minipage}[c]{0.34\columnwidth}
    \captionsetup{justification=raggedright,singlelinecheck=false}
    \caption{\textbf{Per-unit optimization mitigates the diagonal approximation.} ImageNet-1k top-1 k-NN accuracy for AugReg ViT-B/16 at 50\% sparsity under different score parametrizations. Groups are neuron groups per MLP block. All optimized variants use equal compute. Per-unit scores achieve the highest accuracy.}
    \label{tab:free-scores}
  \end{minipage}\hfill
  \begin{minipage}[c]{0.62\columnwidth}
  \centering
  \resizebox{\linewidth}{!}{
  \begin{threeparttable}
    \begin{tabular}{lrc}
      \toprule
      Optimization Granularity & Parameters & ImageNet-1k \\
      \midrule
      None & 0 & 29.4 \\
      Sublayer & 24 & 29.4 \\
      MLP Sublayer + Head (SnapViT)\tnote{$\dagger$} & 156 & 34.6 \\
      8 Groups + Head & 240 & 36.3 \\
      64 Groups + Head & 912 & 43.3 \\
      512 Groups + Head & 6,288 & 51.9 \\
      \rowcolor{recipehighlight} Neuron + Head (Ours) & 37,008 & \textbf{59.7} \\
      \bottomrule
    \end{tabular}
  \begin{tablenotes}
    \small
    \item[$\dagger$] SnapViT's score parametrization under our optimization protocol.
  \end{tablenotes}
  \end{threeparttable}
  }
  \end{minipage}
\end{table}

\noindent\textbf{Robustness to score initialization.} As shown in~\autoref{tab:free-scores}, allowing units to change their relative order independently improves the optimized ranking. We next test whether this flexibility allows us to recover strong rankings from poor initial scores. In \autoref{tab:score-init-ablation} we compare random scores, weight magnitude, Wanda~\citep{sun2024simple}, our default DINO-based scores, and magnitude-augmented SNIP~\citep{kohama2023single}. At 20\% sparsity, optimization lifts Wanda from 7.2\% to 77.8\% accuracy and random initialization from 58.2\% to 76.8\%, placing them within 2.3\% and 3.3\%, respectively, of DINO initialization. Initialization matters more at 60\% sparsity, where DINO leads Wanda by 12.6\% and SNIP by 1.9\%. At 40\% sparsity, SNIP leads DINO by 0.3\%, possibly because its scores use gradients from the model's native cross-entropy objective. Thus, our optimization can repair poor initial rankings, although a strong initialization remains beneficial under deep pruning.

\begin{table}[!htbp]
  \begin{minipage}[c]{0.30\columnwidth}
    \captionsetup{justification=raggedright,singlelinecheck=false}
    \caption{\textbf{Robustness to score initialization.} ImageNet-1k top-1 k-NN accuracy for AugReg ViT-B/16 at 20\%, 40\%, and 60\% sparsity, before (Plain) and after (Ours) ranking optimization. Optimization improves all score initializations, gaining up to 70.6\% for Wanda at 20\% sparsity.}
    \label{tab:score-init-ablation}
  \end{minipage}\hfill
  \begin{minipage}[c]{0.66\columnwidth}
    \centering
    \resizebox{\linewidth}{!}{
      \begin{tabular}{lrrrrrr}
        \toprule
        & \multicolumn{6}{c}{Sparsity} \\
        \cmidrule(lr){2-7}
        & \multicolumn{2}{c}{20\% / 28.2 G} & \multicolumn{2}{c}{40\% / 21.2 G} & \multicolumn{2}{c}{60\% / 14.2 G} \\
        \cmidrule(lr){2-3}\cmidrule(lr){4-5}\cmidrule(lr){6-7}
        Score & Plain & Ours & Plain & Ours & Plain & Ours \\
        \midrule
        Random & 58.2 & 76.8$^{\deltapos{+18.6}}$ & 14.7 & 63.2$^{\deltapos{+48.5}}$ & 7.9 & 35.5$^{\deltapos{+27.6}}$ \\
        Magnitude & 7.3 & 77.0$^{\deltapos{+69.7}}$ & 4.2 & 63.7$^{\deltapos{+59.5}}$ & 2.5 & 33.8$^{\deltapos{+31.3}}$ \\
        Wanda & 7.2 & 77.8$^{\deltapos{+70.6}}$ & 6.2 & 63.1$^{\deltapos{+56.9}}$ & 4.8 & 32.6$^{\deltapos{+27.8}}$ \\
        SNIP Mag. & 78.2 & 79.7$^{\phantom{0}\deltapos{+1.5}}$ & 37.8 & \textbf{69.4}$^{\deltapos{+31.6}}$ & 12.4 & 43.3$^{\deltapos{+30.9}}$ \\
        \rowcolor{recipehighlight} DINO & \textbf{78.6} & \textbf{80.1}$^{\phantom{0}\deltapos{+1.5}}$ & \textbf{42.5} & 69.1$^{\deltapos{+26.6}}$ & \textbf{17.0} & \textbf{45.2}$^{\deltapos{+28.2}}$ \\
        \bottomrule
      \end{tabular}
  }
  \end{minipage}
\end{table}

\noindent\textbf{Joint optimization across sparsity targets improves performance.} An elastic ranking must remain competitive at each deployment budget. We therefore test whether jointly optimizing across sparsity targets compromises performance at individual targets. \autoref{tab:elastic-vs-single} compares joint optimization across 14 sparsity targets from 5\% to 70\% with a separate ranking optimized for each reported target. Compared with single-target optimization, joint optimization improves performance at all three evaluated sparsities, gaining from 0.4\% at 20\% sparsity to 3.6\% at 60\%. These results show that a ranking learned across budgets can outperform separate rankings optimized for individual targets, while avoiding re-optimization when the deployment budget changes.

\begin{table}[htbp]
    \begin{minipage}[c]{0.63\columnwidth}
      \captionsetup{justification=raggedright,singlelinecheck=false}
      \caption{\textbf{Joint optimization across sparsity targets improves performance.} ImageNet-1k top-1 k-NN accuracy for AugReg ViT-B/16. Joint optimization over 14 targets (Elastic) outperforms single-target optimization (Single) at all three sparsities.}
      \label{tab:elastic-vs-single}
    \end{minipage}\hfill
    \begin{minipage}[c]{0.35\columnwidth}
      \centering
      \resizebox{\linewidth}{!}{
        \begin{tabular}{lcc}
          \toprule
          Sparsity & Single & Elastic \\
          \midrule
          \rule{0pt}{2.3ex}20\% / 28.2 G & 79.7 & \cellcolor{recipehighlight}\textbf{80.1}$^{\deltapos{+0.4}}$ \\
          40\% / 21.2 G & 67.2 & \cellcolor{recipehighlight}\textbf{69.1}$^{\deltapos{+1.9}}$ \\
          60\% / 14.2 G & 41.6 & \cellcolor{recipehighlight}\textbf{45.2}$^{\deltapos{+3.6}}$ \\
          \bottomrule
        \end{tabular}
      }
    \end{minipage}
\end{table}

\section{Conclusion}
In this work, we make pre-trained transformer encoders elastic through differentiable soft pruning, jointly optimizing individual neuron and head scores across sparsity targets without any weight updates. The resulting ranking supports pruning up to the deepest optimization target without further optimization. Across 27 vision and text encoders, our method improves the performance-compute trade-off over the evaluated baselines, with the largest gains under deep pruning. Compatibility with a fixed dense index, measured inference speedups, and low budget-switching overhead enable query compute to adapt to fluctuating workloads.

\noindent\textbf{Limitations.} While our method performs strongly on vision and bidirectional text encoders, deep pruning remains challenging for the evaluated decoder-based text embedders. We leave investigating this sensitivity, evaluating our method on diffusion-based image generation and autoregressive language modelling, and exploring elastic post-training using our learned rankings to future work.

\section{Acknowledgements}

This research has received funding from the NWO perspectiefprogramma Foundation models for Industry (FIND) under grant number P23.016. Cees G. M. Snoek is (partially) funded by the Horizon Europe project ELLIOT (GA No. 101214398). We thank SURF for the support in using the National Supercomputer Snellius, and DAS-6 for providing compute used to run the experiments presented in this paper.

\bibliography{iclr2027_conference}
\bibliographystyle{iclr2027_conference}

\newpage
\appendix
\section{Appendix}

\subsection{Datasets}

\noindent\textbf{Image classification.} We evaluate the quality of global representations on ImageNet-1k~\citep{deng2009imagenet} and nine transfer datasets covering natural images, textures, and remote sensing. The datasets span coarse- and fine-grained recognition. \autoref{tab:image-classification-datasets} summarizes the number of classes, training and test set sizes, and average, minimum, and maximum native image resolutions for each dataset. We follow the standard evaluation protocols and report top-1 accuracy on the test split, except for ImageNet-1k, for which we report validation-set accuracy.

\begin{table}[h!]
    \caption{\textbf{Image classification datasets.} Summary table of the image classification datasets.}
    \label{tab:image-classification-datasets}
    \centering
    \resizebox{\columnwidth}{!}{
        \begin{tabular}{lcccccc}
            \toprule
            Dataset & Classes & \makecell{Avg. Native\\Resolution} & \makecell{Min Native\\Resolution} & \makecell{Max Native\\Resolution} & \makecell{Training\\Samples} & \makecell{Test\\Samples} \\
            \midrule
            DTD Textures~\citep{cimpoi2014describing} & 47 & $500\times453$ & $300\times231$ & $800\times701$ & 1880 & 1880 \\
            FGVC Aircraft~\citep{maji13fine-grained} & 100 & $1104\times751$ & $775\times430$ & $1600\times1188$ & 3334 & 3333 \\
            EuroSAT~\citep{helber2019eurosat} & 10 & $64\times64$ & $64\times64$ & $64\times64$ & 21600 & 5400 \\
            CIFAR 10~\citep{krizhevsky2009learning} & 10 & $32\times32$ & $32\times32$ & $32\times32$ & 50000 & 10000 \\
            CIFAR 100~\citep{krizhevsky2009learning} & 100 & $32\times32$ & $32\times32$ & $32\times32$ & 50000 & 10000 \\
            Oxford-IIIT Pets~\citep{parkhi2012cats} & 37 & $431\times383$ & $114\times108$ & $3264\times2606$ & 3680 & 3669 \\
            RESISC 45~\citep{resics45} & 45 & $256\times256$ & $256\times256$ & $256\times256$ & 18900 & 6300 \\
            Flowers 102~\citep{flowers102} & 102 & $624\times538$ & $500\times500$ & $919\times993$ & 1020 & 6149 \\
            Food 101~\citep{bossard14} & 101 & $496\times476$ & $193\times122$ & $512\times512$ & 75750 & 25250 \\
            ImageNet-1k~\citep{deng2009imagenet} & 1000 & $473\times406$ & $8\times10$ & $9331\times6530$ & 1.28M & 50000 \\
            \bottomrule
        \end{tabular}
    }
\end{table}

\noindent\textbf{Semantic segmentation.} We evaluate patch-level representations on the validation split of Pascal VOC 2012~\citep{everingham2010pascal}, reporting the mIoU.

\noindent\textbf{Text embedding and retrieval.} We evaluate text representations using the official implementation of MTEB(eng, v2)~\citep{muennighoff2023mteb, enevoldsen2025mmteb}, which covers 41 English-language tasks across classification, clustering, pair classification, retrieval, reranking, semantic textual similarity, and summarization. We report the mean across tasks. For elastic retrieval with asymmetric document and query encoders, we report mean nDCG@10 across the 15 BEIR datasets~\citep{thakur2021beir}.

\subsection{Additional Experiments}
\label{sec:appendix-additional-experiments}

\subsubsection{How does ranking optimization change allocations?}
\label{sec:allocation-analysis}

\begin{figure}[t]
  \centering
  \includegraphics[width=\linewidth]{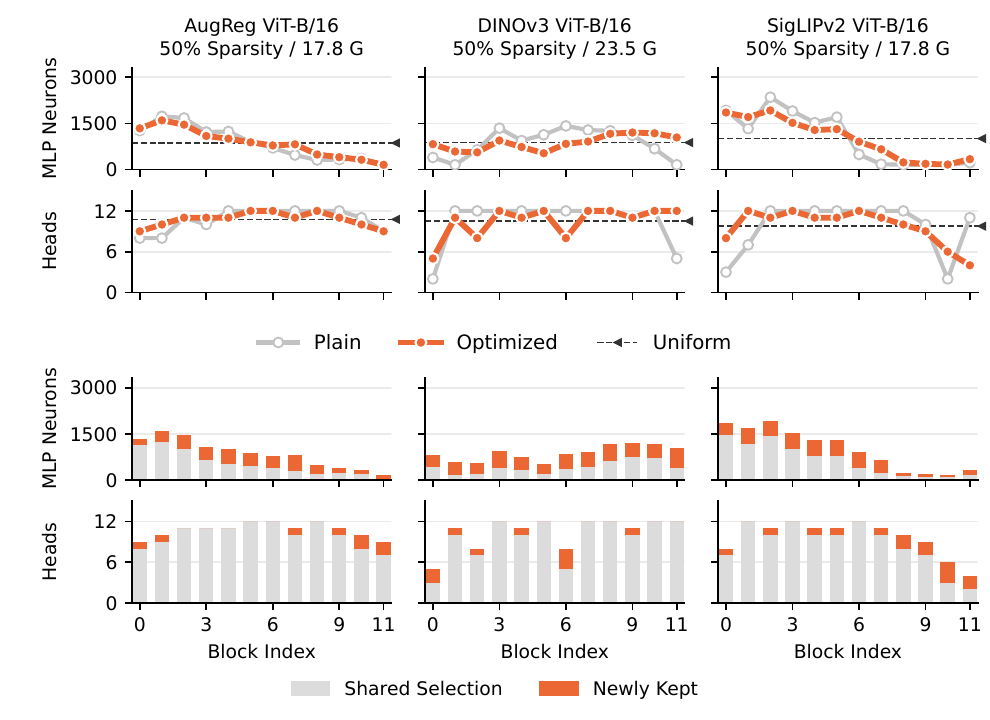}
  \caption{\textbf{Ranking optimization changes both cross-block sparsity allocation and within-block selection.} Retained MLP neurons and attention heads at 50\% sparsity for AugReg, DINOv3, and SigLIPv2 ViT-B/16. The top rows compare per-block allocations from plain and optimized scores with uniform pruning. The bottom rows use the optimized per-block allocation and divide its retained units into those also selected by the plain scores and those newly selected after ranking optimization. Newly retained units appear even where block-level allocations remain nearly unchanged, showing that ranking optimization also reorders units within blocks rather than merely redistributing capacity across them.}
  \label{fig:cross-block}
\end{figure}

In this section, we investigate whether ranking optimization only redistributes retained capacity across blocks or also changes which units are retained within each block.~\autoref{fig:cross-block} compares plain and optimized selections for AugReg, DINOv3, and SigLIPv2 ViT-B/16 models at 50\% sparsity. The top rows show the number of retained MLP neurons and attention heads per block. The bottom rows use the optimized per-block allocation and divide its retained units into those also selected by the plain scores and those newly selected after ranking optimization.

The cross-block effect is backbone-specific. AugReg's allocation changes only modestly, while DINOv3 undergoes a larger redistribution that restores capacity to the final MLP block. For SigLIPv2, optimization preserves the aggressive pruning of later MLP blocks. Across all three backbones, the optimized ranking selects a different subset of units even in blocks where the number retained changes little. This within-block reordering is most visible for MLP neurons. Ranking optimization therefore changes both the allocation of capacity across blocks and the ordering of units within blocks.

\subsubsection{Additional Backbones and Model Scaling}
\label{sec:appendix-additional-backbones}

\begin{figure}[!htbp]
  \centering
  \includegraphics[width=\linewidth]{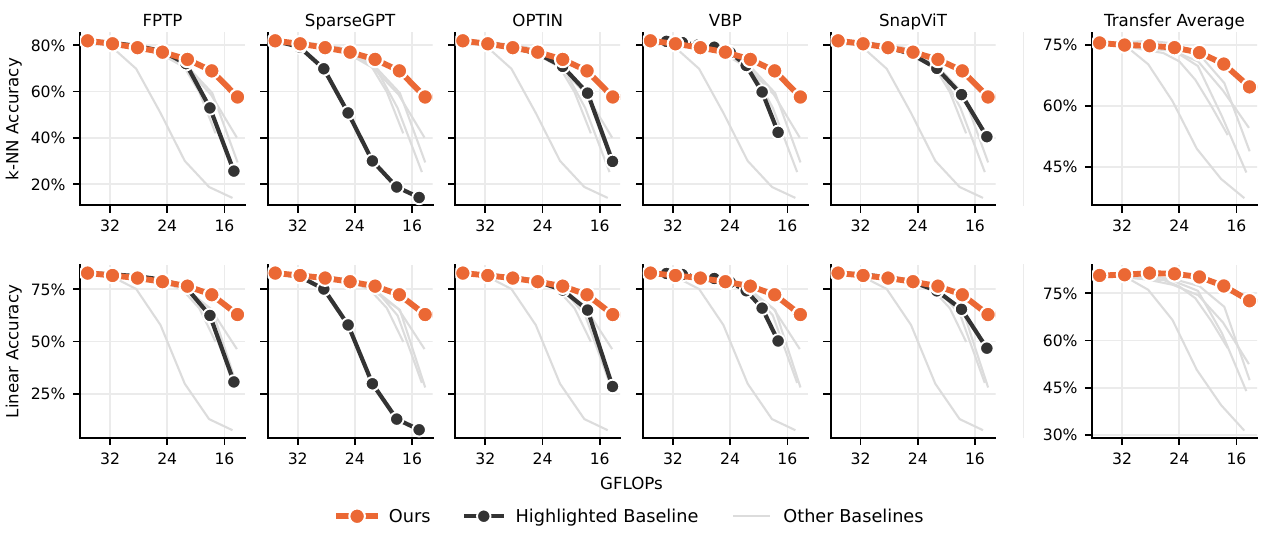}
  \caption{\textbf{Our method retains higher accuracy under deep pruning on DeiT-B/16.} Top-1 k-NN (top) and linear-probing (bottom) accuracy versus GFLOPs on ImageNet-1k and the transfer average. Most baselines are competitive under shallow pruning, but degrade more sharply than our method as sparsity increases.}
  \label{fig:deit-b}
\end{figure}

\noindent\textbf{Original DeiT.} We additionally evaluate DeiT-B/16~\citep{touvron2021training} via k-NN classification and linear probing on ImageNet-1k and the transfer datasets. As shown in~\autoref{fig:deit-b}, our method degrades more smoothly than the baselines as sparsity increases. FPTP, OPTIN, VBP and SnapViT are competitive under shallow pruning, but our method retains higher accuracy under deep pruning in both evaluation protocols.

\begin{figure}[!htbp]
  \centering
  \includegraphics[width=\linewidth]{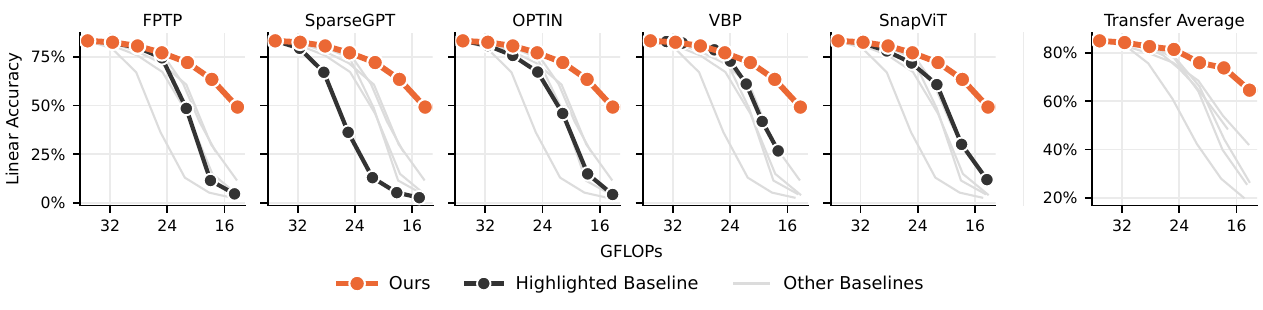}
  \caption{\textbf{Our method degrades more smoothly under linear probing on AugReg ViT-B/16.} Top-1 accuracy in linear classification on ImageNet-1k and transfer datasets for AugReg ViT-B/16 models pruned with our method and the baselines. Compared with the corresponding k-NN results in~\autoref{fig:augreg-scaling}, linear probes retain more accuracy under deep pruning.}
  \label{fig:augreg-linear}
\end{figure}

\noindent\textbf{Linear probing.} We evaluate pruned AugReg ViT-B/16, DINOv3 ViT-H+/16, and SigLIPv2 ViT-G/16 models in linear classification on ImageNet-1k and the transfer datasets. On AugReg, linear probing shows the same qualitative advantage over baselines as k-NN classification, while retaining more accuracy at high sparsity, as shown in~\autoref{fig:augreg-linear}. At 50\% sparsity, DINOv3 ViT-H+/16 retains 84.4\% ImageNet-1k accuracy and 85.4\% on the transfer average, with losses of only 0.8\% and 3.3\% from the dense model while removing 427M parameters. SigLIPv2 ViT-G/16 retains 69.4\% and 73.9\%, with losses of 12.3\% and 17.2\% while removing 573M parameters. At 50\% sparsity, our method leads SnapViT, the runner-up, by 41.2\% and 49.9\% on ImageNet-1k for DINOv3 and SigLIPv2 respectively. These results show that our advantage over baselines also extends to linear classification.

\begin{figure}[!htbp]
  \centering
  \includegraphics[width=\linewidth]{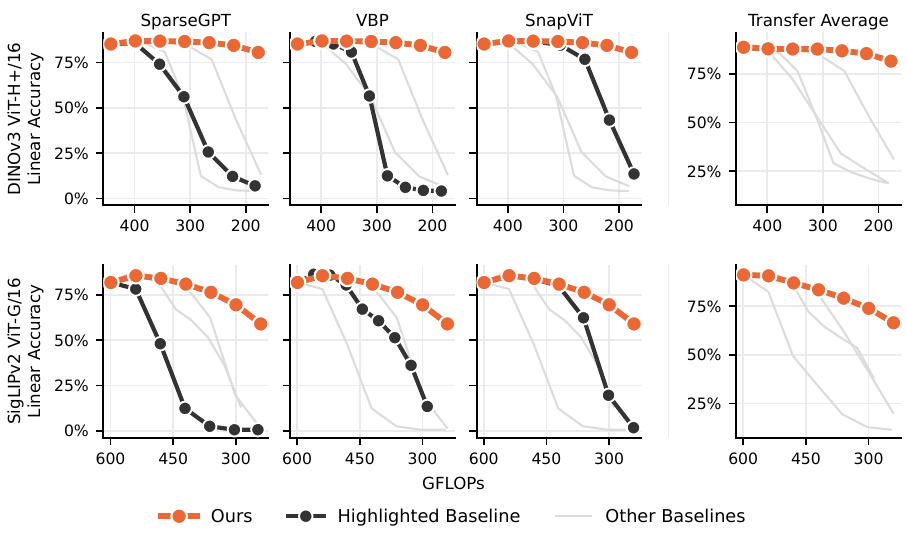}
  \caption{\textbf{Our method outperforms baselines in linear classification under deep pruning on large DINOv3 and SigLIPv2 models.} Top-1 accuracy in linear classification on ImageNet-1k and transfer datasets for DINOv3 ViT-H+/16 (top) and SigLIPv2 ViT-G/16 (bottom), comparing our method with compatible baselines. Notably, the linear accuracy on ImageNet-1k for DINOv3 is essentially flat at up to 50\% sparsity.}
  \label{fig:dinov3-siglip-linear}
\end{figure}

\begin{figure}[t]
  \centering
  \includegraphics[width=\linewidth]{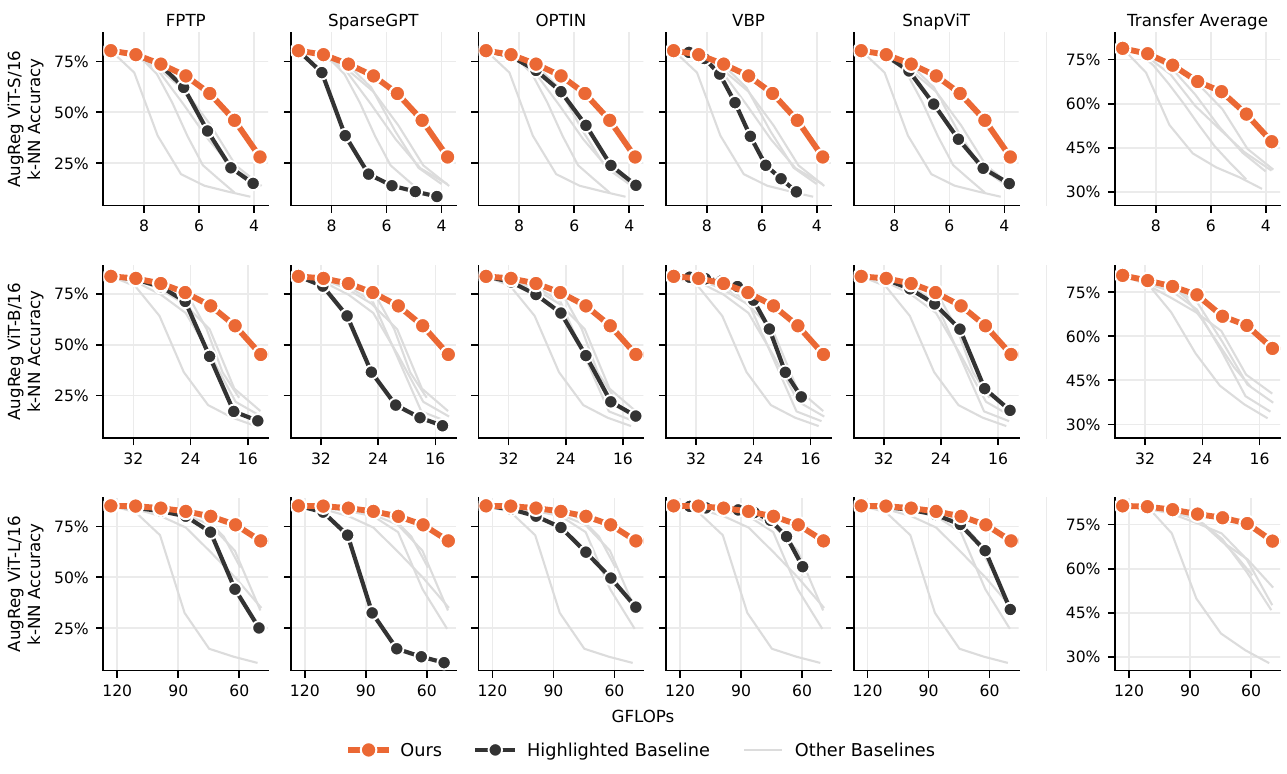}
  \caption{\textbf{Larger AugReg models tolerate more pruning.} Top-1 k-NN accuracy on ImageNet-1k and transfer datasets for AugReg models pruned with our method and the baselines.}
  \label{fig:augreg-scaling}
\end{figure}

\begin{figure}[t]
  \centering
  \includegraphics[width=\linewidth]{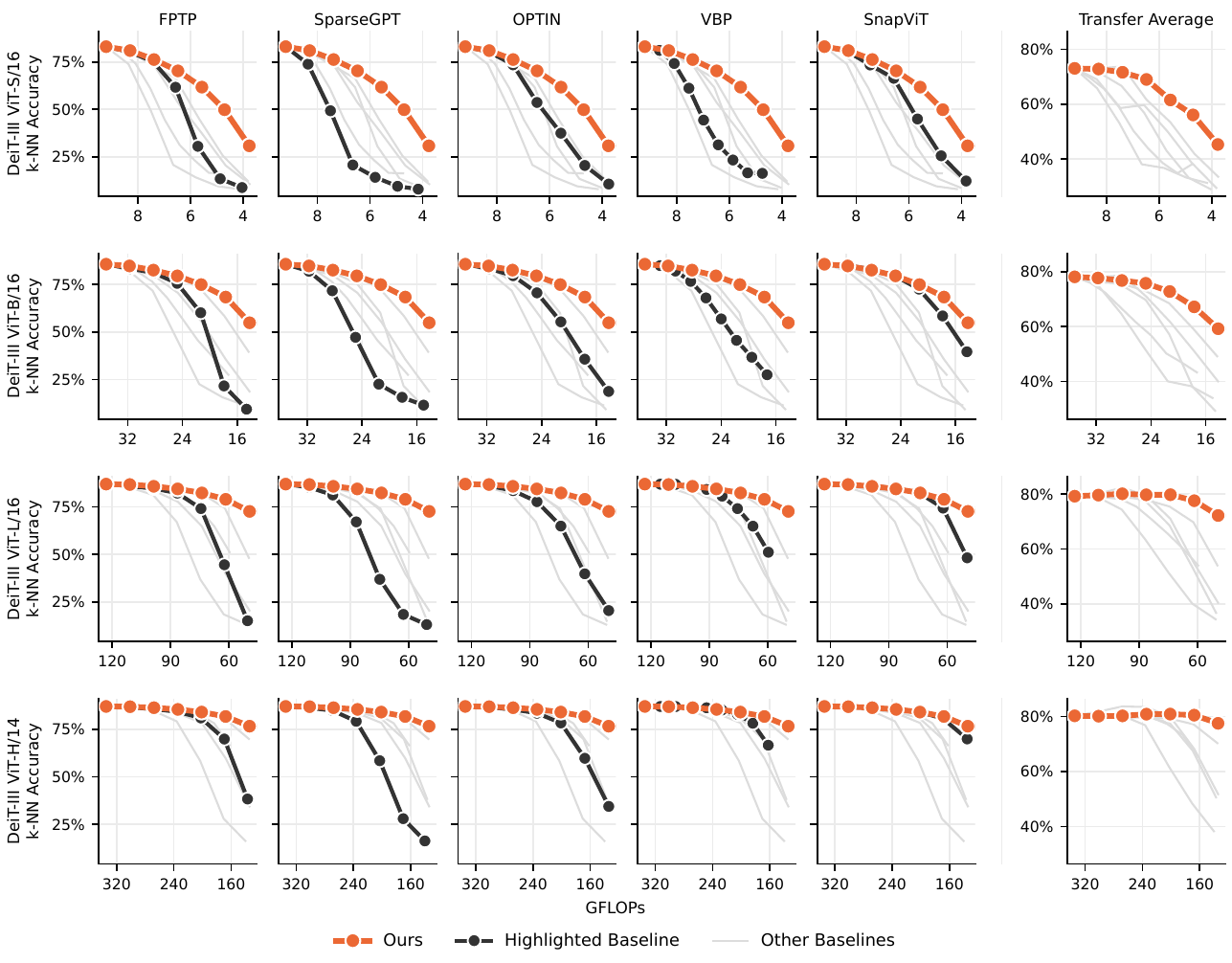}
  \caption{\textbf{Larger DeiT-III models tolerate more pruning.} Top-1 k-NN accuracy on ImageNet-1k and transfer datasets for DeiT-III models pruned with our method and the baselines. At 60\% sparsity, ViT-H/14 loses 10.4\% on ImageNet-1k and 2.73\% on the transfer average, while removing approximately 378M parameters.}
  \label{fig:deit3-scaling}
\end{figure}

\begin{figure}[t]
  \centering
  \includegraphics[width=\linewidth]{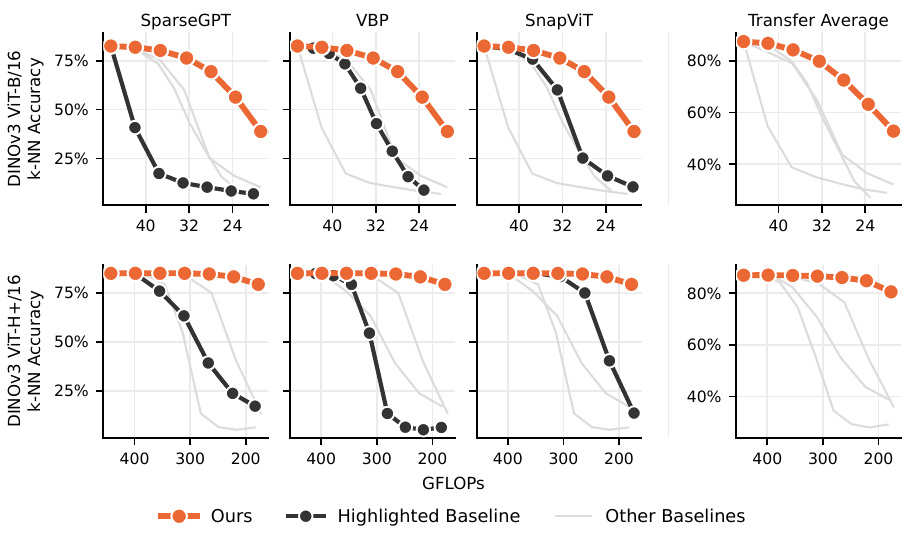}
  \caption{\textbf{Larger DINOv3 models tolerate more pruning.} Top-1 k-NN accuracy on ImageNet-1k and the transfer average for DINOv3 models pruned with our method and the baselines. Our method yields smoother degradation on both models. Pruning the ViT-H+/16 to 60\% sparsity removes 512M parameters, while reducing k-NN accuracy by 5.6\% on ImageNet-1k and 6.4\% on the transfer average.}
  \label{fig:dinov3-scaling}
\end{figure}

\begin{figure}[t]
  \centering
  \includegraphics[width=\linewidth]{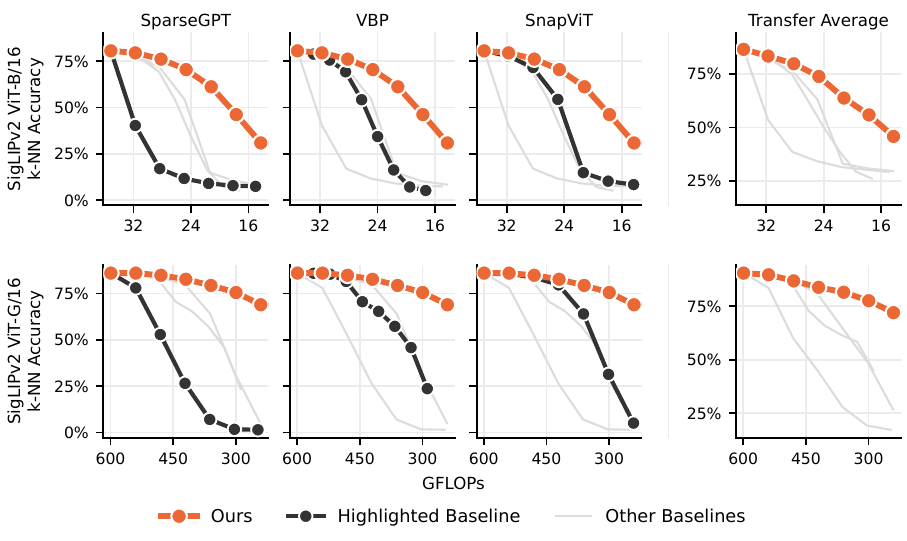}
  \caption{\textbf{Larger SigLIPv2 models tolerate more pruning.} Top-1 k-NN accuracy on ImageNet-1k and the transfer average for SigLIPv2 ViT-B/16 and ViT-G/16 models pruned with our method and the baselines. Our method yields smoother degradation on both models. Pruning the ViT-G/16 to 50\% sparsity removes 573M parameters, while reducing k-NN accuracy by 10.5 percentage points on ImageNet-1k and 12.9 percentage points on the transfer average.}
  \label{fig:siglip-scaling}
\end{figure}

\begin{figure}[t]
  \centering
  \includegraphics[width=\linewidth]{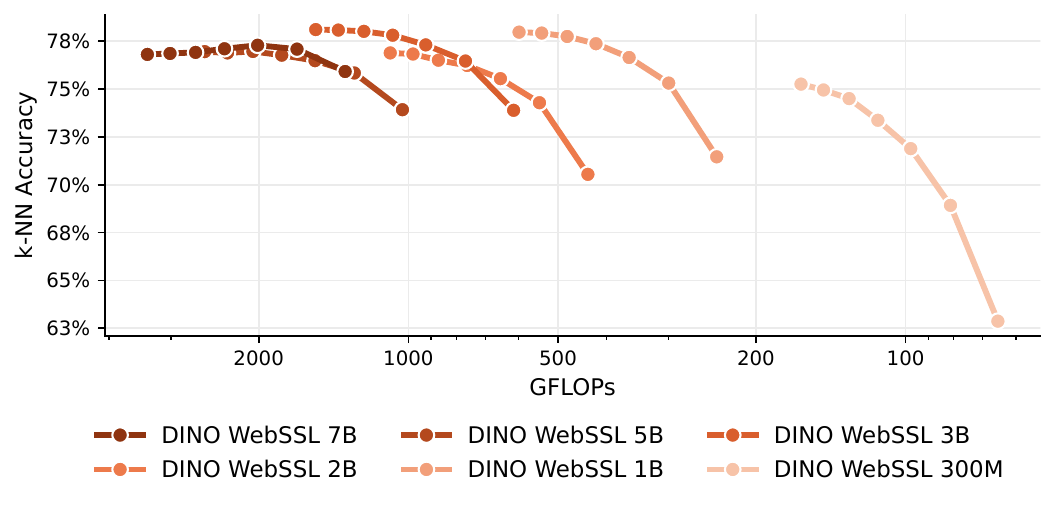}
  \caption{\textbf{Pruned larger models can outperform smaller dense models.} ImageNet-1k top-1 k-NN accuracy versus GFLOPs for the six-model DINO WebSSL family, pruned with our method from dense to 60\% sparsity. Pruned 3B and 7B models pareto-dominate the dense 2B and 5B anchors.}
  \label{fig:dino-webssl-pareto}
\end{figure}

\noindent\textbf{Model scaling.} We provide the full performance versus compute curves discussed in Section~\ref{sec:scaling} for AugReg in~\autoref{fig:augreg-scaling}, DeiT-III in~\autoref{fig:deit3-scaling}, DINOv3 in~\autoref{fig:dinov3-scaling}, and DINO WebSSL in~\autoref{fig:dino-webssl-pareto}. We additionally compare SigLIPv2 ViT-B/16 and ViT-G/16 in~\autoref{fig:siglip-scaling}.

\subsubsection{Transfer Datasets}

\begin{figure}[t]
  \centering
  \includegraphics[width=\linewidth]{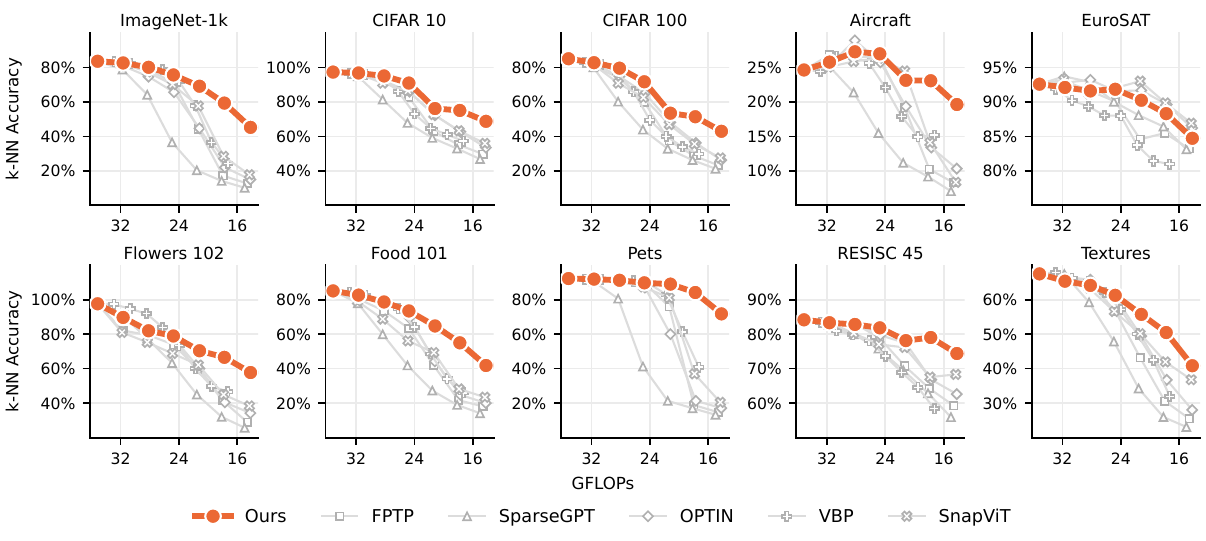}
  \caption{\textbf{AugReg ViT-B/16 degrades unevenly across tasks, with fine-grained tasks being the most sensitive.} Top-1 k-NN accuracy versus GFLOPs curves on ImageNet-1k and the nine transfer datasets for AugReg ViT-B/16 pruned from 10\% to 60\% sparsity with our method and the baselines. At 50\% sparsity, our method loses 24.3\% on ImageNet-1k and over 30\% on CIFAR-100, Flowers 102, and Food 101, but at most 8.1\% on Pets, EuroSAT, and RESISC45. Our method leads the baselines under moderate and deep pruning, except on EuroSAT, where SnapViT and OPTIN retain higher accuracy.}
  \label{fig:augreg-b-transfer-individual-tasks}
\end{figure}

\begin{figure}[t]
  \centering
  \includegraphics[width=\linewidth]{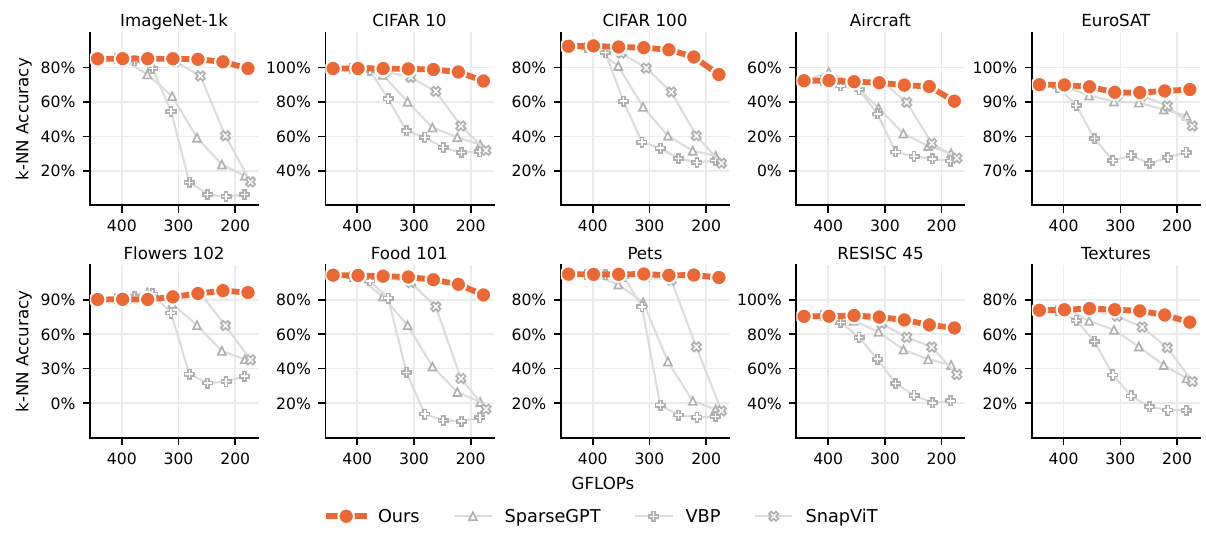}
  \caption{\textbf{DINOv3 ViT-H+/16 is nearly lossless at 50\% sparsity, and improves on Flowers 102.} Top-1 k-NN accuracy versus GFLOPs curves on ImageNet-1k and the nine transfer datasets for DINOv3 ViT-H+/16 pruned from 10\% to 60\% sparsity with our method and the baselines. At 50\% sparsity, our method loses 1.8\% on ImageNet-1k, and between 0.5\% on Pets and 6.2\% on CIFAR-100 on the transfer datasets, while Flowers 102 improves from 90.2\% to 97.9\%. At the same sparsity, it leads SnapViT, the runner-up, by 42.9\% on ImageNet-1k, and from 40\% onward it matches or exceeds every baseline on every dataset.}
  \label{fig:dino-h-transfer-individual-tasks}
\end{figure}

\begin{figure}[t]
  \centering
  \includegraphics[width=\linewidth]{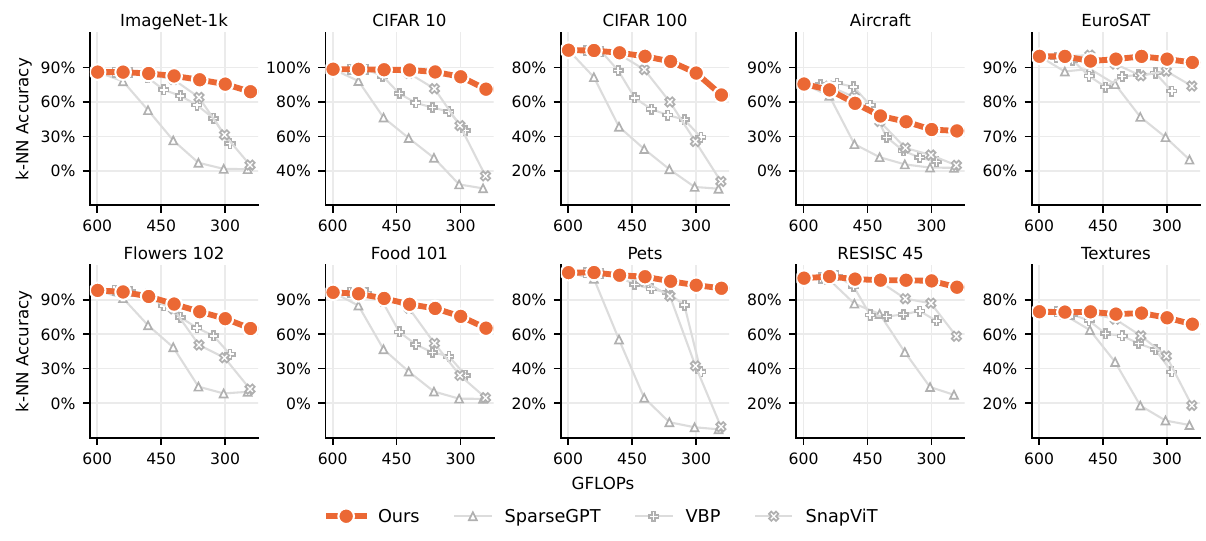}
  \caption{\textbf{SigLIPv2 ViT-G/16 degrades unevenly at 50\% sparsity, with fine-grained tasks being the most sensitive.} Top-1 k-NN accuracy versus GFLOPs curves on ImageNet-1k and the nine transfer datasets for SigLIPv2 ViT-G/16 pruned from 10\% to 60\% sparsity with our method and the baselines. At 50\% sparsity, our method loses 10.5\% on ImageNet-1k, and between 0.8\% on EuroSAT and 39.6\% on Aircraft on the transfer datasets. Under shallow pruning, VBP and SnapViT lead on Aircraft, by 12.9\% and 5.4\% at 20\% sparsity, but from 30\% onward our method leads every baseline on every dataset.}
  \label{fig:siglip-g-transfer-individual-tasks}
\end{figure}

In~\autoref{fig:augreg-b-transfer-individual-tasks}, we plot the top-1 k-NN accuracy of AugReg ViT-B/16 models pruned with our method and the baselines for ImageNet-1k and all transfer datasets. While our method consistently matches or outperforms the baselines except on EuroSAT, the results show significantly different behavior across tasks: performance on Pets is approximately constant, with a dip only at 60\% sparsity, while fine-grained tasks such as Food 101, CIFAR-100, and Flowers 102 degrade more sharply. Interestingly, performance on Aircraft is approximately constant at every sparsity target, but this is most likely explained by the poor performance of the dense model. Furthermore, the good performance on Pets might be an artifact of calibration data alignment, as a significant number of ImageNet-1k classes concern cats and dogs. These results show that our advantage over baselines holds on individual transfer tasks, and that the impact of pruning varies with task granularity and calibration data alignment.

\autoref{fig:dino-h-transfer-individual-tasks} and~\autoref{fig:siglip-g-transfer-individual-tasks} repeat the per-task comparison for DINOv3 ViT-H+/16 and SigLIPv2 ViT-G/16. DINOv3 is remarkably stable: at 50\% sparsity, it stays within 2.0\% of the dense model on ImageNet-1k, CIFAR-10, EuroSAT, and Pets, and, surprisingly, Flowers 102 improves by 7.7\%. Fine-grained tasks degrade only at 60\% sparsity, where CIFAR-100 loses 16.4\% against 5.6\% on ImageNet-1k. SigLIPv2 degrades more and less evenly: at 60\% sparsity, it loses 17.0\% on ImageNet-1k and over 30\% on Aircraft, Flowers 102, and Food 101, while EuroSAT, RESISC45, and Pets lose at most 9.1\%. On both models, baselines are competitive only under shallow pruning, leading on Flowers 102 for DINOv3 and on Aircraft for SigLIPv2, whereas from 40\% and 30\% sparsity, respectively, our method matches or exceeds every baseline on every dataset. Taken together, these results show that our advantage under moderate and deep pruning holds per task on both large models, and that fine-grained tasks, and SigLIPv2 in particular, degrade the most.

\subsubsection{In-Context Semantic Segmentation}
\label{sec:appendix-segmentation}

\begin{figure}[t]
  \centering
  \includegraphics[width=\linewidth]{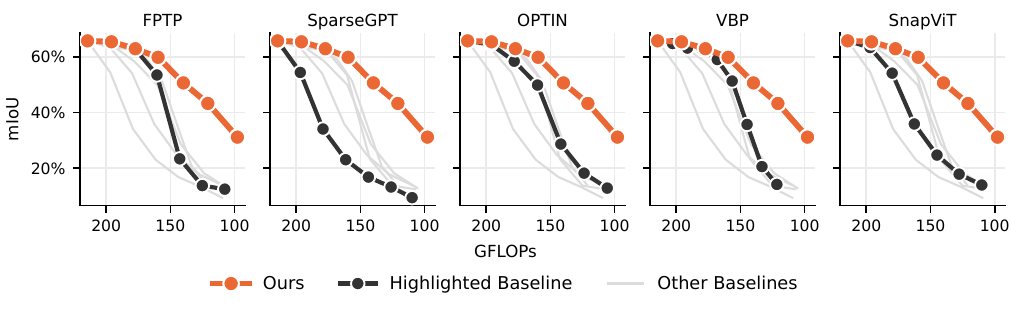}
  \caption{\textbf{Our method preserves patch-level representations.} In-context semantic segmentation mIoU on Pascal VOC 2012 for AugReg ViT-B/16 models pruned with our method and the baselines. Our method retains performance under pruning, improving on the runner-up, OPTIN, by up to 25.1\% at 50\% sparsity.}
  \label{fig:augreg-b-hummingbird}
\end{figure}

\begin{figure}[t]
  \centering
  \includegraphics[width=0.65\linewidth]{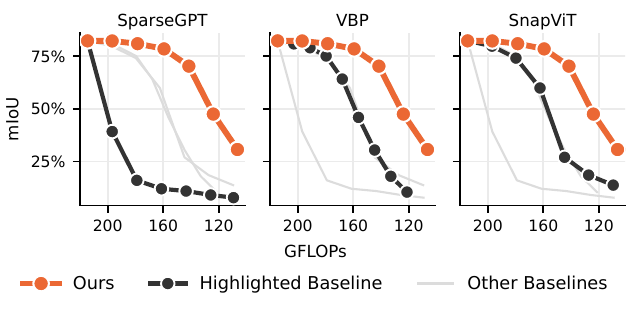}
  \caption{\textbf{Our method preserves DINOv3's patch-level representations under pruning.} In-context semantic segmentation mIoU on Pascal VOC 2012 for DINOv3 ViT-B/16 models pruned with our method and the baselines. Our method degrades smoothly and consistently outperforms each baseline at comparable compute.}
  \label{fig:dinov3-b-hummingbird}
\end{figure}

\autoref{fig:augreg-b-hummingbird} and \autoref{fig:dinov3-b-hummingbird} provide per-baseline breakdowns on Pascal VOC 2012 for AugReg and DINOv3 ViT-B/16. On AugReg, our method outperforms OPTIN, the runner-up, by 25.1\% mIoU at 50\% sparsity. On DINOv3, our method yields a smooth degradation curve and outperforms the baselines across the evaluated compute range. At 40\% sparsity, equivalent to 141.5 GFLOPs, it retains 70.3\% mIoU, outperforming VBP, the runner-up, by 39.8\% despite using 3.1 fewer GFLOPs. Across both backbones, our method preserves patch-level representations more effectively than the evaluated baselines under deep pruning.

\subsubsection{Text Representations}
\label{sec:appendix-text-embeddings}

We provide the full MTEB comparison across all nine text encoders and pruning methods in~\autoref{tab:mteb}.

\begin{table}[!htbp]
  \centering
  \caption{\textbf{Elastic pruning generalizes to text.} MTEB score for dense and elastically pruned models across the BGE, E5, E5-Mistral, and Qwen3 families. Shallow pruning is nearly free: at 20\% sparsity, no model loses more than 3.3 points, and the E5 models lose under 1.0 points. Under deep pruning, bidirectional encoders degrade more gracefully than decoders: at 60\% sparsity, BGE and E5 lose between 5.1 and 10.0 points, while E5-Mistral and Qwen3 lose between 11.5 and 19.5.}
  \label{tab:mteb}
  \resizebox{\columnwidth}{!}{
    \begin{tabular}{llcrrrrrr}
      \toprule
      & & & \multicolumn{6}{c}{Sparsity} \\
      \cmidrule(lr){4-9}
      Backbone & Method & Dense & \multicolumn{1}{c}{10\%} & \multicolumn{1}{c}{20\%} & \multicolumn{1}{c}{30\%} & \multicolumn{1}{c}{40\%} & \multicolumn{1}{c}{50\%} & \multicolumn{1}{c}{60\%} \\
      \midrule
      BGE-S & SnapViT & 64.3 & 63.4$^{-0.9\phantom{0}}$ & \textbf{62.4}$^{-1.9\phantom{0}}$ & \textbf{60.5}$^{-3.8\phantom{0}}$ & 57.6$^{-6.7\phantom{0}}$ & 54.6$^{-9.7\phantom{0}}$ & 51.8$^{-12.5}$ \\
       & SlimGPT &  & 63.2$^{-1.1\phantom{0}}$ & 61.2$^{-3.1\phantom{0}}$ & 58.4$^{-5.9\phantom{0}}$ & 54.7$^{-9.6\phantom{0}}$ & 48.9$^{-15.4}$ & 41.8$^{-22.5}$ \\
       & Ours &  & \textbf{63.7}$^{-0.6\phantom{0}}$ & 62.3$^{-2.0\phantom{0}}$ & 60.3$^{-4.0\phantom{0}}$ & \textbf{58.4}$^{-5.9\phantom{0}}$ & \textbf{56.7}$^{-7.6\phantom{0}}$ & \textbf{54.3}$^{-10.0}$ \\
      \midrule
      BGE-B & SnapViT & 65.2 & \textbf{65.0}$^{-0.2\phantom{0}}$ & \textbf{64.2}$^{-1.0\phantom{0}}$ & \textbf{63.2}$^{-2.0\phantom{0}}$ & 60.1$^{-5.1\phantom{0}}$ & 55.6$^{-9.6\phantom{0}}$ & 49.9$^{-15.3}$ \\
       & SlimGPT &  & 63.8$^{-1.4\phantom{0}}$ & 61.2$^{-4.0\phantom{0}}$ & 57.1$^{-8.1\phantom{0}}$ & 53.4$^{-11.8}$ & 40.9$^{-24.3}$ & 40.6$^{-24.6}$ \\
       & Ours &  & 64.8$^{-0.4\phantom{0}}$ & \textbf{64.2}$^{-1.0\phantom{0}}$ & \textbf{63.2}$^{-2.0\phantom{0}}$ & \textbf{61.8}$^{-3.4\phantom{0}}$ & \textbf{59.5}$^{-5.7\phantom{0}}$ & \textbf{56.4}$^{-8.8\phantom{0}}$ \\
      \midrule
      BGE-L & SnapViT & 65.8 & 65.2$^{-0.6\phantom{0}}$ & 64.2$^{-1.6\phantom{0}}$ & 61.1$^{-4.7\phantom{0}}$ & 54.6$^{-11.2}$ & 48.8$^{-17.0}$ & 43.8$^{-22.0}$ \\
       & SlimGPT &  & 59.1$^{-6.7\phantom{0}}$ & 51.8$^{-14.0}$ & 50.1$^{-15.7}$ & 49.6$^{-16.2}$ & 47.1$^{-18.7}$ & 40.0$^{-25.8}$ \\
       & Ours &  & \textbf{65.5}$^{-0.3\phantom{0}}$ & \textbf{64.8}$^{-1.0\phantom{0}}$ & \textbf{63.6}$^{-2.2\phantom{0}}$ & \textbf{61.2}$^{-4.6\phantom{0}}$ & \textbf{59.5}$^{-6.3\phantom{0}}$ & \textbf{57.1}$^{-8.7\phantom{0}}$ \\
      \midrule
      E5-B & SnapViT & 62.5 & 61.9$^{-0.6\phantom{0}}$ & 60.5$^{-2.0\phantom{0}}$ & 57.0$^{-5.5\phantom{0}}$ & 49.8$^{-12.7}$ & 46.2$^{-16.3}$ & 42.8$^{-19.7}$ \\
       & SlimGPT &  & \textbf{62.4}$^{-0.1\phantom{0}}$ & 59.9$^{-2.6\phantom{0}}$ & 54.1$^{-8.4\phantom{0}}$ & 55.6$^{-6.9\phantom{0}}$ & 51.9$^{-10.6}$ & 48.0$^{-14.5}$ \\
       & Ours &  & 62.1$^{-0.4\phantom{0}}$ & \textbf{62.0}$^{-0.5\phantom{0}}$ & \textbf{60.8}$^{-1.7\phantom{0}}$ & \textbf{59.8}$^{-2.7\phantom{0}}$ & \textbf{57.7}$^{-4.8\phantom{0}}$ & \textbf{55.6}$^{-6.9\phantom{0}}$ \\
      \midrule
      E5-L & SnapViT & 62.4 & 61.9$^{-0.5\phantom{0}}$ & 56.9$^{-5.5\phantom{0}}$ & 47.2$^{-15.2}$ & 39.8$^{-22.6}$ & 37.5$^{-24.9}$ & 30.2$^{-32.2}$ \\
       & SlimGPT &  & 54.7$^{-7.7\phantom{0}}$ & 55.6$^{-6.8\phantom{0}}$ & 54.6$^{-7.8\phantom{0}}$ & 52.0$^{-10.4}$ & 46.1$^{-16.3}$ & 43.5$^{-18.9}$ \\
       & Ours &  & \textbf{62.4}$^{+0.0\phantom{0}}$ & \textbf{61.7}$^{-0.7\phantom{0}}$ & \textbf{61.1}$^{-1.3\phantom{0}}$ & \textbf{60.3}$^{-2.1\phantom{0}}$ & \textbf{59.1}$^{-3.3\phantom{0}}$ & \textbf{57.3}$^{-5.1\phantom{0}}$ \\
      \midrule
      E5-Mistral 7B & SnapViT & 68.3 & 67.8$^{-0.5\phantom{0}}$ & 67.0$^{-1.3\phantom{0}}$ & \textbf{66.4}$^{-1.9\phantom{0}}$ & \textbf{63.9}$^{-4.4\phantom{0}}$ & 56.2$^{-12.1}$ & 43.3$^{-25.0}$ \\
       & SlimGPT &  & 67.5$^{-0.8\phantom{0}}$ & 65.6$^{-2.7\phantom{0}}$ & 62.1$^{-6.2\phantom{0}}$ & 56.3$^{-12.0}$ & 48.9$^{-19.4}$ & 36.7$^{-31.6}$ \\
       & Ours &  & \textbf{68.1}$^{-0.2\phantom{0}}$ & \textbf{67.3}$^{-1.0\phantom{0}}$ & 65.6$^{-2.7\phantom{0}}$ & 62.7$^{-5.6\phantom{0}}$ & \textbf{59.4}$^{-8.9\phantom{0}}$ & \textbf{56.8}$^{-11.5}$ \\
      \midrule
      Qwen3 0.6B & SnapViT & 70.1 & \textbf{69.6}$^{-0.5\phantom{0}}$ & 66.7$^{-3.4\phantom{0}}$ & \textbf{62.9}$^{-7.2\phantom{0}}$ & 57.6$^{-12.5}$ & 47.3$^{-22.8}$ & 39.7$^{-30.4}$ \\
       & SlimGPT &  & 50.9$^{-19.2}$ & 37.4$^{-32.7}$ & 21.0$^{-49.1}$ & 18.3$^{-51.8}$ & 14.1$^{-56.0}$ & 14.1$^{-56.0}$ \\
       & Ours &  & 69.5$^{-0.6\phantom{0}}$ & \textbf{66.8}$^{-3.3\phantom{0}}$ & 61.9$^{-8.2\phantom{0}}$ & \textbf{58.2}$^{-11.9}$ & \textbf{54.4}$^{-15.7}$ & \textbf{52.1}$^{-18.0}$ \\
      \midrule
      Qwen3 4B & SnapViT & 74.1 & \textbf{72.6}$^{-1.5\phantom{0}}$ & \textbf{71.5}$^{-2.6\phantom{0}}$ & \textbf{69.2}$^{-4.9\phantom{0}}$ & \textbf{64.4}$^{-9.7\phantom{0}}$ & 57.2$^{-16.9}$ & 46.3$^{-27.8}$ \\
       & SlimGPT &  & 68.7$^{-5.4\phantom{0}}$ & 59.8$^{-14.3}$ & 44.4$^{-29.7}$ & 34.0$^{-40.1}$ & 22.2$^{-51.9}$ & 14.6$^{-59.5}$ \\
       & Ours &  & \textbf{72.6}$^{-1.5\phantom{0}}$ & 70.8$^{-3.3\phantom{0}}$ & 68.0$^{-6.1\phantom{0}}$ & 63.6$^{-10.5}$ & \textbf{58.6}$^{-15.5}$ & \textbf{54.6}$^{-19.5}$ \\
      \midrule
      Qwen3 8B & SnapViT & 74.6 & 73.9$^{-0.7\phantom{0}}$ & \textbf{72.6}$^{-2.0\phantom{0}}$ & \textbf{70.7}$^{-3.9\phantom{0}}$ & 67.0$^{-7.6\phantom{0}}$ & 58.6$^{-16.0}$ & 42.5$^{-32.1}$ \\
       & SlimGPT &  & 71.8$^{-2.8\phantom{0}}$ & 67.4$^{-7.2\phantom{0}}$ & 61.2$^{-13.4}$ & 50.3$^{-24.3}$ & 39.4$^{-35.2}$ & 26.1$^{-48.5}$ \\
       & Ours &  & \textbf{74.2}$^{-0.4\phantom{0}}$ & \textbf{72.6}$^{-2.0\phantom{0}}$ & 70.3$^{-4.3\phantom{0}}$ & \textbf{67.3}$^{-7.3\phantom{0}}$ & \textbf{63.8}$^{-10.8}$ & \textbf{58.9}$^{-15.7}$ \\
      \bottomrule
    \end{tabular}
  }
\end{table}

Tables~\ref{tab:mteb-no-retrieval} and~\ref{tab:mteb-retrieval} isolate non-retrieval and retrieval performance across the evaluated models. Retrieval is more sensitive to pruning, especially at high sparsity: at 60\% sparsity, BGE-L loses 19.6 points in retrieval nDCG@10, compared with 5.3 points in its non-retrieval score. SnapViT remains competitive under shallow pruning, leading our method on individual encoders by at most 2.6 points, but our method achieves the highest nDCG@10 on all but E5-Mistral 7B at 40\% sparsity and on every encoder at 50\% and 60\%, with margins growing under deeper pruning. For example, at 30\% sparsity, E5-L retains 47.3 nDCG@10 with our method, compared with 17.6 for SnapViT and 32.8 for SlimGPT. Unlike vision transformers, larger text encoders do not consistently retain more of their dense performance under pruning. At 60\% sparsity, Qwen3 4B loses more than 0.6B on both retrieval and non-retrieval tasks, with drops of 26.2 versus 22.2 points and 17.3 versus 16.7 points, respectively. Scaling benefits also depend on the task group: E5-L loses substantially less than E5-B on retrieval at the same sparsity, with drops of 8.0 versus 15.5 points, but their non-retrieval drops are similar at 4.3 and 4.1 points. Our method thus reduces retrieval degradation relative to the baselines, but substantial losses remain at high sparsity.

\begin{figure}[t]
  \centering
  \includegraphics[width=\linewidth]{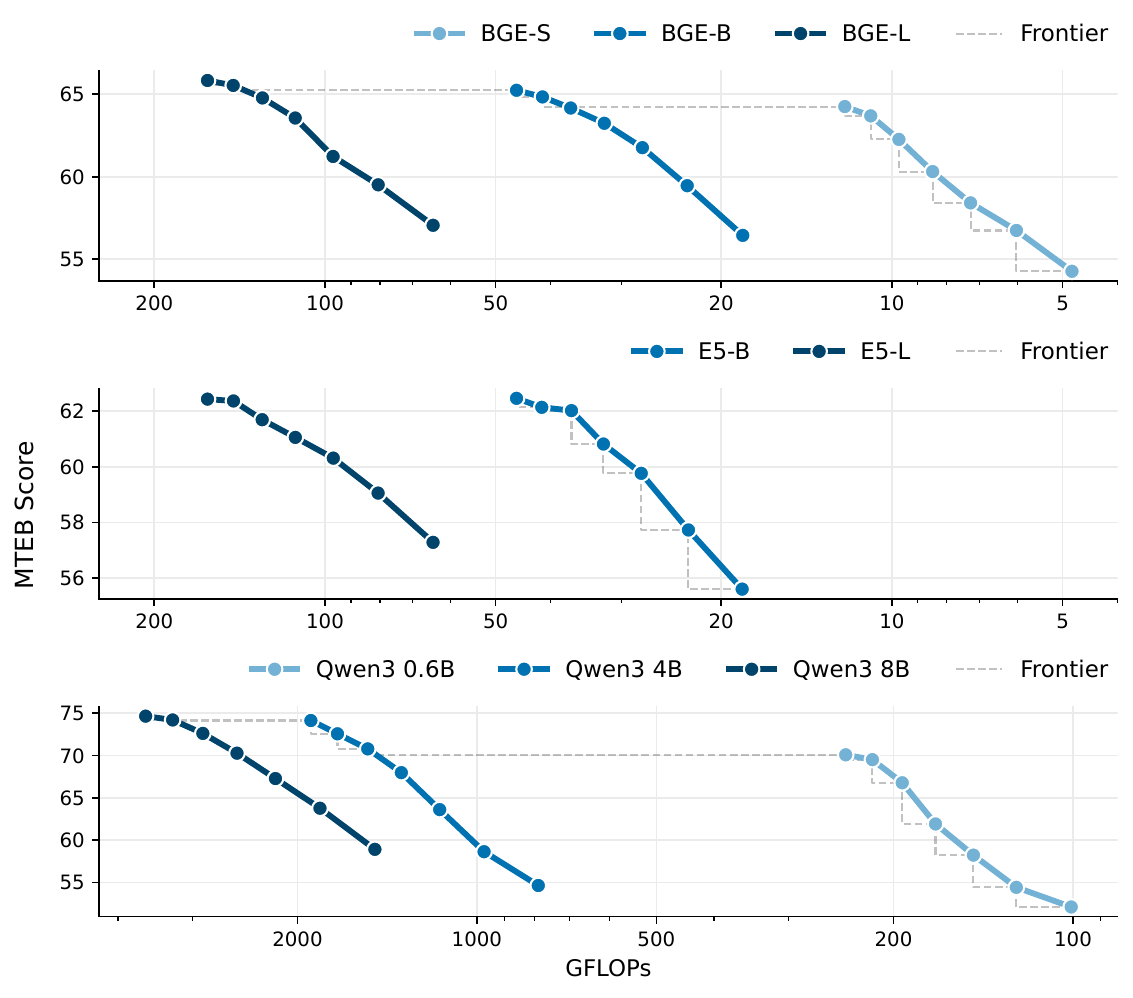}
  \caption{\textbf{Elastic pruning extends the performance-compute frontier of dense text encoders.} MTEB score versus GFLOPs for BGE, E5, and Qwen3 encoders, from dense to 60\% sparsity. Dashed lines indicate the pareto frontier. GFLOPs are shown on a logarithmic scale.}
  \label{fig:pareto}
\end{figure}

\autoref{fig:pareto} shows the performance-compute pareto frontier for each model family. Dense scaling brings little improvement for BGE and E5: BGE scores increase from 64.3 for BGE-S to 65.8 for BGE-L, while E5-B and E5-L score similarly at 62.5 and 62.4. Scale benefits Qwen3 more, with 4B scoring 74.1 versus 70.1 for 0.6B. Yet, further scaling to 8B only yields 74.6. The performance margin available for pruning therefore varies across model sizes and families. Nonetheless, elastic pruning extends the frontier beyond dense models: BGE-B and BGE-L remain on the frontier up to 10\% sparsity, while Qwen3 8B and 4B are on the frontier at up to 10\% and 20\% sparsity, respectively. Pruning the smallest model in a family extends the frontier toward lower compute, reaching below 5 GFLOPs for BGE-S without collapse. With similar dense scores, E5-L is always pareto-dominated by dense E5-B.

\noindent\textbf{Comparison with EffiR.} We compare our pruned E5-Mistral 7B models with EffiR~\citep{lei2026making} on its reported BEIR-13 subset, using the same model to encode both documents and queries. We consider two sets of baselines: E5-Mistral 7B models pruned by coarse layer dropping without retraining, and Mistral 7B models pruned by layer dropping or EffiR's coarse-to-fine strategy and then fine-tuned for retrieval. As shown in~\autoref{tab:effir-drop}, our method achieves higher mean nDCG@10 than the pruning-only baselines in every comparison, reaching 49.9 versus 24.7 at 4.3B parameters. Against the fine-tuned models, \autoref{tab:effir-tuned} shows that our method achieves higher mean nDCG@10 down to 5.7B parameters. At 4.3B parameters, it trails by 5.4 points. Pruning a strong pre-trained retriever thus yields competitive retrieval performance at moderate sparsity without requiring any weight updates.

\begin{table}[t]
  \centering
  \caption{\textbf{Our elastic pruning preserves retrieval performance better than coarse layer dropping.} nDCG@10 on 13 BEIR datasets, with queries and documents encoded by the same pruned model. We compare our elastically pruned E5-Mistral 7B models against similarly sized layer-dropping baselines from EffiR, both without retraining. A$k$, B$k$, and M$k$ denote dropping $k$ attention sublayers, transformer blocks, and MLP sublayers, respectively. Our mean score is higher in every comparison. At 4.3B parameters, our method achieves 49.9 versus 24.7 for dropping 16 MLP sublayers, a gain of 25.2 points.}
  \label{tab:effir-drop}
  \resizebox{\columnwidth}{!}{
    \begin{threeparttable}
      \begin{tabular}{llr*{6}{r}}
        \toprule
        Dataset $\downarrow$ & Method $\downarrow$ & Dense $\downarrow$ & \multicolumn{2}{l}{Attention} & \multicolumn{2}{l}{Blocks} & \multicolumn{2}{l}{MLP} \\
        \cmidrule(lr){4-5}\cmidrule(lr){6-7}\cmidrule(lr){8-9}
        \multicolumn{3}{l}{EffiR Variant $\rightarrow$} & A8 & A16 & B8 & B16 & M8 & M16 \\
        \multicolumn{3}{l}{Params, EffiR/Ours (B) $\rightarrow$} & 6.8/6.4 & 6.4/6.4 & 5.4/5.7 & 3.6/3.6 & 5.7/5.7 & 4.3/4.3 \\
        \midrule
        ArguAna & EffiR$^\dagger$ & 61.6 & 55.3 & 0.1 & 39.0 & 9.3 & 59.0 & 8.4 \\
        & Ours & 61.7 & \textbf{59.9} & \textbf{59.9} & \textbf{60.1} & \textbf{53.0} & \textbf{60.1} & \textbf{57.1} \\
        \midrule
        Climate & EffiR$^\dagger$ & 37.5 & 32.8 & 2.2 & 1.2 & 3.9 & \textbf{39.0} & 9.8 \\
        FEVER & Ours & 38.4 & \textbf{38.4} & \textbf{38.4} & \textbf{36.1} & \textbf{21.4} & 36.1 & \textbf{28.2} \\
        \midrule
        DBPedia & EffiR$^\dagger$ & 48.6 & 44.8 & 0.6 & 19.0 & 9.7 & 44.7 & 15.2 \\
        & Ours & 48.9 & \textbf{48.8} & \textbf{48.8} & \textbf{47.6} & \textbf{29.9} & \textbf{47.6} & \textbf{39.5} \\
        \midrule
        FEVER & EffiR$^\dagger$ & 88.3 & 83.8 & 9.5 & 36.6 & 22.5 & \textbf{89.6} & 56.6 \\
        & Ours & 87.8 & \textbf{88.0} & \textbf{88.0} & \textbf{87.2} & \textbf{61.6} & 87.2 & \textbf{77.4} \\
        \midrule
        FiQA2018 & EffiR$^\dagger$ & 57.2 & 51.8 & 2.4 & 9.1 & 8.1 & \textbf{53.5} & 14.0 \\
        & Ours & 56.5 & \textbf{54.6} & \textbf{54.6} & \textbf{51.0} & \textbf{32.4} & 51.0 & \textbf{40.9} \\
        \midrule
        HotpotQA & EffiR$^\dagger$ & 75.6 & 67.5 & 2.8 & 43.0 & 9.6 & \textbf{75.5} & 18.0 \\
        & Ours & 75.7 & \textbf{75.5} & \textbf{75.5} & \textbf{72.5} & \textbf{37.2} & 72.5 & \textbf{53.0} \\
        \midrule
        NFCorpus & EffiR$^\dagger$ & 38.7 & 36.1 & 2.6 & 11.4 & 17.5 & 37.4 & 18.2 \\
        & Ours & 38.6 & \textbf{38.7} & \textbf{38.7} & \textbf{38.9} & \textbf{29.8} & \textbf{38.9} & \textbf{33.8} \\
        \midrule
        NQ & EffiR$^\dagger$ & 67.1 & 57.0 & 0.4 & 25.6 & 12.1 & \textbf{66.1} & 26.0 \\
        & Ours & 63.6 & \textbf{62.7} & \textbf{62.7} & \textbf{62.3} & \textbf{35.0} & 62.3 & \textbf{54.3} \\
        \midrule
        Quora & EffiR$^\dagger$ & 89.3 & 87.1 & 0.6 & 21.5 & 26.8 & 89.1 & 66.0 \\
        Retrieval & Ours & 89.6 & \textbf{89.6} & \textbf{89.6} & \textbf{89.6} & \textbf{87.9} & \textbf{89.6} & \textbf{89.1} \\
        \midrule
        SCIDOCS & EffiR$^\dagger$ & 16.8 & 13.0 & 0.1 & 7.3 & 2.2 & 15.2 & 5.1 \\
        & Ours & 16.3 & \textbf{17.4} & \textbf{17.4} & \textbf{16.7} & \textbf{9.7} & \textbf{16.7} & \textbf{13.8} \\
        \midrule
        SciFact & EffiR$^\dagger$ & 76.7 & 72.2 & 1.5 & 33.1 & 20.3 & 74.1 & 30.6 \\
        & Ours & 76.4 & \textbf{76.4} & \textbf{76.4} & \textbf{74.5} & \textbf{64.9} & \textbf{74.5} & \textbf{71.5} \\
        \midrule
        TREC & EffiR$^\dagger$ & 86.6 & 85.2 & 12.6 & 55.3 & 20.0 & 80.8 & 45.7 \\
        COVID & Ours & 87.2 & \textbf{87.1} & \textbf{87.1} & \textbf{84.7} & \textbf{51.0} & \textbf{84.7} & \textbf{67.6} \\
        \midrule
        Touche & EffiR$^\dagger$ & 23.6 & 18.8 & 0.0 & 7.2 & 5.6 & 19.1 & 7.6 \\
        2020 & Ours & 26.4 & \textbf{29.8} & \textbf{29.8} & \textbf{30.1} & \textbf{16.6} & \textbf{30.1} & \textbf{23.1} \\
        \midrule
        Average & EffiR$^\dagger$ & 59.1 & 54.3 & 2.7 & 23.8 & 12.9 & 57.2 & 24.7 \\
        & Ours & 59.0 & \textbf{59.0} & \textbf{59.0} & \textbf{57.8} & \textbf{40.8} & \textbf{57.8} & \textbf{49.9} \\
        \bottomrule
      \end{tabular}    
      \begin{tablenotes}
        \small
        \item[$\dagger$] indicates results taken from the EffiR paper~\citep{lei2026making}. Both dense baselines use E5-Mistral 7B, with score differences reflecting evaluation protocols.
      \end{tablenotes}
    \end{threeparttable}  
  }
\end{table}

\begin{table}[t]
  \centering
  \caption{\textbf{Retraining-free pruning is competitive with retrieval-fine-tuned models at moderate sparsity.} nDCG@10 on 13 BEIR datasets, with queries and documents encoded by the same pruned model. We compare our retraining-free pruning of E5-Mistral 7B against similarly sized EffiR models, obtained by pruning Mistral 7B and fine-tuning on MS MARCO. A$k$ and M$k$ denote dropping $k$ attention and MLP sublayers, respectively, and EffiR denotes the full coarse-to-fine method. Our mean score is higher down to 5.7B parameters. At 4.3B, our method achieves 49.9 versus 55.3 for the fine-tuned baseline, a gap of only 5.4 points without any weight updates. On ArguAna, our method leads across all compared sizes.}
  \label{tab:effir-tuned}
  \resizebox{\columnwidth}{!}{
    \begin{threeparttable}
      \begin{tabular}{llr*{8}{r}}
        \toprule
        Dataset $\downarrow$ & Method $\downarrow$ & Dense $\downarrow$ & \multicolumn{2}{l}{Attention} & \multicolumn{4}{l}{MLP} & \multicolumn{1}{l}{Mixed} & \multicolumn{1}{l}{Full} \\
        \cmidrule(lr){4-5}\cmidrule(lr){6-9}\cmidrule(lr){10-10}\cmidrule(lr){11-11}
        \multicolumn{3}{l}{EffiR Variant $\rightarrow$} & A8 & A16 & M8 & M16 & M20 & M24 & M16+A8 & EffiR \\
        \multicolumn{3}{l}{Params, EffiR/Ours (B) $\rightarrow$} & 6.8/6.4 & 6.4/6.4 & 5.7/5.7 & 4.3/4.3 & 3.6/3.6 & 2.9/3.0 & 4.0/4.3 & 3.4/3.6 \\
        \midrule
        ArguAna & EffiR$^\dagger$ & 58.2 & 56.1 & 47.4 & 56.9 & 54.5 & 51.4 & 45.6 & 53.2 & 52.8 \\
         & Ours & 61.7 & \textbf{59.9} & \textbf{59.9} & \textbf{60.1} & \textbf{57.1} & \textbf{53.0} & \textbf{50.7} & \textbf{57.1} & \textbf{53.0} \\
        \midrule
        Climate & EffiR$^\dagger$ & 30.8 & 29.9 & 29.4 & 30.4 & \textbf{31.6} & \textbf{31.9} & \textbf{28.9} & 27.9 & \textbf{30.4} \\
        FEVER & Ours & 38.4 & \textbf{38.4} & \textbf{38.4} & \textbf{36.1} & 28.2 & 21.4 & 21.1 & \textbf{28.2} & 21.4 \\
        \midrule
        DBPedia & EffiR$^\dagger$ & 44.2 & 40.1 & 42.1 & 43.9 & \textbf{42.2} & \textbf{37.4} & \textbf{41.3} & 39.2 & \textbf{43.0} \\
         & Ours & 48.9 & \textbf{48.8} & \textbf{48.8} & \textbf{47.6} & 39.5 & 29.9 & 20.6 & \textbf{39.5} & 29.9 \\
        \midrule
        FEVER & EffiR$^\dagger$ & 83.5 & 79.1 & 77.4 & 82.6 & \textbf{83.3} & \textbf{79.7} & \textbf{80.6} & \textbf{79.9} & \textbf{82.0} \\
         & Ours & 87.8 & \textbf{88.0} & \textbf{88.0} & \textbf{87.2} & 77.4 & 61.6 & 53.2 & 77.4 & 61.6 \\
        \midrule
        FiQA2018 & EffiR$^\dagger$ & 45.9 & 45.5 & 43.5 & 45.8 & \textbf{44.9} & \textbf{43.1} & \textbf{40.3} & \textbf{43.7} & \textbf{42.8} \\
         & Ours & 56.5 & \textbf{54.6} & \textbf{54.6} & \textbf{51.0} & 40.9 & 32.4 & 25.6 & 40.9 & 32.4 \\
        \midrule
        HotpotQA & EffiR$^\dagger$ & 70.5 & 68.1 & 66.2 & 71.0 & \textbf{70.1} & \textbf{67.9} & \textbf{67.5} & \textbf{67.9} & \textbf{67.6} \\
         & Ours & 75.7 & \textbf{75.5} & \textbf{75.5} & \textbf{72.5} & 53.0 & 37.2 & 33.7 & 53.0 & 37.2 \\
        \midrule
        NFCorpus & EffiR$^\dagger$ & 34.7 & 31.6 & 34.2 & 33.3 & \textbf{35.4} & \textbf{33.1} & \textbf{36.3} & \textbf{33.9} & \textbf{35.9} \\
         & Ours & 38.6 & \textbf{38.7} & \textbf{38.7} & \textbf{38.9} & 33.8 & 29.8 & 27.8 & 33.8 & 29.8 \\
        \midrule
        NQ & EffiR$^\dagger$ & 65.0 & \textbf{64.5} & 62.4 & \textbf{65.4} & \textbf{63.8} & \textbf{60.8} & \textbf{58.4} & \textbf{61.9} & \textbf{60.8} \\
         & Ours & 63.6 & 62.7 & \textbf{62.7} & 62.3 & 54.3 & 35.0 & 22.5 & 54.3 & 35.0 \\
        \midrule
        Quora & EffiR$^\dagger$ & 83.1 & 83.3 & 87.9 & 85.9 & 83.7 & 86.5 & \textbf{87.0} & 84.3 & 83.3 \\
        Retrieval & Ours & 89.6 & \textbf{89.6} & \textbf{89.6} & \textbf{89.6} & \textbf{89.1} & \textbf{87.9} & 86.4 & \textbf{89.1} & \textbf{87.9} \\
        \midrule
        SCIDOCS & EffiR$^\dagger$ & 17.2 & 14.3 & 17.2 & \textbf{17.3} & \textbf{17.1} & \textbf{17.0} & \textbf{17.4} & \textbf{16.5} & \textbf{17.4} \\
         & Ours & 16.3 & \textbf{17.4} & \textbf{17.4} & 16.7 & 13.8 & 9.7 & 4.1 & 13.8 & 9.7 \\
        \midrule
        SciFact & EffiR$^\dagger$ & 75.9 & 73.7 & 76.2 & \textbf{76.2} & \textbf{76.4} & \textbf{75.7} & \textbf{74.3} & \textbf{75.9} & \textbf{75.1} \\
         & Ours & 76.4 & \textbf{76.4} & \textbf{76.4} & 74.5 & 71.5 & 64.9 & 63.0 & 71.5 & 64.9 \\
        \midrule
        TREC & EffiR$^\dagger$ & 84.1 & 83.0 & 86.3 & 83.7 & \textbf{84.2} & \textbf{85.7} & \textbf{83.3} & \textbf{85.3} & \textbf{83.3} \\
        COVID & Ours & 87.2 & \textbf{87.1} & \textbf{87.1} & \textbf{84.7} & 67.6 & 51.0 & 41.2 & 67.6 & 51.0 \\
        \midrule
        Touche & EffiR$^\dagger$ & 35.7 & \textbf{30.7} & 28.4 & \textbf{34.4} & \textbf{32.3} & \textbf{28.1} & \textbf{31.1} & \textbf{28.7} & \textbf{31.7} \\
        2020 & Ours & 26.4 & 29.8 & \textbf{29.8} & 30.1 & 23.1 & 16.6 & 14.7 & 23.1 & 16.6 \\
        \midrule
        Average & EffiR$^\dagger$ & 56.1 & 53.8 & 53.7 & 55.9 & \textbf{55.3} & \textbf{53.7} & \textbf{53.2} & \textbf{53.7} & \textbf{54.3} \\
         & Ours & 59.0 & \textbf{59.0} & \textbf{59.0} & \textbf{57.8} & 49.9 & 40.8 & 35.7 & 49.9 & 40.8 \\
        \bottomrule
      \end{tabular}
      \begin{tablenotes}
        \small
        \item[$\dagger$] indicates results taken from the EffiR paper~\citep{lei2026making}. The dense baselines differ as EffiR uses Mistral 7B fine-tuned on MS MARCO, while we use the pre-trained E5-Mistral 7B retriever.
      \end{tablenotes}
    \end{threeparttable}
  }
\end{table}

\begin{table}[t]
  \centering
  \caption{\textbf{Non-retrieval tasks are robust to pruning.} MTEB score excluding retrieval tasks for dense and elastically pruned BGE, E5, E5-Mistral, and Qwen3 models. At 60\% sparsity, BGE-L and E5-L lose 5.3 and 4.3 points, against 19.6 and 8.0 on retrieval tasks as shown in \autoref{tab:mteb-retrieval}.}
  \label{tab:mteb-no-retrieval}
  \resizebox{\columnwidth}{!}{
    \begin{tabular}{llcrrrrrr}
      \toprule
      & & & \multicolumn{6}{c}{Sparsity} \\
      \cmidrule(lr){4-9}
      Backbone & Method & Dense & \multicolumn{1}{c}{10\%} & \multicolumn{1}{c}{20\%} & \multicolumn{1}{c}{30\%} & \multicolumn{1}{c}{40\%} & \multicolumn{1}{c}{50\%} & \multicolumn{1}{c}{60\%} \\
      \midrule
      BGE-S & SnapViT & 67.6 & 67.2$^{-0.4\phantom{0}}$ & 66.5$^{-1.1\phantom{0}}$ & 65.5$^{-2.1\phantom{0}}$ & 64.0$^{-3.6\phantom{0}}$ & 62.5$^{-5.1\phantom{0}}$ & 60.4$^{-7.2\phantom{0}}$ \\
       & SlimGPT &  & 67.2$^{-0.4\phantom{0}}$ & 65.9$^{-1.7\phantom{0}}$ & 63.8$^{-3.8\phantom{0}}$ & 61.3$^{-6.3\phantom{0}}$ & 57.0$^{-10.6}$ & 50.9$^{-16.7}$ \\
       & Ours &  & \textbf{67.3}$^{-0.3\phantom{0}}$ & \textbf{66.7}$^{-0.9\phantom{0}}$ & \textbf{65.6}$^{-2.0\phantom{0}}$ & \textbf{64.5}$^{-3.1\phantom{0}}$ & \textbf{63.3}$^{-4.3\phantom{0}}$ & \textbf{61.9}$^{-5.7\phantom{0}}$ \\
      \midrule
      BGE-B & SnapViT & 68.6 & \textbf{68.4}$^{-0.2\phantom{0}}$ & \textbf{68.0}$^{-0.6\phantom{0}}$ & \textbf{67.4}$^{-1.2\phantom{0}}$ & 65.9$^{-2.7\phantom{0}}$ & 63.0$^{-5.6\phantom{0}}$ & 59.4$^{-9.2\phantom{0}}$ \\
       & SlimGPT &  & 67.6$^{-1.0\phantom{0}}$ & 65.8$^{-2.8\phantom{0}}$ & 63.6$^{-5.0\phantom{0}}$ & 62.3$^{-6.3\phantom{0}}$ & 53.2$^{-15.4}$ & 52.9$^{-15.7}$ \\
       & Ours &  & \textbf{68.4}$^{-0.2\phantom{0}}$ & \textbf{68.0}$^{-0.6\phantom{0}}$ & \textbf{67.4}$^{-1.2\phantom{0}}$ & \textbf{66.4}$^{-2.2\phantom{0}}$ & \textbf{65.0}$^{-3.6\phantom{0}}$ & \textbf{63.3}$^{-5.3\phantom{0}}$ \\
      \midrule
      BGE-L & SnapViT & 69.2 & 68.8$^{-0.4\phantom{0}}$ & 68.3$^{-0.9\phantom{0}}$ & 67.0$^{-2.2\phantom{0}}$ & 63.8$^{-5.4\phantom{0}}$ & 58.9$^{-10.3}$ & 55.2$^{-14.0}$ \\
       & SlimGPT &  & 65.1$^{-4.1\phantom{0}}$ & 60.3$^{-8.9\phantom{0}}$ & 58.0$^{-11.2}$ & 58.7$^{-10.5}$ & 56.3$^{-12.9}$ & 49.1$^{-20.1}$ \\
       & Ours &  & \textbf{69.1}$^{-0.1\phantom{0}}$ & \textbf{68.7}$^{-0.5\phantom{0}}$ & \textbf{68.1}$^{-1.1\phantom{0}}$ & \textbf{66.6}$^{-2.6\phantom{0}}$ & \textbf{65.8}$^{-3.4\phantom{0}}$ & \textbf{63.9}$^{-5.3\phantom{0}}$ \\
      \midrule
      E5-B & SnapViT & 66.6 & 65.8$^{-0.8\phantom{0}}$ & 64.7$^{-1.9\phantom{0}}$ & 61.9$^{-4.7\phantom{0}}$ & 58.2$^{-8.4\phantom{0}}$ & 56.3$^{-10.3}$ & 53.7$^{-12.9}$ \\
       & SlimGPT &  & \textbf{66.5}$^{-0.1\phantom{0}}$ & 65.0$^{-1.6\phantom{0}}$ & 62.1$^{-4.5\phantom{0}}$ & 62.8$^{-3.8\phantom{0}}$ & 59.9$^{-6.7\phantom{0}}$ & 57.4$^{-9.2\phantom{0}}$ \\
       & Ours &  & 66.2$^{-0.4\phantom{0}}$ & \textbf{65.9}$^{-0.7\phantom{0}}$ & \textbf{65.2}$^{-1.4\phantom{0}}$ & \textbf{64.5}$^{-2.1\phantom{0}}$ & \textbf{63.6}$^{-3.0\phantom{0}}$ & \textbf{62.5}$^{-4.1\phantom{0}}$ \\
      \midrule
      E5-L & SnapViT & 66.7 & 66.3$^{-0.4\phantom{0}}$ & 63.2$^{-3.5\phantom{0}}$ & 56.8$^{-9.9\phantom{0}}$ & 49.7$^{-17.0}$ & 46.7$^{-20.0}$ & 38.5$^{-28.2}$ \\
       & SlimGPT &  & 60.9$^{-5.8\phantom{0}}$ & 62.7$^{-4.0\phantom{0}}$ & 61.7$^{-5.0\phantom{0}}$ & 60.5$^{-6.2\phantom{0}}$ & 55.6$^{-11.1}$ & 53.2$^{-13.5}$ \\
       & Ours &  & \textbf{66.5}$^{-0.2\phantom{0}}$ & \textbf{66.0}$^{-0.7\phantom{0}}$ & \textbf{65.5}$^{-1.2\phantom{0}}$ & \textbf{64.6}$^{-2.1\phantom{0}}$ & \textbf{63.7}$^{-3.0\phantom{0}}$ & \textbf{62.4}$^{-4.3\phantom{0}}$ \\
      \midrule
      E5-Mistral 7B & SnapViT & 71.8 & 71.4$^{-0.4\phantom{0}}$ & 70.7$^{-1.1\phantom{0}}$ & \textbf{70.2}$^{-1.6\phantom{0}}$ & \textbf{68.1}$^{-3.7\phantom{0}}$ & 62.4$^{-9.4\phantom{0}}$ & 51.3$^{-20.5}$ \\
       & SlimGPT &  & 71.2$^{-0.6\phantom{0}}$ & 69.3$^{-2.5\phantom{0}}$ & 66.1$^{-5.7\phantom{0}}$ & 61.5$^{-10.3}$ & 57.4$^{-14.4}$ & 45.8$^{-26.0}$ \\
       & Ours &  & \textbf{71.6}$^{-0.2\phantom{0}}$ & \textbf{71.0}$^{-0.8\phantom{0}}$ & 69.3$^{-2.5\phantom{0}}$ & 67.3$^{-4.5\phantom{0}}$ & \textbf{65.0}$^{-6.8\phantom{0}}$ & \textbf{62.5}$^{-9.3\phantom{0}}$ \\
      \midrule
      Qwen3 0.6B & SnapViT & 73.5 & \textbf{73.3}$^{-0.2\phantom{0}}$ & \textbf{71.2}$^{-2.3\phantom{0}}$ & \textbf{67.9}$^{-5.6\phantom{0}}$ & 62.3$^{-11.2}$ & 53.9$^{-19.6}$ & 48.4$^{-25.1}$ \\
       & SlimGPT &  & 60.2$^{-13.3}$ & 47.7$^{-25.8}$ & 27.7$^{-45.8}$ & 24.1$^{-49.4}$ & 18.6$^{-54.9}$ & 18.7$^{-54.8}$ \\
       & Ours &  & 73.2$^{-0.3\phantom{0}}$ & 70.8$^{-2.7\phantom{0}}$ & 65.9$^{-7.6\phantom{0}}$ & \textbf{62.8}$^{-10.7}$ & \textbf{58.8}$^{-14.7}$ & \textbf{56.8}$^{-16.7}$ \\
      \midrule
      Qwen3 4B & SnapViT & 76.6 & \textbf{76.1}$^{-0.5\phantom{0}}$ & \textbf{75.4}$^{-1.2\phantom{0}}$ & \textbf{73.8}$^{-2.8\phantom{0}}$ & \textbf{70.3}$^{-6.3\phantom{0}}$ & \textbf{63.2}$^{-13.4}$ & 54.2$^{-22.4}$ \\
       & SlimGPT &  & 73.3$^{-3.3\phantom{0}}$ & 66.8$^{-9.8\phantom{0}}$ & 53.4$^{-23.2}$ & 41.8$^{-34.8}$ & 28.7$^{-47.9}$ & 19.3$^{-57.3}$ \\
       & Ours &  & 75.7$^{-0.9\phantom{0}}$ & 74.4$^{-2.2\phantom{0}}$ & 71.8$^{-4.8\phantom{0}}$ & 67.6$^{-9.0\phantom{0}}$ & 62.9$^{-13.7}$ & \textbf{59.3}$^{-17.3}$ \\
      \midrule
      Qwen3 8B & SnapViT & 77.0 & 76.6$^{-0.4\phantom{0}}$ & 75.6$^{-1.4\phantom{0}}$ & \textbf{74.1}$^{-2.9\phantom{0}}$ & \textbf{71.3}$^{-5.7\phantom{0}}$ & 64.3$^{-12.7}$ & 51.9$^{-25.1}$ \\
       & SlimGPT &  & 75.3$^{-1.7\phantom{0}}$ & 72.3$^{-4.7\phantom{0}}$ & 67.6$^{-9.4\phantom{0}}$ & 58.8$^{-18.2}$ & 48.2$^{-28.8}$ & 33.8$^{-43.2}$ \\
       & Ours &  & \textbf{76.7}$^{-0.3\phantom{0}}$ & \textbf{75.8}$^{-1.2\phantom{0}}$ & 74.0$^{-3.0\phantom{0}}$ & 71.2$^{-5.8\phantom{0}}$ & \textbf{67.7}$^{-9.3\phantom{0}}$ & \textbf{63.3}$^{-13.7}$ \\
      \bottomrule
    \end{tabular}
  }
\end{table}

\begin{table}[t]
  \centering
  \caption{\textbf{Retrieval is sensitive to deep pruning.} MTEB retrieval score for dense and elastically pruned BGE, E5, E5-Mistral, and Qwen3 models. Retrieval degrades faster than other tasks: at 60\% sparsity, BGE-L loses 19.6 points and Qwen3 8B 22.1. On the other hand, models tolerate shallow pruning, and E5-B slightly exceeds its dense score at 20\% sparsity.}
  \label{tab:mteb-retrieval}
  \resizebox{\columnwidth}{!}{
    \begin{tabular}{llcrrrrrr}
      \toprule
      & & & \multicolumn{6}{c}{Sparsity} \\
      \cmidrule(lr){4-9}
      Backbone & Method & Dense & \multicolumn{1}{c}{10\%} & \multicolumn{1}{c}{20\%} & \multicolumn{1}{c}{30\%} & \multicolumn{1}{c}{40\%} & \multicolumn{1}{c}{50\%} & \multicolumn{1}{c}{60\%} \\
      \midrule
      BGE-S & SnapViT & 53.9 & 51.8$^{-2.1\phantom{0}}$ & \textbf{49.6}$^{-4.3\phantom{0}}$ & \textbf{45.2}$^{-8.7\phantom{0}}$ & 37.7$^{-16.2}$ & 30.3$^{-23.6}$ & 25.3$^{-28.6}$ \\
       & SlimGPT &  & 50.9$^{-3.0\phantom{0}}$ & 46.7$^{-7.2\phantom{0}}$ & 41.6$^{-12.3}$ & 34.2$^{-19.7}$ & 23.8$^{-30.1}$ & 13.4$^{-40.5}$ \\
       & Ours &  & \textbf{52.6}$^{-1.3\phantom{0}}$ & 48.5$^{-5.4\phantom{0}}$ & 43.9$^{-10.0}$ & \textbf{39.6}$^{-14.3}$ & \textbf{36.4}$^{-17.5}$ & \textbf{30.6}$^{-23.3}$ \\
      \midrule
      BGE-B & SnapViT & 54.8 & \textbf{54.2}$^{-0.6\phantom{0}}$ & \textbf{52.3}$^{-2.5\phantom{0}}$ & \textbf{50.2}$^{-4.6\phantom{0}}$ & 42.4$^{-12.4}$ & 32.8$^{-22.0}$ & 20.5$^{-34.3}$ \\
       & SlimGPT &  & 52.1$^{-2.7\phantom{0}}$ & 47.2$^{-7.6\phantom{0}}$ & 36.9$^{-17.9}$ & 25.9$^{-28.9}$ & 2.6$^{-52.2}$ & 2.2$^{-52.6}$ \\
       & Ours &  & 53.7$^{-1.1\phantom{0}}$ & \textbf{52.3}$^{-2.5\phantom{0}}$ & \textbf{50.2}$^{-4.6\phantom{0}}$ & \textbf{47.4}$^{-7.4\phantom{0}}$ & \textbf{42.3}$^{-12.5}$ & \textbf{35.2}$^{-19.6}$ \\
      \midrule
      BGE-L & SnapViT & 55.5 & 54.0$^{-1.5\phantom{0}}$ & 51.5$^{-4.0\phantom{0}}$ & 42.8$^{-12.7}$ & 26.1$^{-29.4}$ & 17.5$^{-38.0}$ & 8.7$^{-46.8}$ \\
       & SlimGPT &  & 40.2$^{-15.3}$ & 25.6$^{-29.9}$ & 25.4$^{-30.1}$ & 21.6$^{-33.9}$ & 18.3$^{-37.2}$ & 11.9$^{-43.6}$ \\
       & Ours &  & \textbf{54.6}$^{-0.9\phantom{0}}$ & \textbf{52.5}$^{-3.0\phantom{0}}$ & \textbf{49.4}$^{-6.1\phantom{0}}$ & \textbf{44.6}$^{-10.9}$ & \textbf{40.1}$^{-15.4}$ & \textbf{35.9}$^{-19.6}$ \\
      \midrule
      E5-B & SnapViT & 49.7 & \textbf{49.9}$^{+0.2\phantom{0}}$ & 47.3$^{-2.4\phantom{0}}$ & 41.8$^{-7.9\phantom{0}}$ & 23.6$^{-26.1}$ & 15.0$^{-34.7}$ & 8.8$^{-40.9}$ \\
       & SlimGPT &  & \textbf{49.9}$^{+0.2\phantom{0}}$ & 44.1$^{-5.6\phantom{0}}$ & 29.3$^{-20.4}$ & 33.1$^{-16.6}$ & 26.9$^{-22.8}$ & 18.9$^{-30.8}$ \\
       & Ours &  & 49.5$^{-0.2\phantom{0}}$ & \textbf{50.1}$^{+0.4\phantom{0}}$ & \textbf{47.3}$^{-2.4\phantom{0}}$ & \textbf{45.0}$^{-4.7\phantom{0}}$ & \textbf{39.6}$^{-10.1}$ & \textbf{34.2}$^{-15.5}$ \\
      \midrule
      E5-L & SnapViT & 49.3 & 48.0$^{-1.3\phantom{0}}$ & 37.2$^{-12.1}$ & 17.6$^{-31.7}$ & 8.9$^{-40.4}$ & 8.9$^{-40.4}$ & 4.4$^{-44.9}$ \\
       & SlimGPT &  & 35.5$^{-13.8}$ & 33.7$^{-15.6}$ & 32.8$^{-16.5}$ & 25.8$^{-23.5}$ & 16.5$^{-32.8}$ & 13.3$^{-36.0}$ \\
       & Ours &  & \textbf{49.4}$^{+0.1\phantom{0}}$ & \textbf{48.2}$^{-1.1\phantom{0}}$ & \textbf{47.3}$^{-2.0\phantom{0}}$ & \textbf{47.0}$^{-2.3\phantom{0}}$ & \textbf{44.7}$^{-4.6\phantom{0}}$ & \textbf{41.3}$^{-8.0\phantom{0}}$ \\
      \midrule
      E5-Mistral 7B & SnapViT & 57.6 & 56.8$^{-0.8\phantom{0}}$ & 55.4$^{-2.2\phantom{0}}$ & \textbf{54.5}$^{-3.1\phantom{0}}$ & \textbf{50.9}$^{-6.7\phantom{0}}$ & 37.0$^{-20.6}$ & 18.3$^{-39.3}$ \\
       & SlimGPT &  & 56.1$^{-1.5\phantom{0}}$ & 54.1$^{-3.5\phantom{0}}$ & 49.9$^{-7.7\phantom{0}}$ & 40.0$^{-17.6}$ & 22.5$^{-35.1}$ & 8.6$^{-49.0}$ \\
       & Ours &  & \textbf{57.2}$^{-0.4\phantom{0}}$ & \textbf{55.8}$^{-1.8\phantom{0}}$ & 53.8$^{-3.8\phantom{0}}$ & 48.3$^{-9.3\phantom{0}}$ & \textbf{42.0}$^{-15.6}$ & \textbf{39.3}$^{-18.3}$ \\
      \midrule
      Qwen3 0.6B & SnapViT & 59.6 & \textbf{58.3}$^{-1.3\phantom{0}}$ & 52.7$^{-6.9\phantom{0}}$ & 47.6$^{-12.0}$ & 42.7$^{-16.9}$ & 26.8$^{-32.8}$ & 12.6$^{-47.0}$ \\
       & SlimGPT &  & 22.2$^{-37.4}$ & 5.4$^{-54.2}$ & 0.3$^{-59.3}$ & 0.1$^{-59.5}$ & 0.1$^{-59.5}$ & 0.0$^{-59.6}$ \\
       & Ours &  & 58.0$^{-1.6\phantom{0}}$ & \textbf{54.3}$^{-5.3\phantom{0}}$ & \textbf{49.5}$^{-10.1}$ & \textbf{44.0}$^{-15.6}$ & \textbf{40.8}$^{-18.8}$ & \textbf{37.4}$^{-22.2}$ \\
      \midrule
      Qwen3 4B & SnapViT & 66.4 & 61.9$^{-4.5\phantom{0}}$ & 59.2$^{-7.2\phantom{0}}$ & 55.2$^{-11.2}$ & 46.0$^{-20.4}$ & 38.6$^{-27.8}$ & 21.8$^{-44.6}$ \\
       & SlimGPT &  & 54.4$^{-12.0}$ & 37.9$^{-28.5}$ & 16.4$^{-50.0}$ & 9.6$^{-56.8}$ & 2.3$^{-64.1}$ & 0.0$^{-66.4}$ \\
       & Ours &  & \textbf{62.9}$^{-3.5\phantom{0}}$ & \textbf{59.5}$^{-6.9\phantom{0}}$ & \textbf{56.1}$^{-10.3}$ & \textbf{51.2}$^{-15.2}$ & \textbf{45.4}$^{-21.0}$ & \textbf{40.2}$^{-26.2}$ \\
      \midrule
      Qwen3 8B & SnapViT & 67.4 & 65.6$^{-1.8\phantom{0}}$ & \textbf{63.1}$^{-4.3\phantom{0}}$ & \textbf{60.0}$^{-7.4\phantom{0}}$ & 53.6$^{-13.8}$ & 41.1$^{-26.3}$ & 13.7$^{-53.7}$ \\
       & SlimGPT &  & 61.0$^{-6.4\phantom{0}}$ & 52.3$^{-15.1}$ & 41.5$^{-25.9}$ & 24.0$^{-43.4}$ & 12.2$^{-55.2}$ & 2.0$^{-65.4}$ \\
       & Ours &  & \textbf{66.4}$^{-1.0\phantom{0}}$ & 62.8$^{-4.6\phantom{0}}$ & 58.8$^{-8.6\phantom{0}}$ & \textbf{55.0}$^{-12.4}$ & \textbf{51.7}$^{-15.7}$ & \textbf{45.3}$^{-22.1}$ \\
      \bottomrule
    \end{tabular}
  }
\end{table}

\subsubsection{Elastic Text Retrieval}
\label{sec:appendix-elastic-text-retrieval}

\autoref{tab:beir-elastic-additional} reports mean nDCG@10 for all nine text encoders, where the document index is encoded once by the dense model and queries by its pruned counterparts. Our method outperforms both baselines at every evaluated sparsity on BGE-S and E5-B, and from 40\% sparsity onward on every other backbone except E5-Mistral 7B, where SnapViT leads by up to 1.8 points at up to 40\% sparsity. At 50\% and 60\% sparsity, our method outperforms both baselines on every backbone. Our advantage in elastic retrieval therefore extends across encoder sizes and families over a broad range of query budgets.

\begin{table}[htbp]
  \centering
  \caption{\textbf{Our method preserves elastic retrieval performance across all nine text encoders.} Mean nDCG@10 on the 15 BEIR tasks across all nine text encoders, where documents are encoded by the dense model and queries by its pruned counterparts. Our method outperforms both baselines on every backbone at 50\% and 60\% sparsity.}
  \label{tab:beir-elastic-additional}
  \resizebox{\columnwidth}{!}{
    \begin{tabular}{llcrrrrrr}
      \toprule
      & & & \multicolumn{6}{c}{Sparsity} \\
      \cmidrule(lr){4-9}
      Backbone & Method & Dense & \multicolumn{1}{c}{10\%} & \multicolumn{1}{c}{20\%} & \multicolumn{1}{c}{30\%} & \multicolumn{1}{c}{40\%} & \multicolumn{1}{c}{50\%} & \multicolumn{1}{c}{60\%} \\
      \midrule
      BGE-S & SnapViT & 51.7 & 50.0$^{-1.7\phantom{0}}$ & 46.5$^{-5.2\phantom{0}}$ & 41.6$^{-10.1}$ & 26.6$^{-25.1}$ & 14.2$^{-37.5}$ & 4.9$^{-46.8}$ \\
       & SlimGPT &  & 48.8$^{-2.9\phantom{0}}$ & 41.6$^{-10.1}$ & 28.5$^{-23.2}$ & 9.8$^{-41.9}$ & 2.4$^{-49.3}$ & 0.2$^{-51.5}$ \\
       & Ours &  & \textbf{51.4}$^{-0.3\phantom{0}}$ & \textbf{48.3}$^{-3.4\phantom{0}}$ & \textbf{44.5}$^{-7.2\phantom{0}}$ & \textbf{40.0}$^{-11.7}$ & \textbf{36.7}$^{-15.0}$ & \textbf{32.0}$^{-19.7}$ \\
      \midrule
      BGE-B & SnapViT & 53.2 & \textbf{52.9}$^{-0.3\phantom{0}}$ & \textbf{51.4}$^{-1.8\phantom{0}}$ & 49.6$^{-3.6\phantom{0}}$ & 45.3$^{-7.9\phantom{0}}$ & 29.7$^{-23.5}$ & 12.6$^{-40.6}$ \\
       & SlimGPT &  & 51.4$^{-1.8\phantom{0}}$ & 48.3$^{-4.9\phantom{0}}$ & 43.0$^{-10.2}$ & 32.9$^{-20.3}$ & 2.6$^{-50.6}$ & 0.6$^{-52.6}$ \\
       & Ours &  & 52.4$^{-0.8\phantom{0}}$ & 51.0$^{-2.2\phantom{0}}$ & \textbf{49.7}$^{-3.5\phantom{0}}$ & \textbf{46.9}$^{-6.3\phantom{0}}$ & \textbf{42.8}$^{-10.4}$ & \textbf{35.3}$^{-17.9}$ \\
      \midrule
      BGE-L & SnapViT & 54.3 & 53.2$^{-1.1\phantom{0}}$ & \textbf{52.1}$^{-2.2\phantom{0}}$ & 47.1$^{-7.2\phantom{0}}$ & 30.0$^{-24.3}$ & 17.4$^{-36.9}$ & 5.5$^{-48.8}$ \\
       & SlimGPT &  & 47.5$^{-6.8\phantom{0}}$ & 34.7$^{-19.6}$ & 31.3$^{-23.0}$ & 26.7$^{-27.6}$ & 19.4$^{-34.9}$ & 8.1$^{-46.2}$ \\
       & Ours &  & \textbf{53.6}$^{-0.7\phantom{0}}$ & 52.0$^{-2.3\phantom{0}}$ & \textbf{48.9}$^{-5.4\phantom{0}}$ & \textbf{43.4}$^{-10.9}$ & \textbf{38.4}$^{-15.9}$ & \textbf{33.6}$^{-20.7}$ \\
      \midrule
      E5-B & SnapViT & 50.2 & 49.8$^{-0.4\phantom{0}}$ & 48.1$^{-2.1\phantom{0}}$ & 44.0$^{-6.2\phantom{0}}$ & 30.3$^{-19.9}$ & 19.7$^{-30.5}$ & 4.7$^{-45.5}$ \\
       & SlimGPT &  & 49.8$^{-0.4\phantom{0}}$ & 46.4$^{-3.8\phantom{0}}$ & 35.1$^{-15.1}$ & 41.3$^{-8.9\phantom{0}}$ & 34.4$^{-15.8}$ & 17.4$^{-32.8}$ \\
       & Ours &  & \textbf{50.0}$^{-0.2\phantom{0}}$ & \textbf{49.9}$^{-0.3\phantom{0}}$ & \textbf{49.2}$^{-1.0\phantom{0}}$ & \textbf{47.5}$^{-2.7\phantom{0}}$ & \textbf{45.3}$^{-4.9\phantom{0}}$ & \textbf{41.8}$^{-8.4\phantom{0}}$ \\
      \midrule
      E5-L & SnapViT & 50.5 & \textbf{50.6}$^{+0.1\phantom{0}}$ & 44.1$^{-6.4\phantom{0}}$ & 23.7$^{-26.8}$ & 5.9$^{-44.6}$ & 1.9$^{-48.6}$ & 0.1$^{-50.4}$ \\
       & SlimGPT &  & 43.6$^{-6.9\phantom{0}}$ & 41.1$^{-9.4\phantom{0}}$ & 37.8$^{-12.7}$ & 33.3$^{-17.2}$ & 24.3$^{-26.2}$ & 16.0$^{-34.5}$ \\
       & Ours &  & 50.3$^{-0.2\phantom{0}}$ & \textbf{49.8}$^{-0.7\phantom{0}}$ & \textbf{49.0}$^{-1.5\phantom{0}}$ & \textbf{47.5}$^{-3.0\phantom{0}}$ & \textbf{45.4}$^{-5.1\phantom{0}}$ & \textbf{41.2}$^{-9.3\phantom{0}}$ \\
      \midrule
      E5-Mistral 7B & SnapViT & 56.9 & 56.6$^{-0.3\phantom{0}}$ & \textbf{55.8}$^{-1.1\phantom{0}}$ & \textbf{54.9}$^{-2.0\phantom{0}}$ & \textbf{51.5}$^{-5.4\phantom{0}}$ & 40.9$^{-16.0}$ & 15.4$^{-41.5}$ \\
       & SlimGPT &  & \textbf{56.8}$^{-0.1\phantom{0}}$ & 55.7$^{-1.2\phantom{0}}$ & 52.6$^{-4.3\phantom{0}}$ & 40.4$^{-16.5}$ & 29.2$^{-27.7}$ & 13.2$^{-43.7}$ \\
       & Ours &  & 56.7$^{-0.2\phantom{0}}$ & 55.7$^{-1.2\phantom{0}}$ & 53.7$^{-3.2\phantom{0}}$ & 49.7$^{-7.2\phantom{0}}$ & \textbf{41.7}$^{-15.2}$ & \textbf{33.7}$^{-23.2}$ \\
      \midrule
      Qwen3 0.6B & SnapViT & 55.2 & 54.6$^{-0.6\phantom{0}}$ & 52.7$^{-2.5\phantom{0}}$ & \textbf{50.3}$^{-4.9\phantom{0}}$ & 44.4$^{-10.8}$ & 25.9$^{-29.3}$ & 9.0$^{-46.2}$ \\
       & SlimGPT &  & 36.9$^{-18.3}$ & 8.2$^{-47.0}$ & 1.2$^{-54.0}$ & 0.1$^{-55.1}$ & 0.1$^{-55.1}$ & 0.1$^{-55.1}$ \\
       & Ours &  & \textbf{54.7}$^{-0.5\phantom{0}}$ & \textbf{53.2}$^{-2.0\phantom{0}}$ & 50.1$^{-5.1\phantom{0}}$ & \textbf{47.2}$^{-8.0\phantom{0}}$ & \textbf{41.0}$^{-14.2}$ & \textbf{34.1}$^{-21.1}$ \\
      \midrule
      Qwen3 4B & SnapViT & 61.4 & \textbf{60.4}$^{-1.0\phantom{0}}$ & \textbf{59.1}$^{-2.3\phantom{0}}$ & \textbf{56.3}$^{-5.1\phantom{0}}$ & 53.0$^{-8.4\phantom{0}}$ & 45.5$^{-15.9}$ & 23.9$^{-37.5}$ \\
       & SlimGPT &  & 58.5$^{-2.9\phantom{0}}$ & 50.3$^{-11.1}$ & 23.9$^{-37.5}$ & 13.2$^{-48.2}$ & 3.5$^{-57.9}$ & 0.2$^{-61.2}$ \\
       & Ours &  & 59.9$^{-1.5\phantom{0}}$ & 58.4$^{-3.0\phantom{0}}$ & 56.2$^{-5.2\phantom{0}}$ & \textbf{53.3}$^{-8.1\phantom{0}}$ & \textbf{47.8}$^{-13.6}$ & \textbf{41.5}$^{-19.9}$ \\
      \midrule
      Qwen3 8B & SnapViT & 62.6 & 62.0$^{-0.6\phantom{0}}$ & 60.3$^{-2.3\phantom{0}}$ & 58.8$^{-3.8\phantom{0}}$ & 55.6$^{-7.0\phantom{0}}$ & 47.8$^{-14.8}$ & 7.1$^{-55.5}$ \\
       & SlimGPT &  & 60.9$^{-1.7\phantom{0}}$ & 57.6$^{-5.0\phantom{0}}$ & 49.6$^{-13.0}$ & 21.0$^{-41.6}$ & 7.8$^{-54.8}$ & 2.5$^{-60.1}$ \\
       & Ours &  & \textbf{62.2}$^{-0.4\phantom{0}}$ & \textbf{61.2}$^{-1.4\phantom{0}}$ & \textbf{59.2}$^{-3.4\phantom{0}}$ & \textbf{56.2}$^{-6.4\phantom{0}}$ & \textbf{52.5}$^{-10.1}$ & \textbf{43.6}$^{-19.0}$ \\
      \bottomrule
    \end{tabular}
  }
\end{table}

\subsubsection{Weight Correction}
\label{sec:appendix-weight-correction}

\begin{table}[t]
  \caption{\textbf{Our method is competitive against corrected baselines, and is compatible with weight correction.} Top-1 k-NN accuracy on ImageNet-1k for pruned AugReg ViT-B/16, without (Plain) and with (Corrected) post-pruning weight correction. Uncorrected, our method outperforms SparseGPT and FPTP, only trailing corrected SnapViT by at most 4.0\%. Corrected, it matches SnapViT at up to 40\% sparsity and surpasses it beyond.}
  \label{tab:correction}
  \resizebox{\columnwidth}{!}{
    \begin{tabular}{lcccccccccccc}
      \toprule
      & \multicolumn{12}{c}{Sparsity} \\
      \cmidrule(lr){2-13}
       & \multicolumn{2}{c}{10\% / 31.7 G} & \multicolumn{2}{c}{20\% / 28.2 G} & \multicolumn{2}{c}{30\% / 24.8 G} & \multicolumn{2}{c}{40\% / 21.2 G} & \multicolumn{2}{c}{50\% / 17.8 G} & \multicolumn{2}{c}{60\% / 14.2 G} \\
      \cmidrule(lr){2-3}\cmidrule(lr){4-5}\cmidrule(lr){6-7}\cmidrule(lr){8-9}\cmidrule(lr){10-11}\cmidrule(lr){12-13}
      Method & Plain & Corrected & Plain & Corrected & Plain & Corrected & Plain & Corrected & Plain & Corrected & Plain & Corrected \\
      \midrule
      SparseGPT & 78.8 & 82.4$^{\phantom{0}\deltapos{+3.6}}$ & 64.2 & 79.3$^{\deltapos{+15.1}}$ & 36.5 & 72.2$^{\deltapos{+35.7}}$ & 20.3 & 59.9$^{\deltapos{+39.6}}$ & 14.1 & 42.4$^{\deltapos{+28.3}}$ & 10.2 & 29.1$^{\deltapos{+18.9}}$ \\
      FPTP & 82.0 & 82.1$^{\phantom{0}\deltapos{+0.1}}$ & 78.5 & 79.5$^{\phantom{0}\deltapos{+1.0}}$ & 71.3 & 76.2$^{\phantom{0}\deltapos{+4.9}}$ & 44.3 & 57.0$^{\deltapos{+12.7}}$ & 17.2 & 19.3$^{\phantom{0}\deltapos{+2.1}}$ & 12.6 & 13.3$^{\phantom{0}\deltapos{+0.7}}$ \\
      SnapViT & 82.2 & \textbf{82.8}$^{\phantom{0}\deltapos{+0.6}}$ & 77.5 & \textbf{81.0}$^{\phantom{0}\deltapos{+3.5}}$ & 70.1 & \textbf{78.4}$^{\phantom{0}\deltapos{+8.3}}$ & 57.7 & \textbf{73.0}$^{\deltapos{+15.3}}$ & 28.4 & 63.3$^{\deltapos{+34.9}}$ & 17.7 & 47.5$^{\deltapos{+29.8}}$ \\
      \rowcolor{recipehighlight} Ours & \textbf{82.6} & \textbf{82.8}$^{\phantom{0}\deltapos{+0.2}}$ & \textbf{80.1} & \textbf{81.0}$^{\phantom{0}\deltapos{+0.9}}$ & \textbf{75.7} & 78.1$^{\phantom{0}\deltapos{+2.4}}$ & \textbf{69.1} & 72.6$^{\phantom{0}\deltapos{+3.5}}$ & \textbf{59.3} & \textbf{64.3}$^{\phantom{0}\deltapos{+5.0}}$ & \textbf{45.2} & \textbf{54.0}$^{\phantom{0}\deltapos{+8.8}}$ \\
      \bottomrule
    \end{tabular}
  }
\end{table}

\begin{table}[t]
  \caption{\textbf{Our correction-free method matches or outperforms weight-correction methods on DINOv3 ViT-H+/16.} Top-1 k-NN accuracy on ImageNet-1k for DINOv3 ViT-H+/16 pruned with corrected SparseGPT and SnapViT, versus our correction-free method. At 10\% and 20\% sparsity, our method remains within 0.1\% of the best corrected baseline. Our advantage grows with sparsity, to up to 18.2\% over corrected SnapViT, the runner-up, at 60\%.}
  \label{tab:correction-dinov3}
  \resizebox{\columnwidth}{!}{
    \begin{tabular}{lccccccc}
      \toprule
      & & \multicolumn{6}{c}{Sparsity} \\
      \cmidrule(lr){3-8}
      Method & Correction & 10\% / 398.6 G & 20\% / 354.4 G & 30\% / 310.0 G & 40\% / 265.8 G & 50\% / 221.7 G & 60\% / 177.5 G \\
      \midrule
      SparseGPT & \ding{52} & \textbf{85.2} & 85.1 & 80.6 & 69.6 & 55.4 & 38.3 \\
      SnapViT & \ding{52}  & \textbf{85.2} & \textbf{85.2} & 85.0 & 83.5 & 76.8 & 61.2 \\
      \rowcolor{recipehighlight} Ours & \ding{56} &  85.1 & 85.1 & \textbf{85.1} & \textbf{84.7} & \textbf{83.3} & \textbf{79.4} \\
      \bottomrule
    \end{tabular}
  }
\end{table}

In our main experiments, we evaluate our method and the baselines without retraining or weight correction to isolate the effect of unit selection. Weight correction adjusts the remaining weights to compensate for those removed by pruning. We test whether our method remains competitive when baselines use weight correction, and whether applying correction to our method further improves accuracy. As correction produces a separate set of weights for each sparsity target, applying it to our method sacrifices the ability to serve multiple sparsity targets with a single set of weights.

\noindent\textbf{Vision encoders.}~\autoref{tab:correction} compares pruning methods with and without weight correction on AugReg ViT-B/16, evaluated via top-1 k-NN accuracy on ImageNet-1k. For our method, we apply SparseGPT-like weight correction~\citep{frantar2023sparsegpt}. Without correction, our method outperforms all uncorrected baselines at every sparsity target. It also outperforms corrected SparseGPT at every sparsity target and corrected FPTP at every target except 30\%, while remaining within 4.0\% of corrected SnapViT across all targets. Correction uniformly improves our method, by up to 8.8\% at 60\% sparsity, where it surpasses corrected SnapViT by 6.5\%. Furthermore,~\autoref{tab:correction-dinov3} shows that our uncorrected method outperforms both corrected baselines on DINOv3 ViT-H+/16 from 30\% sparsity onward, leading corrected SnapViT by 18.2\% at 60\%. Overall, these results show that our method remains competitive with corrected baselines, and can benefit from weight correction.

\begin{table}[t]
  \caption{\textbf{Our uncorrected text encoders are competitive with corrected SlimGPT.}  MTEB(eng, v2) score for the nine text encoders, comparing our method without correction against SlimGPT with and without correction. Our method outperforms corrected SlimGPT at every evaluated sparsity target on BGE-B, BGE-L, Qwen3 0.6B, and Qwen3 8B, and on seven of nine encoders between 40\% and 60\% sparsity.}
  \label{tab:text-correction}
  \resizebox{\columnwidth}{!}{
    \begin{tabular}{llcrrrrrr}
      \toprule
      & & & \multicolumn{6}{c}{Sparsity} \\
      \cmidrule(lr){4-9}
      Backbone & Method & Dense & \multicolumn{1}{c}{10\%} & \multicolumn{1}{c}{20\%} & \multicolumn{1}{c}{30\%} & \multicolumn{1}{c}{40\%} & \multicolumn{1}{c}{50\%} & \multicolumn{1}{c}{60\%} \\
      \midrule
      BGE-S & SlimGPT & 64.3 & 63.2$^{-1.1\phantom{0}}$ & 61.2$^{-3.1\phantom{0}}$ & 58.4$^{-5.9\phantom{0}}$ & 54.7$^{-9.6\phantom{0}}$ & 48.9$^{-15.4}$ & 41.8$^{-22.5}$ \\
       & SlimGPT (Corrected) &  & 63.6$^{-0.7\phantom{0}}$ & \textbf{62.5}$^{-1.8\phantom{0}}$ & \textbf{60.4}$^{-3.9\phantom{0}}$ & 58.3$^{-6.0\phantom{0}}$ & 54.0$^{-10.3}$ & 48.3$^{-16.0}$ \\
       & Ours &  & \textbf{63.7}$^{-0.6\phantom{0}}$ & 62.3$^{-2.0\phantom{0}}$ & 60.3$^{-4.0\phantom{0}}$ & \textbf{58.4}$^{-5.9\phantom{0}}$ & \textbf{56.7}$^{-7.6\phantom{0}}$ & \textbf{54.3}$^{-10.0}$ \\
      \midrule
      BGE-B & SlimGPT & 65.2 & 63.8$^{-1.4\phantom{0}}$ & 61.2$^{-4.0\phantom{0}}$ & 57.1$^{-8.1\phantom{0}}$ & 53.4$^{-11.8}$ & 40.9$^{-24.3}$ & 40.6$^{-24.6}$ \\
       & SlimGPT (Corrected) &  & 64.1$^{-1.1\phantom{0}}$ & 63.1$^{-2.1\phantom{0}}$ & 61.6$^{-3.6\phantom{0}}$ & 58.1$^{-7.1\phantom{0}}$ & 55.2$^{-10.0}$ & 32.2$^{-33.0}$ \\
       & Ours &  & \textbf{64.8}$^{-0.4\phantom{0}}$ & \textbf{64.2}$^{-1.0\phantom{0}}$ & \textbf{63.2}$^{-2.0\phantom{0}}$ & \textbf{61.8}$^{-3.4\phantom{0}}$ & \textbf{59.5}$^{-5.7\phantom{0}}$ & \textbf{56.4}$^{-8.8\phantom{0}}$ \\
      \midrule
      BGE-L & SlimGPT & 65.8 & 59.1$^{-6.7\phantom{0}}$ & 51.8$^{-14.0}$ & 50.1$^{-15.7}$ & 49.6$^{-16.2}$ & 47.1$^{-18.7}$ & 40.0$^{-25.8}$ \\
       & SlimGPT (Corrected) &  & 63.6$^{-2.2\phantom{0}}$ & 58.0$^{-7.8\phantom{0}}$ & 56.1$^{-9.7\phantom{0}}$ & 53.3$^{-12.5}$ & 53.0$^{-12.8}$ & 45.7$^{-20.1}$ \\
       & Ours &  & \textbf{65.5}$^{-0.3\phantom{0}}$ & \textbf{64.8}$^{-1.0\phantom{0}}$ & \textbf{63.6}$^{-2.2\phantom{0}}$ & \textbf{61.2}$^{-4.6\phantom{0}}$ & \textbf{59.5}$^{-6.3\phantom{0}}$ & \textbf{57.1}$^{-8.7\phantom{0}}$ \\
      \midrule
      E5-B & SlimGPT & 62.5 & \textbf{62.4}$^{-0.1\phantom{0}}$ & 59.9$^{-2.6\phantom{0}}$ & 54.1$^{-8.4\phantom{0}}$ & 55.6$^{-6.9\phantom{0}}$ & 51.9$^{-10.6}$ & 48.0$^{-14.5}$ \\
       & SlimGPT (Corrected) &  & 62.2$^{-0.3\phantom{0}}$ & 61.6$^{-0.9\phantom{0}}$ & 59.6$^{-2.9\phantom{0}}$ & 57.3$^{-5.2\phantom{0}}$ & 53.5$^{-9.0\phantom{0}}$ & 48.2$^{-14.3}$ \\
       & Ours &  & 62.1$^{-0.4\phantom{0}}$ & \textbf{62.0}$^{-0.5\phantom{0}}$ & \textbf{60.8}$^{-1.7\phantom{0}}$ & \textbf{59.8}$^{-2.7\phantom{0}}$ & \textbf{57.7}$^{-4.8\phantom{0}}$ & \textbf{55.6}$^{-6.9\phantom{0}}$ \\
      \midrule
      E5-L & SlimGPT & 62.4 & 54.7$^{-7.7\phantom{0}}$ & 55.6$^{-6.8\phantom{0}}$ & 54.6$^{-7.8\phantom{0}}$ & 52.0$^{-10.4}$ & 46.1$^{-16.3}$ & 43.5$^{-18.9}$ \\
       & SlimGPT (Corrected) &  & \textbf{63.3}$^{+0.9\phantom{0}}$ & \textbf{62.6}$^{+0.2\phantom{0}}$ & \textbf{61.3}$^{-1.1\phantom{0}}$ & 59.2$^{-3.2\phantom{0}}$ & 56.0$^{-6.4\phantom{0}}$ & 53.1$^{-9.3\phantom{0}}$ \\
       & Ours &  & 62.4$^{+0.0\phantom{0}}$ & 61.7$^{-0.7\phantom{0}}$ & 61.1$^{-1.3\phantom{0}}$ & \textbf{60.3}$^{-2.1\phantom{0}}$ & \textbf{59.1}$^{-3.3\phantom{0}}$ & \textbf{57.3}$^{-5.1\phantom{0}}$ \\
      \midrule
      E5-Mistral 7B & SlimGPT & 68.3 & 67.5$^{-0.8\phantom{0}}$ & 65.6$^{-2.7\phantom{0}}$ & 62.1$^{-6.2\phantom{0}}$ & 56.3$^{-12.0}$ & 48.9$^{-19.4}$ & 36.7$^{-31.6}$ \\
       & SlimGPT (Corrected) &  & 67.9$^{-0.4\phantom{0}}$ & 67.1$^{-1.2\phantom{0}}$ & 65.4$^{-2.9\phantom{0}}$ & \textbf{63.2}$^{-5.1\phantom{0}}$ & \textbf{60.7}$^{-7.6\phantom{0}}$ & \textbf{57.2}$^{-11.1}$ \\
       & Ours &  & \textbf{68.1}$^{-0.2\phantom{0}}$ & \textbf{67.3}$^{-1.0\phantom{0}}$ & \textbf{65.6}$^{-2.7\phantom{0}}$ & 62.7$^{-5.6\phantom{0}}$ & 59.4$^{-8.9\phantom{0}}$ & 56.8$^{-11.5}$ \\
      \midrule
      Qwen3 0.6B & SlimGPT & 70.1 & 50.9$^{-19.2}$ & 37.4$^{-32.7}$ & 21.0$^{-49.1}$ & 18.3$^{-51.8}$ & 14.1$^{-56.0}$ & 14.1$^{-56.0}$ \\
       & SlimGPT (Corrected) &  & 69.1$^{-1.0\phantom{0}}$ & 62.0$^{-8.1\phantom{0}}$ & 56.6$^{-13.5}$ & 53.2$^{-16.9}$ & 47.4$^{-22.7}$ & 42.0$^{-28.1}$ \\
       & Ours &  & \textbf{69.5}$^{-0.6\phantom{0}}$ & \textbf{66.8}$^{-3.3\phantom{0}}$ & \textbf{61.9}$^{-8.2\phantom{0}}$ & \textbf{58.2}$^{-11.9}$ & \textbf{54.4}$^{-15.7}$ & \textbf{52.1}$^{-18.0}$ \\
      \midrule
      Qwen3 4B & SlimGPT & 74.1 & 68.7$^{-5.4\phantom{0}}$ & 59.8$^{-14.3}$ & 44.4$^{-29.7}$ & 34.0$^{-40.1}$ & 22.2$^{-51.9}$ & 14.6$^{-59.5}$ \\
       & SlimGPT (Corrected) &  & \textbf{73.0}$^{-1.1\phantom{0}}$ & \textbf{71.6}$^{-2.5\phantom{0}}$ & \textbf{68.6}$^{-5.5\phantom{0}}$ & \textbf{64.9}$^{-9.2\phantom{0}}$ & \textbf{59.9}$^{-14.2}$ & \textbf{56.1}$^{-18.0}$ \\
       & Ours &  & 72.6$^{-1.5\phantom{0}}$ & 70.8$^{-3.3\phantom{0}}$ & 68.0$^{-6.1\phantom{0}}$ & 63.6$^{-10.5}$ & 58.6$^{-15.5}$ & 54.6$^{-19.5}$ \\
      \midrule
      Qwen3 8B & SlimGPT & 74.6 & 71.8$^{-2.8\phantom{0}}$ & 67.4$^{-7.2\phantom{0}}$ & 61.2$^{-13.4}$ & 50.3$^{-24.3}$ & 39.4$^{-35.2}$ & 26.1$^{-48.5}$ \\
       & SlimGPT (Corrected) &  & 73.6$^{-1.0\phantom{0}}$ & 72.0$^{-2.6\phantom{0}}$ & 69.4$^{-5.2\phantom{0}}$ & 65.7$^{-8.9\phantom{0}}$ & 60.4$^{-14.2}$ & 56.7$^{-17.9}$ \\
       & Ours &  & \textbf{74.2}$^{-0.4\phantom{0}}$ & \textbf{72.6}$^{-2.0\phantom{0}}$ & \textbf{70.3}$^{-4.3\phantom{0}}$ & \textbf{67.3}$^{-7.3\phantom{0}}$ & \textbf{63.8}$^{-10.8}$ & \textbf{58.9}$^{-15.7}$ \\
      \bottomrule
    \end{tabular}
  }
\end{table}

\noindent\textbf{Text encoders.}~\autoref{tab:text-correction} compares our uncorrected method with SlimGPT with and without weight correction on MTEB(eng, v2) across all nine text encoders. Our uncorrected method outperforms corrected SlimGPT at every evaluated sparsity target on BGE-B, BGE-L, Qwen3 0.6B, and Qwen3 8B. Between 40\% and 60\% sparsity, it outperforms corrected SlimGPT on seven of nine encoders. Corrected SlimGPT outperforms our method on E5-Mistral 7B and Qwen3 4B in this range, but by 1.5 points at most. These results show that our approach is competitive with corrected SlimGPT under deep pruning, while using a single set of weights across sparsity targets.

\subsection{Experimental Details}

\subsubsection{Backbones}
\label{sec:appendix-backbones}

For vision, we evaluate 18 transformer encoders ranging from 22M to 7B parameters. For AugReg~\citep{steiner2022how}, we evaluate ViT-S/16, ViT-B/16, and ViT-L/16 models. For DeiT~\citep{touvron2021training}, we evaluate the ViT-B/16. For DeiT-III~\citep{touvron2022deit}, we evaluate ViT-S/16, ViT-B/16, ViT-L/16, and ViT-H/14 variants. For DINOv3~\citep{simeoni2026dinov3}, we evaluate the ViT-B/16 and ViT-H+/16. For SigLIPv2~\citep{tschannen2025siglip}, we evaluate the ViT-B/16 and ViT-G/16. For DINO WebSSL~\citep{fan2025scaling}, we evaluate all six variants, i.e., 300M, 1B, 2B, 3B, 5B, and 7B.

For text, we evaluate nine encoders ranging from 33M to 8B parameters, including five bidirectional encoders, i.e., BGE-S, BGE-B, and BGE-L~\citep{xiao2024c}, and E5-B and E5-L~\citep{wang2022text}, ranging from 33M to 334M parameters. We also include four decoder-based embedders, namely Qwen3 0.6B, Qwen3 4B, and Qwen3 8B~\citep{zhang2025qwen3}, and E5-Mistral 7B~\citep{wang2024improving}.

\subsubsection{Initial Importance Score Estimation}
\label{sec:appendix-importance-scores}

In this section, we describe the modality-specific proxy losses used to compute the initial importance scores, and their calibration data.

\noindent\textbf{Vision Encoders} Following~\cite{simoncini2026elastic}, we use the self-supervised DINO~\citep{caron2021emerging} objective, as it does not require labels or a specialized head. We use two global crops $x_n^g$ with scale in $[0.25, 1]$, and 10 local crops $x_m^l$ with scale in $[0.05, 0.25]$, without any color augmentation. A teacher model encodes the global crops, while the student encodes all crops. In our case, the teacher and student are the same model, but no gradients are computed for the teacher. The loss is then taken as the cross-entropy between each (teacher, student) crop pair, skipping same-view pairs:
\begin{equation}
  \mathcal{L_\text{DINO}} = \sum_{n=1}^{n_g}\sum_{m=1}^{n_l} \mathcal{L_\text{CE}}\left(\text{sg} \left[  f\left(x_n^g\right)' \right], f\left(x_m^l\right)'\right),
\end{equation}
where $f(\cdot)'$ are L2-normalized features and $\text{sg}[\cdot]$ is the stop-gradient operator. Unlike DINO, we use no projection head, and we apply neither centering nor sharpening. Unless specified otherwise, we use 1000 random samples from ImageNet-1k~\citep{deng2009imagenet} for calibration. The batch size depends on model size: 32 for ViT-S, 16 for ViT-B, 8 for ViT-L, and 1 for ViT-H and ViT-G. For DINO WebSSL models we always use a batch size of 2.

\noindent\textbf{Text Encoders} We use the symmetrized in-batch InfoNCE objective~\citep{oord2018representation, gunther2023jina} formulated as:
\begin{equation}
    \mathcal{L} = \frac{\mathcal{L_\text{CE}}(z, y) + \mathcal{L_\text{CE}}(z^\top, y)}{2},
    \quad
    z = \frac{QP^\top}{\tau},
    \quad
    \begin{aligned}
        Q_i = f(q_i)', \\
        P_i = f(p_i)',
    \end{aligned}
    \quad
    y = [0, ..., n - 1],
\end{equation}
where $q_i$ is a query, $p_i$ a passage, $f(\cdot)'$ the L2-normalized pooled embedding, using each model's own pooling, $\tau = 0.02$ the temperature, $y$ the labels, and $n$ the batch size. As each query has a single positive passage at the same in-batch index, the labels $y$ are the identity mapping. The batch size depends on model size: 64 for BGE-S, 32 for BGE-B and E5-B, 16 for BGE-L, E5-L, and Qwen3 0.6B, 4 for Qwen3 4B, 2 for Qwen3 8B, and 8 for E5-Mistral 7B. As for the calibration data, we use 1000 (query, passage) pairs from MS MARCO~\citep{bajaj2016ms} across all our experiments, with samples truncated to a maximum length of 256 tokens.

\noindent\textbf{Per-unit reduction} We obtain each unit's score by averaging its associated weight scores, as described in Section~\ref{sec:method-importance-scores}, with two architecture-specific modifications. For models using SwiGLU MLPs~\citep{shazeer2020glu}, i.e., DINOv3 ViT-H+/16 and DINO WebSSL, we average scores only over the two input projections. For grouped-query attention~\citep{ainslie2023gqa} models, i.e., Qwen3 and E5-Mistral, we prune individual query heads and score each head using its corresponding columns in the output projection, as the key and value projections are shared across query heads.

\subsubsection{Differentiable Soft Pruning}
\label{sec:appendix-soft-pruning}

In this section, we detail the hyperparameters and calibration data of the differentiable soft pruning of Section~\ref{sec:method-score-refinement}, and how the resulting ranking is applied at test time.

\noindent\textbf{Optimization.} We initialize the log-scores $z$ by clamping the importance scores to a minimum of $10^{-12}$ before taking logarithms. For vision encoders, we then optimize $z$ using Adam~\citep{kingma2014adam}, with a learning rate $\eta = 0.1$, parameters $\beta_1 = 0.9$, $\beta_2 = 0.999$ and no weight decay, for 300 steps. We optimize over 14 sparsity targets, from 5\% to 70\% in 5\% increments. At each step and for each sparsity target $b$, we compute the threshold $\kappa_b$ from the current scores, forward the calibration batch through the soft-pruned model and compute the loss. The per-target loss terms are then averaged. We use a sigmoid temperature $\tau = 0.2$. The batch size depends on the model size. For vision we use 64 for a ViT-S, 32 for a ViT-B, 16 for a ViT-L, 8 for a ViT-H, 4 for a ViT-G. For DINO WebSSL models, we use a batch size of 16 across all configurations. Text encoders use the same configuration, differing only in $\eta = 0.2$, $\tau = 0.1$ and batch size: 64 for BGE-S, 32 for BGE-B and E5-B, 16 for BGE-L, E5-L, and Qwen3 0.6B, 8 for E5-Mistral 7B, 4 for Qwen3 4B, and 2 for Qwen3 8B.

\noindent\textbf{Calibration data.} For vision encoders, we use up to 2000 random samples from ImageNet-1k~\citep{deng2009imagenet}. For text encoders, we instead use up to 1000 passages from MS MARCO~\citep{bajaj2016ms}, truncated to a maximum length of 256 tokens. As the batch size shrinks with model size, large models do not see the whole calibration set: the number of unique samples is $\min(300 \times \text{batch size}, |D|)$, e.g., 2000 for a ViT-B, 1200 for a ViT-G, and 600 for Qwen3 8B.

\noindent\textbf{Objectives.} Where possible, the loss $\ell$ is the model's native objective: cross-entropy for supervised models, i.e., AugReg, DeiT and DeiT-III, and the sigmoid image-text contrastive loss of SigLIP~\citep{zhai2023sigmoid} between the soft-pruned image embedding and pre-computed text embeddings of all ImageNet-1k class names for SigLIPv2. For DINOv3, DINO WebSSL, and text encoders, we instead use the cosine distance between soft-pruned and dense features.

\noindent\textbf{Test-time pruning.} To prune a model to a sparsity target $b$, we walk the optimized ranking from the top, keep units until their cumulative FLOP cost reaches $(1 - b)$ of the dense model's, and remove the rest. As a safeguard, every block retains at least 5\% of its MLP neurons and 20\% of its attention heads.

\subsubsection{Theoretical FLOPs of functional units}
\label{sec:unit-flops}

For the ranking optimization and the computation of pruning thresholds $\kappa_b$, we only consider FLOPs associated with prunable functional units, i.e., the input and output projections for MLP neurons and query, key, value, and output projections, attention matrix multiplications, and softmax operations for attention heads, counting each multiplication and addition as one FLOP, equivalent to two FLOPs per MAC. Let $E$ denote the embedding dimension, $H_d$ the attention head dimension, and $N$ the number of tokens, defined as $N = \lfloor W/P \rfloor^2 + 1$ for vision, given a square image of width $W$ and patch size $P$, where 1 accounts for the \texttt{[CLS]} token. As the length of text sequences varies, we use $N=256$ for text encoders. We omit embedding lookup and the comparatively small costs of normalization, activation functions, residual connections, pooling, and prediction heads.

The FLOP cost of one MLP neuron is $C_{\mathrm{MLP}} = 2JEN$, where $J$ is the number of linear projections involving that neuron. A standard MLP has $J=2$, corresponding to its input and output projections, whereas a SwiGLU MLP~\citep{shazeer2020glu} has $J=3$, corresponding to its gate, up, and down projections. For standard multi-head attention, the cost of one head is $C_{\mathrm{MHA}} = 8EH_dN + (4H_d + 3)N^2$, where the first term accounts for the query, key, value, and output projections, and the second term accounts for computing attention logits, applying softmax, and reducing the values. For grouped-query attention~\citep{ainslie2023gqa}, multiple query heads share each key-value head. Let $A_q$ and $A_{kv}$ denote the numbers of retained query and key-value heads in a block. Its exact attention cost is $C_{\mathrm{GQA}} = 4EH_dN(A_q + A_{kv}) + (4H_d + 3)A_qN^2$. For ranking, we assign each query head the amortized cost $C_q = 4EH_dN(1 + 1/R) + (4H_d + 3)N^2$, where $R$ is the number of query heads sharing one key-value head. Pruning a query head removes its query projection, corresponding output-projection columns, and attention computation, whereas the shared key and value projections are removed only when every associated query head is pruned. We use the exact block cost when reporting model FLOPs. Under this accounting, for both dense and pruned models, we report the total cost $C_{\mathrm{model}}$ as
\begin{equation}
    C_{\mathrm{model}} = C_{\mathrm{input}} + \sum_{\ell=1}^{L}\left(F_\ell C_{\mathrm{MLP}} + C_{\mathrm{attn}}^{(\ell)}\right),
\end{equation}
where $F_\ell$ is the number of retained feed-forward neurons. For standard multi-head attention, $C_{\mathrm{attn}}^{(\ell)}$ is the number of retained heads times $C_{\mathrm{MHA}}$. For grouped-query attention, it is $C_{\mathrm{GQA}}$ evaluated using the retained query-head and key-value-head counts. For vision models, $C_{\mathrm{input}} = 6(N-1)P^2E$ is the patch-projection cost. For text models, we omit embedding lookup and set $C_{\mathrm{input}}=0$.

\subsubsection{Evaluation}
\label{sec:appendix-evaluation}

\noindent\textbf{k-NN classification.} For each dataset, we build an index using L2-normalized training features. We then encode each test image, L2-normalize its features, retrieve the top-20 nearest neighbors by cosine similarity, and predict its label via majority voting. In the symmetric setting, the index is rebuilt for each sparsity target using the same pruned encoder as the queries, and in the elastic setting, it is built once using dense features.

\noindent\textbf{Linear probing.} For each sparsity target, we train a linear head for 100 epochs on frozen, non-normalized features, using SGD with momentum 0.9, batch size 256, learning rate $10^{-3}$ decayed to 0 with a cosine schedule, and no weight decay. We report the last epoch performance.

\noindent\textbf{In-context semantic segmentation.} We probe the quality of pruned models' dense features via in-context semantic segmentation using the Hummingbird protocol~\citep{balazevic2023towards}, as implemented by \citet{pariza2024hbird}. We evaluate models on Pascal VOC 2012~\citep{everingham2010pascal} with $512 \times 512$ inputs, two augmentation epochs, a memory bank of $1024 \times 10^4$ patches, and 30 nearest neighbors, using Faiss~\citep{douze2024faiss, johnson2019billion} as the retrieval backend.

\subsection{Does pruning translate to speedups?}
\label{sec:appendix-speedups}

In the main text, we use theoretical GFLOPs to measure computational cost across methods. Here, we expand on the performance results reported in Section~\ref{sec:elastic-inference}, measuring throughput at batch size 128 across sparsity targets, latency across batch sizes at 60\% sparsity, and the cost of model switching. Furthermore, our main experiments do not constrain pruning to produce GPU-friendly matrix shapes. While attention heads are pruned whole, so that the remaining projections are always a multiple of the head dimension, which is generally GPU-friendly, MLPs may end up with ragged hidden dimensions. We therefore also measure the effect of constraining pruning to remove MLP neurons in blocks of $A \in \{8, 64\}$, obtained by rounding the ranking post-hoc and re-distributing odd neurons. We denote unconstrained pruning as $A=1$. All measurements use \texttt{PyTorch 2.13.0+cu130} and \texttt{timm 1.0.28} in fp16, and we report performance for both eager execution and \texttt{torch.compile}. Jobs are run on a shared node, and are allocated a single NVIDIA A100 with 40 GB of VRAM, 8 cores of an Intel Xeon Platinum 8360Y at 2.40 GHz and 60 GB of RAM. We measure throughput as total samples divided by the elapsed time over 200 consecutive forward passes at batch size 128, after 10 warmup passes, synchronizing the device before and after the timed loop. For latency, we report the median of 500 individually timed forward passes after 20 warmup passes, synchronizing the device after each pass. As for the cost of switching between budgets, we measure it across 72 transitions $b_1 \rightarrow b_2$. For each transition, we run 50 timed swaps, and measure the median switching cost. We report the mean of these medians.

\autoref{tab:tiling-accuracy} shows that constraining pruning to GPU-friendly blocks generally preserves accuracy at low-to-moderate sparsity, but can have a larger effect at high sparsity. At 60\% sparsity, blocks of 8 and 64 trail unconstrained pruning by 3.4 and 3.2 points on ImageNet-1k. On the transfer average, they gain 2.9 and 2.5 points at 40\% sparsity. Tiling also matters for speed: as shown in~\autoref{tab:tiling-throughput}, in eager execution unconstrained pruning at shallow sparsity is slower than the dense model, dropping to $0.70\times$ the dense throughput at 20\% sparsity, while pruning in blocks is consistently faster, by up to $1.63\times$ over unconstrained pruning at the same sparsity. Compilation largely closes this gap, as compiled models pruned to 60\% sparsity reach up to $1.80\times$ the dense throughput. Latency follows a similar trend, as shown in~\autoref{tab:tiling-latency}. Under eager execution, latency is dominated by kernel launch overhead up to batch size 16, where pruning yields no speedup, and at larger batches tiling significantly improves performance, with $A=64$ reaching $1.60\times$ versus $1.26\times$ for unconstrained pruning at batch size 128. Compilation reduces the improvement due to tiling, and increases speedups, up to $1.79\times$ at batch size 128, yet at batch size 1 pruning still has no effect under default compilation. Compiling with \texttt{reduce-overhead} mode, which captures CUDA graphs, removes the remaining launch overhead and yields a $1.26\times$ speedup with $A=64$ and $1.09\times$ for unconstrained pruning.

Finally, to quantify the cost of switching between budgets, we evaluate AugReg ViT-B/16 with $A=8$, switching between the dense model and sparsity targets from 10\% to 60\% in both sequential and random order. The process holds one resident set of dense weights, with pruned models represented as truncated views. Since the fused QKV projections cannot be truncated as a contiguous slice, we gather them into a fixed buffer per block, shared across budgets. With weights and buffers stored in fp32 and fp16 inference, these buffers require approximately 85 MB of additional weight storage. Compiled graphs are cached per budget. The mean switching cost is 0.46 milliseconds, less than 10\% of a batch-1 forward in eager mode. Compiling a newly seen budget takes 12 to 49 seconds, but this cost can be paid once at startup for the deployed budgets.

\begin{table}[t]
    \centering
    \caption{\textbf{GPU-friendly tiling helps latency.} Latency of dense and 60\% pruned AugReg ViT-B/16 models across batch sizes, under eager execution and \texttt{torch.compile}, for unconstrained pruning, i.e., $A=1$, and pruning in blocks of 8 or 64 neurons, reported as speedup over the dense model. Under eager execution, latency is launch-bound up to batch size 16 and pruning yields no speedup, while tiling accounts for most of the gains at larger batches. Compilation lifts speedups across the board, up to $1.79\times$ at batch size 128, and at batch size 1 gains only materialize without launch overhead.}
    \label{tab:tiling-latency}
    \begin{threeparttable}
        \begin{tabular}{lcccccccc}
            \toprule
             & \multicolumn{4}{c}{Eager} & \multicolumn{4}{c}{Compiled} \\
            \cmidrule(lr){2-5}\cmidrule(lr){6-9}
            Batch & Dense & A=1 & A=8 & A=64 & Dense & A=1 & A=8 & A=64 \\
            \midrule
            1 & 5.402~ms & 1.00$\times$ & 0.95$\times$ & 1.00$\times$ & 3.050~ms & 0.98$\times$ & 0.97$\times$ & 1.00$\times$ \\
            1\tnote{$\dagger$} & -- & -- & -- & -- & 1.812~ms & 1.09$\times$ & 1.25$\times$ & 1.26$\times$ \\
            4 & 5.666~ms & 1.06$\times$ & 1.02$\times$ & 1.03$\times$ & 3.196~ms & 0.99$\times$ & 1.00$\times$ & 1.04$\times$ \\
            16 & 5.806~ms & 1.01$\times$ & 0.99$\times$ & 0.97$\times$ & 5.226~ms & 1.42$\times$ & 1.52$\times$ & 1.57$\times$ \\
            32 & 10.461~ms & 1.22$\times$ & 1.54$\times$ & 1.57$\times$ & 9.214~ms & 1.59$\times$ & 1.67$\times$ & 1.72$\times$ \\
            64 & 19.003~ms & 1.26$\times$ & 1.56$\times$ & 1.58$\times$ & 16.468~ms & 1.68$\times$ & 1.72$\times$ & 1.76$\times$ \\
            128 & 35.851~ms & 1.26$\times$ & 1.58$\times$ & 1.60$\times$ & 31.178~ms & 1.73$\times$ & 1.75$\times$ & 1.79$\times$ \\
            \bottomrule
        \end{tabular}
        \begin{tablenotes}
          \small
          \item[$\dagger$] indicates \texttt{reduce-overhead} compilation mode, which removes kernel launch overhead.
        \end{tablenotes}
    \end{threeparttable}
\end{table}

\begin{table}[t]
    \centering
    \caption{\textbf{GPU-friendly tiling can affect accuracy at high sparsity.} Top-1 k-NN accuracy on ImageNet-1k and transfer datasets for pruned AugReg ViT-B/16, with unconstrained pruning, i.e., $A=1$, or pruning constrained to remove MLP neurons in blocks of 8 or 64. Superscripts indicate the delta relative to $A=1$. Tiling costs little at up to moderate sparsity, but at 60\% sparsity pruning in blocks of 8 and 64 trails unconstrained pruning by 3.4 and 3.2 points on ImageNet-1k, respectively.}
    \label{tab:tiling-accuracy}
    \begin{tabular}{lcccccc}
      \toprule
       & \multicolumn{3}{c}{ImageNet-1k} & \multicolumn{3}{c}{Transfer Average} \\
      \cmidrule(lr){2-4}\cmidrule(lr){5-7}
      GFLOPs & A=1 & A=8 & A=64 & A=1 & A=8 & A=64 \\
      \midrule
      31.7 & \textbf{82.6} & 82.5$^{\deltaneg{-0.1}}$ & 82.4$^{\deltaneg{-0.2}}$ & \textbf{78.9} & \textbf{78.9}$^{\deltapos{+0.0}}$ & 78.8$^{\deltaneg{-0.1}}$ \\
      28.2 & \textbf{80.1} & 79.6$^{\deltaneg{-0.5}}$ & 79.5$^{\deltaneg{-0.6}}$ & \textbf{76.9} & 76.7$^{\deltaneg{-0.2}}$ & 76.7$^{\deltaneg{-0.2}}$ \\
      24.8 & \textbf{75.7} & 75.4$^{\deltaneg{-0.3}}$ & 75.4$^{\deltaneg{-0.3}}$ & \textbf{74.1} & 73.4$^{\deltaneg{-0.7}}$ & 73.3$^{\deltaneg{-0.8}}$ \\
      21.2 & \textbf{69.1} & \textbf{69.1}$^{\deltapos{+0.0}}$ & \textbf{69.1}$^{\deltapos{+0.0}}$ & 66.8 & \textbf{69.7}$^{\deltapos{+2.9}}$ & 69.3$^{\deltapos{+2.5}}$ \\
      17.8 & \textbf{59.3} & 58.6$^{\deltaneg{-0.7}}$ & 58.2$^{\deltaneg{-1.1}}$ & 63.7 & \textbf{63.8}$^{\deltapos{+0.1}}$ & 63.7$^{\deltapos{+0.0}}$ \\
      14.2 & \textbf{45.2} & 41.8$^{\deltaneg{-3.4}}$ & 42.0$^{\deltaneg{-3.2}}$ & \textbf{55.9} & 55.5$^{\deltaneg{-0.4}}$ & 55.5$^{\deltaneg{-0.4}}$ \\
      \bottomrule
    \end{tabular}         
\end{table}

\begin{table}[t]
    \centering
    \caption{\textbf{Pruning improves throughput.} Throughput of dense and pruned AugReg ViT-B/16 models in fp16 at batch size 128, under eager execution and \texttt{torch.compile}, for unconstrained pruning, i.e., $A=1$, and pruning in blocks of 8 or 64 neurons, reported as speedup over the dense model. In eager mode, ragged widths can be slower than the dense model, while pruning in blocks improves throughput by up to $1.63\times$ over unconstrained pruning at 20\% sparsity. Compilation nearly eliminates this gap, and models pruned to 60\% sparsity reach up to $1.80\times$ dense throughput.}
    \label{tab:tiling-throughput}
    \begin{tabular}{lcccccc}
        \toprule
         & \multicolumn{3}{c}{Eager} & \multicolumn{3}{c}{Compiled} \\
        \cmidrule(lr){2-4}\cmidrule(lr){5-7}
        GFLOPs & A=1 & A=8 & A=64 & A=1 & A=8 & A=64 \\
        \midrule
        Dense (35.1) & \multicolumn{3}{c}{3572~samples/s} & \multicolumn{3}{c}{4143~samples/s} \\
        \midrule
        31.7 & 0.75$\times$ & 1.03$\times$ & 1.06$\times$ & 1.03$\times$ & 1.07$\times$ & 1.08$\times$ \\
        28.2 & 0.70$\times$ & 1.10$\times$ & 1.14$\times$ & 1.11$\times$ & 1.16$\times$ & 1.17$\times$ \\
        24.8 & 0.91$\times$ & 1.19$\times$ & 1.23$\times$ & 1.24$\times$ & 1.27$\times$ & 1.28$\times$ \\
        21.2 & 1.03$\times$ & 1.30$\times$ & 1.34$\times$ & 1.35$\times$ & 1.39$\times$ & 1.42$\times$ \\
        17.8 & 1.16$\times$ & 1.44$\times$ & 1.47$\times$ & 1.53$\times$ & 1.56$\times$ & 1.59$\times$ \\
        14.2 & 1.27$\times$ & 1.59$\times$ & 1.61$\times$ & 1.74$\times$ & 1.77$\times$ & 1.80$\times$ \\
    \bottomrule
    \end{tabular}
\end{table}

\subsection{Runtimes}

\begin{table}[t]
    \centering
    \caption{\textbf{Our method is competitive on runtime.} Wall-clock time to prune models to all six sparsity targets evaluated in this paper, on a single NVIDIA RTX A6000. Our method and SnapViT produce any-budget models in a single run, OPTIN reuses its leave-one-out search across targets, and all other methods must be run end-to-end per target.}
    \label{tab:pruning-cost}
    \begin{tabular}{l r@{\,}l r@{\,}l r@{\,}l r@{\,}l}
        \toprule
        Method & \multicolumn{2}{c}{ViT-S/16} & \multicolumn{2}{c}{ViT-B/16}
               & \multicolumn{2}{c}{ViT-L/16} & \multicolumn{2}{c}{ViT-H/14} \\
        \midrule
        SparseGPT         & 4.5  & min & 5.0  & min & 11   & min & 20    & min \\
        FPTP              & 5.6  & min & 9.0  & min & 19   & min & 35.6  & min \\
        OPTIN             & 12.5 & min & 1.0 & h   & 8.0 & h   & 34.6 & h \\
        VBP               & 2.6 & h   & 6.6 & h   & 20.6 & h  & 53.7 & h \\
        SnapViT           & 2.2 & min & 4.0  & min & 11.2 & min & 28.8  & min \\
        Ours              & 8.7 & min & 13.4 & min & 24.5 & min & 40.7 & min \\
        \bottomrule
    \end{tabular}
\end{table}

In~\autoref{tab:pruning-cost}, we compare the runtime of our approach against the baselines, measured as the total wall-clock time to produce all six sparsity targets evaluated in this paper, on a machine with one NVIDIA RTX A6000 with 48 GB of VRAM, 16 CPU threads of an AMD EPYC 7402P and 40 GB of RAM. Our method scales sub-linearly with parameter count, requiring 8.7 minutes for a ViT-S/16 and 40.7 minutes for ViT-H/14, a $4.7\times$ increase for a $29\times$ larger model. While SparseGPT, SnapViT and FPTP are faster, all four methods, including ours, operate in the same minutes regime. In contrast, OPTIN and VBP scale poorly, requiring 34.6 and 53.7 hours respectively for ViT-H/14, over $50\times$ more than our approach, even under the assumption that OPTIN's leave-one-out search is computed once and reused across all targets.

\subsection{Baselines}
\label{sec:appendix-baselines}

\noindent\textbf{FPTP~\citep{fptp}.} Also known as ``Fast Post-Training Pruning'', FPTP is a structured pruning framework, originally designed for encoder language models, which, similarly to our method, prunes neurons and whole attention heads. The method derives initial importance scores via a diagonal approximation of the Fisher Information Matrix $\mathcal{I}$, producing an initial FLOP- or latency-aware pruning mask for neurons and attention heads. Then, on a per layer basis, they compute the block diagonal Fisher, including off-diagonal entries, and search the initial mask with a greedy algorithm, with the goal of minimizing the expected loss increase under pruning. Finally, the masks of surviving units are relaxed to real values and tuned via least squares to minimize the per-layer output reconstruction error. We use the official implementation\footnote{\url{https://github.com/WoosukKwon/retraining-free-pruning}}, extend it to support ViTs, and set the number of samples used for estimating the Fisher matrix to 1000, for a fair comparison with other methods.

\noindent\textbf{SparseGPT~\citep{frantar2023sparsegpt}.} SparseGPT is a one shot pruning and weight correction algorithm from the OBD/OBS~\citep{lecun1989optimal, hassibi1992second} lineage, originally designed to perform unstructured and semi-structured pruning for large language models. We use the official code implementation\footnote{\url{https://github.com/IST-DASLab/sparsegpt}}, extend it to support structured pruning and ViTs, and report results with and without weight correction. We adopt the official hyperparameter choices, except for the number of samples used to estimate the curvature, for which we use 1000 for a fair comparison with other methods.

\noindent\textbf{OPTIN~\citep{optin}.} Also known as ``One-shot Pruning Technique for Interchangeable Networks'', is a retraining- and correction-free structured pruning approach for transformers and CNNs. It scores each attention head and MLP neuron by masking it and measuring the induced discrepancy in all subsequent blocks' embeddings via manifold distillation and in the output logits via the KL divergence, followed by mask search as in FPTP. The authors propose $\text{OPTIN}_\beta$, which prunes only weights, and $\text{OPTIN}_\tau$, which includes a token-reduction schedule. We prune models with the official implementation\footnote{\url{https://github.com/Skhaki18/optin-transformer-pruning}} and the default hyperparameters for the $\text{OPTIN}_\beta$ configuration. The leave-one-out search uses 32 calibration samples.

\noindent\textbf{VBP~\citep{berisha2025variance}.} Variance Based Pruning is a structured pruning method for ViTs, which prunes the MLP neurons with the lowest measured variance over a calibration set. The mean activation of pruned neurons is then folded in the MLP output projection bias to compensate for their removal. The authors report results for retraining-free pruning and after 10 epochs of ImageNet-1k fine-tuning. For a fair comparison with other methods, we only report retraining-free results. The calibration set used to collect activation means and variances is the entire 1.28M samples ImageNet-1k training set. We prune models using the official implementation\footnote{\url{https://github.com/boschresearch/variance-based-pruning}} and the default hyperparameters.

\noindent\textbf{SnapViT~\citep{simoncini2026elastic}.} SnapViT is a method to elastify pre-trained ViTs. The authors first estimate the local loss curvature as the Hessian diagonal using squared gradients obtained from a DINO loss, and then estimate cross-structure interactions, approximating the Hessian off-diagonal components, via evolutionary search, which minimizes the divergence in cosine similarity between the dense and pruned models. The search then yields separate global rankings of MLP neurons and attention heads, which can be pruned in order to realize any target sparsity. We implemented SnapViT in our framework following the official implementation\footnote{\url{https://github.com/WalterSimoncini/snapvit}}, and use the same configuration recommended by the authors for producing elastic models.

\noindent\textbf{SlimGPT~\citep{ling2024slimgpt}.} SlimGPT is a structured pruning method for large language models from the OBD/OBS~\citep{lecun1989optimal, hassibi1992second} lineage. It greedily removes attention heads and MLP neurons using inverse-Hessian estimates of layerwise output reconstruction error, correcting the surviving weights. Heads are removed one by one, while neurons are removed in progressively smaller groups, to balance efficiency and accuracy. Furthermore, the per-block pruning ratio is progressively increased with depth on a logarithmic ramp to limit error accumulation. We evaluate SlimGPT without fine-tuning, reporting uncorrected results in the main text and corrected results in Appendix~\ref{sec:appendix-weight-correction}. We use the official implementation with the author-recommended defaults, except for calibration and Hessian damping. We calibrate on MS MARCO using 256 sequences of 2048 tokens for decoder-based models and 1024 sequences of 512 tokens for bidirectional embedders, respecting the latter's maximum sequence length. Both configurations use the same total token budget, approximately matching that of our method. We add 1\% of the mean Hessian diagonal to each diagonal entry for BGE-S, E5-Mistral 7B, Qwen3 4B, and Qwen3 8B, and for E5-B only when evaluating without weight correction, to ensure the Hessian is positive definite.

\noindent\textbf{EffiR~\citep{lei2026making}.} EffiR compresses large language models for dense retrieval by dropping MLP sublayers and pruning neurons in those retained. Sublayers are ranked by the cosine distance between their inputs and outputs after residual addition. After layer dropping, neuron importance scores are learned on MS MARCO using an InfoNCE loss and a sparsity penalty. The scores are then ranked globally to determine a fixed pruning mask, after which the model undergoes one epoch of LoRA fine-tuning on MS MARCO using an InfoNCE loss and ranking distillation from a BGE reranker. We use the authors' reported BEIR-13 results for two comparisons: layer dropping without retraining on E5-Mistral 7B in~\autoref{tab:effir-drop}, and pruning followed by retrieval fine-tuning on Mistral 7B in~\autoref{tab:effir-tuned}. The latter includes both layer dropping and the full coarse-to-fine pipeline.

\subsection{Additional Ablations}
\label{sec:appendix-ablations}

\subsubsection{Task-aligned Calibration}

\begin{table}[t]
  \centering
  \caption{\textbf{Task-aligned calibration improves performance.} Top-1 k-NN accuracy on transfer datasets for AugReg ViT-B/16 pruned with our method, using rankings optimized on ImageNet-1k or the target task's training split. Task-aligned calibration improves 25 of 27 settings, gaining up to 26.6\% on Flowers 102 at 40\% sparsity and 20.8\% on Pets at 60\%.}
  \label{tab:augreg-refine-calibration-target}
  \resizebox{\columnwidth}{!}{
    \begin{tabular}{lccccccc}
      \toprule
      & & \multicolumn{6}{c}{Sparsity} \\
      \cmidrule(lr){3-8}
       & & \multicolumn{2}{c}{20\% / 28.2 G} & \multicolumn{2}{c}{40\% / 21.2 G} & \multicolumn{2}{c}{60\% / 14.2 G} \\
      \cmidrule(lr){3-4}\cmidrule(lr){5-6}\cmidrule(lr){7-8}
      Dataset & Dense & ImageNet & Task & ImageNet & Task & ImageNet & Task \\
      \midrule
      CIFAR 10 & 97.3 & 93.5 & \textbf{97.0}$^{\phantom{0}\deltapos{+3.5}}$ & 82.4 & \textbf{96.1}$^{\deltapos{+13.7}}$ & 69.5 & \textbf{86.8}$^{\deltapos{+17.3}}$ \\
      CIFAR 100 & 85.0 & 75.5 & \textbf{83.8}$^{\phantom{0}\deltapos{+8.3}}$ & 58.1 & \textbf{78.0}$^{\deltapos{+19.9}}$ & 43.3 & \textbf{56.5}$^{\deltapos{+13.2}}$ \\
      Aircraft & 24.6 & \textbf{26.7} & 25.0$^{\phantom{0}\deltaneg{-1.7}}$ & 21.4 & \textbf{23.7}$^{\phantom{0}\deltapos{+2.3}}$ & 12.5 & \textbf{16.7}$^{\phantom{0}\deltapos{+4.2}}$ \\
      EuroSAT & 92.6 & 92.1 & \textbf{93.3}$^{\phantom{0}\deltapos{+1.2}}$ & 91.6 & \textbf{94.2}$^{\phantom{0}\deltapos{+2.6}}$ & 84.4 & \textbf{90.7}$^{\phantom{0}\deltapos{+6.3}}$ \\
      Flowers 102 & 97.6 & 80.1 & \textbf{97.4}$^{\deltapos{+17.3}}$ & 67.5 & \textbf{94.1}$^{\deltapos{+26.6}}$ & 52.8 & \textbf{67.5}$^{\deltapos{+14.7}}$ \\
      Food 101 & 85.1 & 77.9 & \textbf{84.4}$^{\phantom{0}\deltapos{+6.5}}$ & 62.5 & \textbf{80.1}$^{\deltapos{+17.6}}$ & 36.7 & \textbf{49.1}$^{\deltapos{+12.4}}$ \\
      Pets & 92.3 & 91.2 & \textbf{91.7}$^{\phantom{0}\deltapos{+0.5}}$ & 86.7 & \textbf{90.2}$^{\phantom{0}\deltapos{+3.5}}$ & 58.3 & \textbf{79.1}$^{\deltapos{+20.8}}$ \\
      RESISC 45 & 84.2 & 81.4 & \textbf{83.9}$^{\phantom{0}\deltapos{+2.5}}$ & 79.0 & \textbf{82.7}$^{\phantom{0}\deltapos{+3.7}}$ & \textbf{69.1} & 68.7$^{\phantom{0}\deltaneg{-0.4}}$ \\
      Textures & 67.5 & 63.5 & \textbf{66.3}$^{\phantom{0}\deltapos{+2.8}}$ & 55.4 & \textbf{60.4}$^{\phantom{0}\deltapos{+5.0}}$ & 37.3 & \textbf{44.5}$^{\phantom{0}\deltapos{+7.2}}$ \\
      \bottomrule
    \end{tabular}      
  }
\end{table}

Our default ImageNet-1k calibration data may not represent the downstream task's data distribution. We therefore test whether calibrating on the downstream task's training split improves pruning performance, optimizing rankings by minimizing the cosine distance between pruned and dense features. \autoref{tab:augreg-refine-calibration-target} shows that task-aligned calibration improves performance in 25 of the 27 settings shown, with the largest gains at moderate-to-high sparsity, e.g., 26.6\% on Flowers 102 at 40\% sparsity and 20.8\% on Pets at 60\%. Task-specific calibration can therefore further improve pruning performance when downstream data is available.

\subsubsection{Choice of Ranking Optimization Loss}
\label{sec:appendix-refinement-loss}

Minimizing the cosine distance between pruned and dense embeddings is applicable to any encoder, but preserving dense embeddings may not fully align with downstream performance. Differentiable soft pruning lets us instead optimize unit scores using the model's native training objective or a close surrogate. We test whether this choice improves downstream accuracy, keeping the DINO-based initial scores fixed. For AugReg ViT-B/16, we compare cosine-distance alignment with cross-entropy classification, and for SigLIPv2 ViT-B/16, we compare it with the sigmoid image-text loss term. We report top-1 k-NN accuracy on ImageNet-1k and nine transfer datasets at 20\%, 40\%, and 60\% sparsity in~\autoref{tab:refine-native-loss-augreg} and~\autoref{tab:refine-native-loss-siglip}.

\begin{table}[htbp]
  \centering
  \caption{\textbf{Optimizing rankings with the backbone's native loss improves pruning performance.} Top-1 k-NN accuracy for AugReg ViT-B/16 pruned with our method, with rankings optimized by cosine-distance alignment to dense features (Cosine) or with the native cross-entropy loss (Native). Column groups correspond to 20\%, 40\%, and 60\% sparsity. The native loss gains up to 13.5\% on Pets at 60\% sparsity, but trails cosine by up to 6.2\% on CIFAR 100 at 40\%.}
  \label{tab:refine-native-loss-augreg}
  \begin{tabular}{lcccccc}
    \toprule
    & \multicolumn{6}{c}{Sparsity} \\
    \cmidrule(lr){2-7}
     & \multicolumn{2}{c}{20\% / 28.2 G} & \multicolumn{2}{c}{40\% / 21.2 G} & \multicolumn{2}{c}{60\% / 14.2 G} \\
    \cmidrule(lr){2-3}\cmidrule(lr){4-5}\cmidrule(lr){6-7}
    Dataset & Cosine & Native & Cosine & Native & Cosine & Native \\
    \midrule
    ImageNet-1k & 79.2 & \textbf{80.1}$^{\phantom{0}\deltapos{+0.9}}$ & 64.2 & \textbf{69.1}$^{\phantom{0}\deltapos{+4.9}}$ & 36.3 & \textbf{45.2}$^{\phantom{0}\deltapos{+8.9}}$ \\
    CIFAR 10 & 92.9 & \textbf{95.1}$^{\phantom{0}\deltapos{+2.2}}$ & \textbf{81.8} & 76.2$^{\phantom{0}\deltaneg{-5.6}}$ & 66.7 & \textbf{68.7}$^{\phantom{0}\deltapos{+2.0}}$ \\
    CIFAR 100 & 74.1 & \textbf{79.6}$^{\phantom{0}\deltapos{+5.5}}$ & \textbf{59.6} & 53.4$^{\phantom{0}\deltaneg{-6.2}}$ & 41.3 & \textbf{43.0}$^{\phantom{0}\deltapos{+1.7}}$ \\
    Aircraft & 24.9 & \textbf{27.3}$^{\phantom{0}\deltapos{+2.4}}$ & 20.1 & \textbf{23.1}$^{\phantom{0}\deltapos{+3.0}}$ & 12.7 & \textbf{19.6}$^{\phantom{0}\deltapos{+6.9}}$ \\
    EuroSAT & \textbf{92.3} & 91.6$^{\phantom{0}\deltaneg{-0.7}}$ & \textbf{91.6} & 90.2$^{\phantom{0}\deltaneg{-1.4}}$ & 83.3 & \textbf{84.7}$^{\phantom{0}\deltapos{+1.4}}$ \\
    Flowers 102 & 80.1 & \textbf{82.0}$^{\phantom{0}\deltapos{+1.9}}$ & 69.7 & \textbf{70.4}$^{\phantom{0}\deltapos{+0.7}}$ & 52.9 & \textbf{57.7}$^{\phantom{0}\deltapos{+4.8}}$ \\
    Food 101 & \textbf{79.5} & 78.7$^{\phantom{0}\deltaneg{-0.8}}$ & 61.4 & \textbf{64.8}$^{\phantom{0}\deltapos{+3.4}}$ & 37.8 & \textbf{41.8}$^{\phantom{0}\deltapos{+4.0}}$ \\
    Pets & 91.1 & \textbf{91.3}$^{\phantom{0}\deltapos{+0.2}}$ & 84.0 & \textbf{89.0}$^{\phantom{0}\deltapos{+5.0}}$ & 58.3 & \textbf{71.8}$^{\deltapos{+13.5}}$ \\
    RESISC 45 & 81.3 & \textbf{82.8}$^{\phantom{0}\deltapos{+1.5}}$ & \textbf{80.0} & 78.1$^{\phantom{0}\deltaneg{-1.9}}$ & 67.8 & \textbf{74.4}$^{\phantom{0}\deltapos{+6.6}}$ \\
    Textures & \textbf{64.4} & 64.1$^{\phantom{0}\deltaneg{-0.3}}$ & 53.7 & \textbf{55.7}$^{\phantom{0}\deltapos{+2.0}}$ & 37.1 & \textbf{40.9}$^{\phantom{0}\deltapos{+3.8}}$ \\
    \bottomrule
  \end{tabular}
\end{table}

\begin{table}[htbp]
  \centering
  \caption{\textbf{Optimizing rankings with the backbone's native loss improves pruning performance.} Top-1 k-NN accuracy for SigLIPv2 ViT-B/16 pruned with our method, with rankings optimized by cosine-distance alignment to dense features (Cosine) or with the native sigmoid image-text loss (Native). Column groups correspond to 20\%, 40\%, and 60\% sparsity. The native loss gains up to 19.0\% on Pets at 60\% sparsity, but trails cosine by up to 2.9\% on EuroSAT at 40\%.}
  \label{tab:refine-native-loss-siglip}
  \begin{tabular}{lcccccc}
    \toprule
    & \multicolumn{6}{c}{Sparsity} \\
    \cmidrule(lr){2-7}
     & \multicolumn{2}{c}{20\% / 28.2 G} & \multicolumn{2}{c}{40\% / 21.2 G} & \multicolumn{2}{c}{60\% / 14.3 G} \\
    \cmidrule(lr){2-3}\cmidrule(lr){4-5}\cmidrule(lr){6-7}
    Dataset & Cosine & Native & Cosine & Native & Cosine & Native \\
    \midrule
    ImageNet-1k & 74.7 & \textbf{76.1}$^{\phantom{0}\deltapos{+1.4}}$ & 53.5 & \textbf{61.1}$^{\phantom{0}\deltapos{+7.6}}$ & 20.8 & \textbf{30.9}$^{\deltapos{+10.1}}$ \\
    CIFAR 10 & 92.6 & \textbf{95.0}$^{\phantom{0}\deltapos{+2.4}}$ & 67.4 & \textbf{71.6}$^{\phantom{0}\deltapos{+4.2}}$ & 52.3 & \textbf{57.1}$^{\phantom{0}\deltapos{+4.8}}$ \\
    CIFAR 100 & 72.9 & \textbf{77.3}$^{\phantom{0}\deltapos{+4.4}}$ & 42.6 & \textbf{46.3}$^{\phantom{0}\deltapos{+3.7}}$ & 26.7 & \textbf{31.2}$^{\phantom{0}\deltapos{+4.5}}$ \\
    Aircraft & 34.6 & \textbf{46.5}$^{\deltapos{+11.9}}$ & 15.1 & \textbf{23.3}$^{\phantom{0}\deltapos{+8.2}}$ & 8.6 & \textbf{13.3}$^{\phantom{0}\deltapos{+4.7}}$ \\
    EuroSAT & 90.9 & \textbf{91.9}$^{\phantom{0}\deltapos{+1.0}}$ & \textbf{89.6} & 86.7$^{\phantom{0}\deltaneg{-2.9}}$ & \textbf{81.6} & 80.8$^{\phantom{0}\deltaneg{-0.8}}$ \\
    Flowers 102 & \textbf{77.7} & 74.9$^{\phantom{0}\deltaneg{-2.8}}$ & 55.4 & \textbf{66.5}$^{\deltapos{+11.1}}$ & 39.6 & \textbf{48.9}$^{\phantom{0}\deltapos{+9.3}}$ \\
    Food 101 & 83.2 & \textbf{83.9}$^{\phantom{0}\deltapos{+0.7}}$ & 52.3 & \textbf{59.5}$^{\phantom{0}\deltapos{+7.2}}$ & 20.7 & \textbf{32.0}$^{\deltapos{+11.3}}$ \\
    Pets & 87.7 & \textbf{89.8}$^{\phantom{0}\deltapos{+2.1}}$ & 68.1 & \textbf{80.8}$^{\deltapos{+12.7}}$ & 25.4 & \textbf{44.4}$^{\deltapos{+19.0}}$ \\
    RESISC 45 & 87.5 & \textbf{88.7}$^{\phantom{0}\deltapos{+1.2}}$ & 75.8 & \textbf{80.5}$^{\phantom{0}\deltapos{+4.7}}$ & 61.3 & \textbf{66.0}$^{\phantom{0}\deltapos{+4.7}}$ \\
    Textures & 66.0 & \textbf{69.1}$^{\phantom{0}\deltapos{+3.1}}$ & 51.1 & \textbf{58.1}$^{\phantom{0}\deltapos{+7.0}}$ & 31.4 & \textbf{38.3}$^{\phantom{0}\deltapos{+6.9}}$ \\
    \bottomrule
  \end{tabular}
\end{table}

The native loss improves ImageNet-1k accuracy at every evaluated sparsity for both backbones, with gains increasing from 0.9\% to 8.9\% for AugReg and from 1.4\% to 10.1\% for SigLIPv2 as sparsity increases from 20\% to 60\%. It also generally performs better on transfer datasets, most notably on Pets at 60\% sparsity, where it gains 19.0\% for SigLIPv2 and 13.5\% for AugReg. The largest regressions among the reported targets occur on EuroSAT for SigLIPv2, by 2.9\% at 40\% sparsity, and on CIFAR 100 for AugReg, by 6.2\% at 40\% sparsity. These results support using the native loss when available, with gains extending across sparsity levels and most transfer settings.

\subsubsection{Calibration Set Size}
\label{sec:appendix-calibration-size}

\begin{table}[htbp]
    \centering
    \caption{\textbf{More unique calibration samples generally improve ranking optimization.} Top-1 k-NN accuracy on ImageNet-1k for AugReg ViT-B/16 pruned with our method, varying the number of unique ImageNet-1k samples available to ranking optimization under fixed compute, i.e., 300 steps, at batch size 32. Superscripts give the change relative to 64 samples, and $\dagger$ marks our default. Using 9600 samples gains up to 14.3 points over 64 samples at 50\% sparsity, and up to 2.6 points over the default at 40\%, but trails it by 2.0 points at 60\%.}
    \label{tab:refine-num-samples}
    \resizebox{\columnwidth}{!}{
        \begin{threeparttable}
          \begin{tabular}{lccccccccc}
            \toprule
             & \multicolumn{9}{c}{Unique Samples} \\
            \cmidrule(lr){2-10}
            GFLOPs & 64 & 128 & 256 & 512 & 1000 & 2000\tnote{$\dagger$} & 4000 & 8000 & 9600 \\
            \midrule
            31.7 & 81.8 & 81.8$^{\phantom{0}\deltapos{+0.0}}$ & 82.1$^{\phantom{0}\deltapos{+0.3}}$ & 82.2$^{\phantom{0}\deltapos{+0.4}}$ & 82.5$^{\phantom{0}\deltapos{+0.7}}$ & 82.6$^{\phantom{0}\deltapos{+0.8}}$ & 82.9$^{\phantom{0}\deltapos{+1.1}}$ & 82.9$^{\phantom{0}\deltapos{+1.1}}$ & \textbf{83.0}$^{\phantom{0}\deltapos{+1.2}}$ \\
            28.2 & 77.4 & 77.6$^{\phantom{0}\deltapos{+0.2}}$ & 78.1$^{\phantom{0}\deltapos{+0.7}}$ & 78.6$^{\phantom{0}\deltapos{+1.2}}$ & 79.1$^{\phantom{0}\deltapos{+1.7}}$ & 80.1$^{\phantom{0}\deltapos{+2.7}}$ & 80.5$^{\phantom{0}\deltapos{+3.1}}$ & 81.0$^{\phantom{0}\deltapos{+3.6}}$ & \textbf{81.1}$^{\phantom{0}\deltapos{+3.7}}$ \\
            24.8 & 70.3 & 71.3$^{\phantom{0}\deltapos{+1.0}}$ & 71.4$^{\phantom{0}\deltapos{+1.1}}$ & 72.8$^{\phantom{0}\deltapos{+2.5}}$ & 74.1$^{\phantom{0}\deltapos{+3.8}}$ & 75.7$^{\phantom{0}\deltapos{+5.4}}$ & 77.1$^{\phantom{0}\deltapos{+6.8}}$ & 77.6$^{\phantom{0}\deltapos{+7.3}}$ & \textbf{77.7}$^{\phantom{0}\deltapos{+7.4}}$ \\
            21.2 & 59.4 & 62.1$^{\phantom{0}\deltapos{+2.7}}$ & 63.0$^{\phantom{0}\deltapos{+3.6}}$ & 65.2$^{\phantom{0}\deltapos{+5.8}}$ & 67.6$^{\phantom{0}\deltapos{+8.2}}$ & 69.1$^{\phantom{0}\deltapos{+9.7}}$ & 70.7$^{\deltapos{+11.3}}$ & 71.1$^{\deltapos{+11.7}}$ & \textbf{71.7}$^{\deltapos{+12.3}}$ \\
            17.8 & 46.6 & 49.2$^{\phantom{0}\deltapos{+2.6}}$ & 51.7$^{\phantom{0}\deltapos{+5.1}}$ & 55.1$^{\phantom{0}\deltapos{+8.5}}$ & 57.2$^{\deltapos{+10.6}}$ & 59.3$^{\deltapos{+12.7}}$ & 59.8$^{\deltapos{+13.2}}$ & 60.2$^{\deltapos{+13.6}}$ & \textbf{60.9}$^{\deltapos{+14.3}}$ \\
            14.2 & 30.2 & 32.5$^{\phantom{0}\deltapos{+2.3}}$ & 35.4$^{\phantom{0}\deltapos{+5.2}}$ & 38.7$^{\phantom{0}\deltapos{+8.5}}$ & 43.7$^{\deltapos{+13.5}}$ & \textbf{45.2}$^{\deltapos{+15.0}}$ & 42.9$^{\deltapos{+12.7}}$ & 43.9$^{\deltapos{+13.7}}$ & 43.2$^{\deltapos{+13.0}}$ \\
            \bottomrule
            \end{tabular}          
            \begin{tablenotes}
              \small
              \item[$\dagger$] standard recipe.
            \end{tablenotes}
        \end{threeparttable}
    }
\end{table}

Optimizing on a small calibration set may limit how well the ranking generalizes. We therefore vary the number of unique calibration samples while keeping the optimization compute fixed. In~\autoref{tab:refine-num-samples}, we use ImageNet-1k samples and run 300 steps at batch size 32, corresponding to a total of 9600 observed samples. Increasing the number of samples generally improves performance. Using 9600 unique samples improves accuracy by up to 14.3\% at 50\% sparsity compared to using only 64 samples. Relative to the standard recipe of 2000 samples, using 9600 samples improves accuracy by up to 2.6\% at 40\% sparsity, but trails it by 2.0\% at 60\% sparsity. Increasing calibration diversity improves accuracy over very small sets, but expanding beyond our default 2000 samples does not consistently improve performance.

\subsubsection{Unified importance ranking}
\label{sec:appendix-unified-ranking}

A unified ranking lets optimization allocate sparsity across blocks and unit types without a fixed neuron-to-head pruning schedule. We test how this flexibility affects downstream performance compared with separate rankings and more constrained thresholding. We compare it with a fixed schedule of paired neuron and head pruning fractions, following the approach of SnapViT~\citep{simoncini2026elastic}. We optimize this schedule using either cross-block thresholding, with one threshold per unit type across all blocks, or per-block thresholding, with separate thresholds for neurons and heads within each block. At evaluation, both alternatives use separate global rankings for MLP neurons and attention heads. All three approaches optimize over 14 settings. Our unified approach uses sparsity targets $b$ from 5\% to 70\% in 5\% increments. For the alternatives, the schedule maps each target $b$ to neuron and head pruning fractions of $b + 0.05$ and $\max(0, b - 0.10)$, respectively. We evaluate sparsity targets from 10\% to 60\% in 10\% increments, interpolating both neuron and head pruning fractions between anchors where needed to match computational costs.

We report top-1 k-NN accuracy on ImageNet-1k and transfer datasets for AugReg ViT-B/16, DINOv3 ViT-B/16, and SigLIPv2 ViT-B/16 in~\autoref{tab:unified-ranking-aureg},~\autoref{tab:unified-ranking-dinov3}, and~\autoref{tab:unified-ranking-siglipv2}, respectively. Cross-block thresholding generally improves on per-block thresholding, with the largest ImageNet-1k gain of 34.9\% for DINOv3 at 40\% sparsity. The impact of additionally unifying neurons and heads varies across backbones. Unified ranking performs similarly to cross-block thresholding on AugReg and improves ImageNet-1k accuracy by up to 8.7\% on DINOv3 at 50\% sparsity. On SigLIPv2, it generally trails cross-block thresholding, by up to 2.5\% on the transfer average at 40\% sparsity. Nonetheless, a unified ranking lets end users specify a single sparsity target and removes the need for a hand-crafted allocation schedule.

\begin{table}[htbp]
  \centering
  \caption{\textbf{Unified ranking matches cross-block thresholding on AugReg.} Top-1 k-NN accuracy on ImageNet-1k and transfer datasets for AugReg ViT-B/16 pruned with our method, with separate neuron and head rankings under per-block or cross-block thresholding, or our default unified ranking. Superscripts indicate the delta relative to per-block.}
  \label{tab:unified-ranking-aureg}
  \resizebox{\columnwidth}{!}{
    \begin{tabular}{lcccccc}
      \toprule
      & \multicolumn{6}{c}{Sparsity} \\
      \cmidrule(lr){2-7}
      Threshold & 10\% / 31.7 G & 20\% / 28.2 G & 30\% / 24.8 G & 40\% / 21.2 G & 50\% / 17.8 G & 60\% / 14.2 G \\
      \midrule
      \multicolumn{7}{l}{\textit{ImageNet-1k}} \\
      Per-block & 82.5 & 79.7 & 74.3 & 65.2 & 54.0 & 40.6 \\
      Cross-block & 82.5\makebox[0pt][l]{$^{\deltapos{+0.0}}$} & \textbf{80.2}\makebox[0pt][l]{$^{\deltapos{+0.5}}$} & \textbf{76.2}\makebox[0pt][l]{$^{\deltapos{+1.9}}$} & \textbf{69.6}\makebox[0pt][l]{$^{\deltapos{+4.4}}$} & 57.2\makebox[0pt][l]{$^{\deltapos{+3.2}}$} & \textbf{45.7}\makebox[0pt][l]{$^{\deltapos{+5.1}}$} \\
      Unified & \textbf{82.6}\makebox[0pt][l]{$^{\deltapos{+0.1}}$} & 80.1\makebox[0pt][l]{$^{\deltapos{+0.4}}$} & 75.7\makebox[0pt][l]{$^{\deltapos{+1.4}}$} & 69.1\makebox[0pt][l]{$^{\deltapos{+3.9}}$} & \textbf{59.3}\makebox[0pt][l]{$^{\deltapos{+5.3}}$} & 45.2\makebox[0pt][l]{$^{\deltapos{+4.6}}$} \\
      \midrule
      \multicolumn{7}{l}{\textit{Transfer Average}} \\
      Per-block & 78.7 & 76.0 & 71.6 & 65.5 & 60.3 & 52.9 \\
      Cross-block & \textbf{79.1}\makebox[0pt][l]{$^{\deltapos{+0.4}}$} & 76.8\makebox[0pt][l]{$^{\deltapos{+0.8}}$} & 73.8\makebox[0pt][l]{$^{\deltapos{+2.2}}$} & \textbf{68.7}\makebox[0pt][l]{$^{\deltapos{+3.2}}$} & 60.6\makebox[0pt][l]{$^{\deltapos{+0.3}}$} & 55.8\makebox[0pt][l]{$^{\deltapos{+2.9}}$} \\
      Unified & 78.9\makebox[0pt][l]{$^{\deltapos{+0.2}}$} & \textbf{76.9}\makebox[0pt][l]{$^{\deltapos{+0.9}}$} & \textbf{74.1}\makebox[0pt][l]{$^{\deltapos{+2.5}}$} & 66.8\makebox[0pt][l]{$^{\deltapos{+1.3}}$} & \textbf{63.7}\makebox[0pt][l]{$^{\deltapos{+3.4}}$} & \textbf{55.9}\makebox[0pt][l]{$^{\deltapos{+3.0}}$} \\
      \bottomrule
    \end{tabular}
  }
\end{table}

\begin{table}[htbp]
  \centering
  \caption{\textbf{Unified ranking improves DINOv3 performance at high sparsity.} Top-1 k-NN accuracy on ImageNet-1k and transfer datasets for DINOv3 ViT-B/16 pruned with our method, with separate neuron and head rankings under per-block or cross-block thresholding, or our default unified ranking. Superscripts indicate the delta relative to per-block. Cross-block thresholding gains up to 34.9\% over per-block on ImageNet-1k at 40\% sparsity, and unifying neurons and heads adds up to 8.7\% at 50\%.}
  \label{tab:unified-ranking-dinov3}
  \resizebox{\columnwidth}{!}{
    \begin{tabular}{lcccccc}
      \toprule
      & \multicolumn{6}{c}{Sparsity} \\
      \cmidrule(lr){2-7}
      Threshold & 10\% / 41.9 G & 20\% / 37.3 G & 30\% / 32.5 G & 40\% / 28.0 G & 50\% / 23.4 G & 60\% / 18.8 G \\
      \midrule
      \multicolumn{7}{l}{\textit{ImageNet-1k}} \\
      Per-block & 81.4 & 77.8 & 69.3 & 29.9 & 15.5 & 8.0 \\
      Cross-block & \textbf{82.0}\makebox[0pt][l]{$^{\phantom{0}\deltapos{+0.6}}$} & \textbf{80.5}\makebox[0pt][l]{$^{\phantom{0}\deltapos{+2.7}}$} & 72.2\makebox[0pt][l]{$^{\phantom{0}\deltapos{+2.9}}$} & 64.8\makebox[0pt][l]{$^{\deltapos{+34.9}}$} & 47.7\makebox[0pt][l]{$^{\deltapos{+32.2}}$} & 32.1\makebox[0pt][l]{$^{\deltapos{+24.1}}$} \\
      Unified & 81.9\makebox[0pt][l]{$^{\phantom{0}\deltapos{+0.5}}$} & 80.2\makebox[0pt][l]{$^{\phantom{0}\deltapos{+2.4}}$} & \textbf{76.4}\makebox[0pt][l]{$^{\phantom{0}\deltapos{+7.1}}$} & \textbf{69.4}\makebox[0pt][l]{$^{\deltapos{+39.5}}$} & \textbf{56.4}\makebox[0pt][l]{$^{\deltapos{+40.9}}$} & \textbf{38.9}\makebox[0pt][l]{$^{\deltapos{+30.9}}$} \\
      \midrule
      \multicolumn{7}{l}{\textit{Transfer Average}} \\
      Per-block & 85.8 & 79.8 & 72.0 & 47.4 & 37.4 & 30.1 \\
      Cross-block & 86.2\makebox[0pt][l]{$^{\phantom{0}\deltapos{+0.4}}$} & \textbf{84.6}\makebox[0pt][l]{$^{\phantom{0}\deltapos{+4.8}}$} & 75.1\makebox[0pt][l]{$^{\phantom{0}\deltapos{+3.1}}$} & 68.2\makebox[0pt][l]{$^{\deltapos{+20.8}}$} & 58.9\makebox[0pt][l]{$^{\deltapos{+21.5}}$} & 51.6\makebox[0pt][l]{$^{\deltapos{+21.5}}$} \\
      Unified & \textbf{86.8}\makebox[0pt][l]{$^{\phantom{0}\deltapos{+1.0}}$} & 84.2\makebox[0pt][l]{$^{\phantom{0}\deltapos{+4.4}}$} & \textbf{79.8}\makebox[0pt][l]{$^{\phantom{0}\deltapos{+7.8}}$} & \textbf{72.6}\makebox[0pt][l]{$^{\deltapos{+25.2}}$} & \textbf{63.2}\makebox[0pt][l]{$^{\deltapos{+25.8}}$} & \textbf{52.9}\makebox[0pt][l]{$^{\deltapos{+22.8}}$} \\
      \bottomrule
    \end{tabular}
  }
\end{table}

\begin{table}[htbp]
  \centering
  \caption{\textbf{Cross-block thresholding performs best for SigLIPv2.} Top-1 k-NN accuracy on ImageNet-1k and transfer datasets for SigLIPv2 ViT-B/16 pruned with our method, with separate neuron and head rankings under per-block or cross-block thresholding, or our default unified ranking. Superscripts indicate the delta relative to per-block. Unified ranking trails cross-block thresholding by up to 2.5\% on the transfer average at 40\% sparsity.}
  \label{tab:unified-ranking-siglipv2}
  \resizebox{\columnwidth}{!}{
    \begin{tabular}{lcccccc}
      \toprule
      & \multicolumn{6}{c}{Sparsity} \\
      \cmidrule(lr){2-7}
      Threshold & 10\% / 31.7 G & 20\% / 28.2 G & 30\% / 24.7 G & 40\% / 21.2 G & 50\% / 17.7 G & 60\% / 14.3 G \\
      \midrule
      \multicolumn{7}{l}{\textit{ImageNet-1k}} \\
      Per-block & 79.0 & 71.2 & 65.6 & 51.4 & 40.6 & 26.8 \\
      Cross-block & \textbf{79.4}\makebox[0pt][l]{$^{\phantom{0}\deltapos{+0.4}}$} & \textbf{77.2}\makebox[0pt][l]{$^{\phantom{0}\deltapos{+6.0}}$} & \textbf{72.0}\makebox[0pt][l]{$^{\phantom{0}\deltapos{+6.4}}$} & \textbf{63.2}\makebox[0pt][l]{$^{\deltapos{+11.8}}$} & \textbf{48.4}\makebox[0pt][l]{$^{\phantom{0}\deltapos{+7.8}}$} & \textbf{33.8}\makebox[0pt][l]{$^{\phantom{0}\deltapos{+7.0}}$} \\
      Unified & \textbf{79.4}\makebox[0pt][l]{$^{\phantom{0}\deltapos{+0.4}}$} & 76.1\makebox[0pt][l]{$^{\phantom{0}\deltapos{+4.9}}$} & 70.4\makebox[0pt][l]{$^{\phantom{0}\deltapos{+4.8}}$} & 61.1\makebox[0pt][l]{$^{\phantom{0}\deltapos{+9.7}}$} & 46.2\makebox[0pt][l]{$^{\phantom{0}\deltapos{+5.6}}$} & 30.9\makebox[0pt][l]{$^{\phantom{0}\deltapos{+4.1}}$} \\
      \midrule
      \multicolumn{7}{l}{\textit{Transfer Average}} \\
      Per-block & 82.9 & 74.2 & 69.5 & 57.3 & 51.9 & 43.9 \\
      Cross-block & \textbf{84.0}\makebox[0pt][l]{$^{\phantom{0}\deltapos{+1.1}}$} & \textbf{80.8}\makebox[0pt][l]{$^{\phantom{0}\deltapos{+6.6}}$} & \textbf{74.3}\makebox[0pt][l]{$^{\phantom{0}\deltapos{+4.8}}$} & \textbf{66.2}\makebox[0pt][l]{$^{\phantom{0}\deltapos{+8.9}}$} & \textbf{55.8}\makebox[0pt][l]{$^{\phantom{0}\deltapos{+3.9}}$} & \textbf{48.2}\makebox[0pt][l]{$^{\phantom{0}\deltapos{+4.3}}$} \\
      Unified & 83.3\makebox[0pt][l]{$^{\phantom{0}\deltapos{+0.4}}$} & 79.7\makebox[0pt][l]{$^{\phantom{0}\deltapos{+5.5}}$} & 73.7\makebox[0pt][l]{$^{\phantom{0}\deltapos{+4.2}}$} & 63.7\makebox[0pt][l]{$^{\phantom{0}\deltapos{+6.4}}$} & \textbf{55.8}\makebox[0pt][l]{$^{\phantom{0}\deltapos{+3.9}}$} & 45.8\makebox[0pt][l]{$^{\phantom{0}\deltapos{+1.9}}$} \\
      \bottomrule
    \end{tabular}
  }
\end{table}

\subsubsection{Elastic interpolation}
\label{sec:interpolation-ablation}

\begin{table}[htbp]
  \centering
  \caption{\textbf{Our rankings interpolate to unseen sparsity targets.} Top-1 k-NN accuracy on ImageNet-1k and transfer datasets for pruned AugReg ViT-B/16 with a ranking optimized over seven sparsity anchors, from 10\% to 70\% in 10\% increments, and evaluated every 5\%. Targets marked $\dagger$ lie between anchors. Performance degrades smoothly at the evaluated intermediate targets.}
  \label{tab:interpolation-ablation}
  \resizebox{\columnwidth}{!}{
    \begin{threeparttable}
      \begin{tabular}{lccccccccccc}
        \toprule
        & \multicolumn{11}{c}{Sparsity} \\
        \cmidrule(lr){2-12}
         & 10\% & 15\%$\dagger$ & 20\% & 25\%$\dagger$ & 30\% & 35\%$\dagger$ & 40\% & 45\%$\dagger$ & 50\% & 55\%$\dagger$ & 60\% \\
        \midrule
        ImageNet-1k & 82.7 & 81.5 & 79.9 & 78.2 & 75.9 & 73.3 & 69.4 & 64.2 & 59.1 & 52.1 & 43.3 \\
        Transfer Average & 78.9 & 77.8 & 76.8 & 75.6 & 73.3 & 71.0 & 68.6 & 66.2 & 63.5 & 59.3 & 55.2 \\
        \bottomrule
      \end{tabular}        
      \begin{tablenotes}
        \small
        \item[$\dagger$] indicates interpolated sparsities.
      \end{tablenotes}
    \end{threeparttable}
  }
\end{table}

Our ranking optimization procedure optimizes a single ranking over a discrete set of sparsity anchors. In our main experiments, the evaluated sparsity targets coincide with a subset of such anchors. In this section, we instead investigate whether performance interpolates smoothly for sparsity targets off the anchor grid. To do so, we optimize a ranking for an AugReg ViT-B/16 model over seven sparsity targets from 10\% to 70\% in 10\% increments, rather than the default 14, and evaluate models pruned at every 5\% step between 10\% and 60\%. We report the top-1 k-NN accuracy on ImageNet-1k and transfer datasets in~\autoref{tab:interpolation-ablation}. Accuracy decreases monotonically across all 11 evaluation points, with no dip for sparsity targets off the anchor grid: each interpolated target scores within 0.1 points of, or above, the linear interpolation between its neighboring anchors. These results show that the learned ranking maintains smooth performance between optimization anchors, allowing deployment at the evaluated intermediate budgets without further optimization.

\subsubsection{Ranking optimization temperature}
\label{sec:temperature-ablation}

The sigmoid temperature controls how closely soft masks approximate hard pruning and how widely useful gradients extend around each threshold. We test how this tradeoff affects the learned ranking, sweeping $\tau \in \{0.01, 0.05, 0.1, 0.2, 0.5, 1.0, 2.5, 5.0\}$ and reporting top-1 k-NN accuracy on ImageNet-1k and transfer datasets for pruned AugReg ViT-B/16 models in~\autoref{tab:temperature-ablation}.

\begin{table}[t]
  \centering
  \caption{\textbf{Low temperatures improve ranking optimization.} Top-1 k-NN accuracy on ImageNet-1k and transfer datasets for AugReg ViT-B/16 pruned with our method, varying the sigmoid temperature $\tau$ of the ranking optimization. The first row uses plain scores, without ranking optimization. The default setting is highlighted in \recipetag{orange}. Low temperatures perform best, with mean accuracy peaking at $\tau = 0.1$. Mean accuracy decreases steadily above $0.2$ and approaches the unoptimized baseline at $\tau = 5.0$.}
  \label{tab:temperature-ablation}
  \begin{tabular}{lcccc}
    \toprule
    & \multicolumn{3}{c}{Sparsity} \\
    \cmidrule(lr){2-4}
    Temperature & 20\% / 28.2 G & 40\% / 21.2 G & 60\% / 14.2 G & Mean \\
    \midrule
    \multicolumn{5}{l}{\textit{ImageNet-1k}} \\
    No optimization & 78.6 & 42.5 & 17.0 & 46.1 \\
    $\tau$ = 0.01 & 79.7 & 67.7 & 41.6 & 63.0 \\
    $\tau$ = 0.05 & \textbf{80.4} & 69.7 & 44.1 & 64.8 \\
    $\tau$ = 0.1 & 80.3 & \textbf{70.6} & 44.5 & \textbf{65.1} \\
    \rowcolor{recipehighlight} $\tau$ = 0.2 & 80.1 & 69.1 & \textbf{45.2} & 64.8 \\
    $\tau$ = 0.5 & 79.7 & 67.6 & 38.7 & 62.0 \\
    $\tau$ = 1.0 & 78.5 & 63.6 & 33.2 & 58.4 \\
    $\tau$ = 2.5 & 76.1 & 55.9 & 16.5 & 49.5 \\
    $\tau$ = 5.0 & 72.4 & 47.3 & 16.5 & 45.4 \\
    \midrule
    \multicolumn{5}{l}{\textit{Transfer Average}} \\
    No optimization & 74.6 & 54.2 & 39.7 & 56.2 \\
    $\tau$ = 0.01 & 76.9 & 68.2 & 55.7 & 66.9 \\
    $\tau$ = 0.05 & 77.1 & 69.9 & \textbf{56.7} & 67.9 \\
    $\tau$ = 0.1 & \textbf{77.5} & \textbf{71.2} & 56.6 & \textbf{68.4} \\
    \rowcolor{recipehighlight} $\tau$ = 0.2 & 76.9 & 66.8 & 55.9 & 66.5 \\
    $\tau$ = 0.5 & 76.8 & 68.7 & 52.2 & 65.9 \\
    $\tau$ = 1.0 & 76.1 & 68.4 & 49.1 & 64.5 \\
    $\tau$ = 2.5 & 73.4 & 63.9 & 38.4 & 58.6 \\
    $\tau$ = 5.0 & 70.8 & 59.4 & 40.3 & 56.8 \\
    \bottomrule
  \end{tabular}
\end{table}

Low temperatures perform best, with mean accuracy peaking at $\tau = 0.1$. The sharper $\tau = 0.01$ setting yields lower mean accuracy, while mean accuracy decreases steadily above $\tau = 0.2$ and approaches the baseline without ranking optimization at $\tau = 5.0$. These results suggest that the sigmoid must be sharp enough to approximate hard pruning while remaining smooth enough to propagate gradients to units near the threshold.

\subsubsection{Run-to-Run Variability}

Stochastic optimization may yield rankings with different downstream performance across runs. We therefore repeat our optimization with three random seeds on AugReg ViT-B/16. We report the mean and sample standard deviation of top-1 k-NN accuracy on ImageNet-1k and the transfer dataset average in~\autoref{tab:seeds}. For transfer, we first average accuracy across the nine datasets within each seed, then compute the mean and sample standard deviation across seeds. The sample standard deviation is at most 0.65\% on ImageNet-1k at 60\% sparsity and 1.81\% on the transfer average at 40\%. Variability is otherwise below 0.56\%. Performance varies little across these three runs, except for greater variability in transfer accuracy at 40\% sparsity.

\begin{table}[htbp]
  \centering
  \caption{\textbf{Performance is generally stable across random seeds.} Top-1 k-NN accuracy on ImageNet-1k and the transfer average for AugReg ViT-B/16 models pruned with our method, reported as mean $\pm$ sample standard deviation across three seeds. Transfer accuracy is first averaged across the nine datasets within each seed.}
  \label{tab:seeds}
  \resizebox{\columnwidth}{!}{
    \begin{tabular}{lcccccc}
      \toprule
      & \multicolumn{6}{c}{Sparsity} \\
      \cmidrule(lr){2-7}
       & 10\% / 31.7 G & 20\% / 28.2 G & 30\% / 24.8 G & 40\% / 21.2 G & 50\% / 17.8 G & 60\% / 14.2 G \\
      \midrule
      ImageNet-1k & $82.5 \pm 0.14$ & $79.9 \pm 0.14$ & $75.6 \pm 0.17$ & $69.1 \pm 0.09$ & $59.5 \pm 0.54$ & $44.5 \pm 0.65$ \\
      Transfer Average & $78.7 \pm 0.20$ & $76.7 \pm 0.23$ & $73.6 \pm 0.44$ & $68.7 \pm 1.81$ & $64.1 \pm 0.55$ & $55.9 \pm 0.12$ \\
      \bottomrule
    \end{tabular}
  }
\end{table}

\end{document}